\documentclass[11pt]{article}
\usepackage[letterpaper,margin=1in]{geometry}
\usepackage{arxiv}

\IfFileExists{latexml.sty}{\usepackage{latexml}}{\newif\iflatexml\latexmlfalse}

\usepackage{amsmath,amssymb,amsthm}
\usepackage{bm,bbm}

\usepackage{amsmath,amsfonts,bm}

\def\eqref#1{equation~\ref{#1}}

\def\1{\bm{1}}

\def\rmU{{\mathbf{U}}}

\DeclareMathAlphabet{\mathsfit}{\encodingdefault}{\sfdefault}{m}{sl}
\SetMathAlphabet{\mathsfit}{bold}{\encodingdefault}{\sfdefault}{bx}{n}

\newcommand{\R}{\mathbb{R}}

\newcommand{\softmax}{\mathrm{softmax}}

\DeclareMathOperator*{\argmax}{arg\,max}
\DeclareMathOperator*{\argmin}{arg\,min}

\usepackage{url}

\usepackage{bm}
\usepackage{amssymb}
\usepackage[toc,page]{appendix}
\usepackage{titletoc}
\usepackage{graphicx}
\usepackage{booktabs}
\usepackage{enumerate}
\usepackage{natbib}
\usepackage{microtype}
\usepackage{listings}
\usepackage{algorithm}
\usepackage{algorithmic}
\usepackage{comment}
\usepackage{afterpage,wrapfig}
\usepackage[colorlinks,citecolor=blue,urlcolor=blue]{hyperref}
\usepackage{fancyhdr}
\usepackage{multicol}
\usepackage{multirow}
\usepackage{tikz}
\usetikzlibrary{arrows.meta, positioning, calc}

\usepackage{adjustbox}
\usepackage{lineno}
\usepackage{apptools}
\usepackage{bbm}
\usepackage{caption}
\usepackage{subcaption}
\usepackage{xcolor,colortbl}
\newtheorem{proposition}{Proposition}

\newtheorem{lemma}{Lemma}

\title{Bayesian optimization on amortized search space}

\iflatexml
\author{
Tristan Fauvel\textsuperscript{*} \\
Theremia \\
Paris, France
\and
Federico Pavone\textsuperscript{*}\thanks{Corresponding author: \texttt{federicopavone@theremia.health}} \\
Theremia \\
Paris, France
\and
Titouan Guillou \\
Theremia \\
Paris, France
\and
Pau Redon Mu\~{n}oz \\
Theremia \\
Paris, France
\and
Chloé Geoffroy \\
Theremia \\
Paris, France
}
\else
\author{
Tristan Fauvel\footnotemark[1] \\
Theremia \\
Paris, France
\And
Federico Pavone\thanks{Equal contribution.}\hspace{5pt}\thanks{Corresponding author: \texttt{federicopavone@theremia.health}}\\
Theremia \\
Paris, France
\And
Titouan Guillou \\
Theremia \\
Paris, France
\authorbreak
Pau Redon Mu\~{n}oz\\
Theremia \\
Paris, France
\And
Chloé Geoffroy \\
Theremia \\
Paris, France
}
\fi

\begin{document}

\maketitle

\begin{abstract}
In real-world black-box optimization problems, a mechanistic simulator is often available that approximately predicts the behavior of the system as a function of the parameters being optimized. Here, we show how an amortized inverse map of such a simulator can be used in Bayesian optimization to restrict the search space while using a Gaussian process kernel that is aware of the original problem geometry. This allows for efficiently leveraging the low intrinsic dimensionality of the optimization problem. 
We characterize its approximation bias and information gain, and derive an intrinsic-dimension-dependent regret rate under  additional geometric and Gaussian Process assumptions.  We illustrate the method with an application to drug dosing personalization in Parkinson's disease.  We use, as a simulator, a pharmacokinetic/pharmacodynamic (PK/PD) model of a drug, a mechanistic model calibrated on population data with quantified parameter uncertainty. A dosing neural network is trained offline against the model, amortizing the map from patient PK/PD parameters to the regimen predicted to be optimal, and online Bayesian optimization then searches over this learned, low-dimensional representation rather than over raw dosages. Numerical simulations demonstrate the improved performance of our method in this application, as well as under simulator misspecification. 
\end{abstract}

\section{Introduction}
\label{sec:introduction}

We consider the problem of optimizing a system with objective function $U$ and parameters $\mathbf {x}$ in a search space (or design space) $\mathcal{D}\subseteq\mathbb{R}^d$:

\begin{equation}\label{eq:setup}
    x^\star = \argmax_{\mathbf{x}\in\mathcal{D}}
U(\mathbf{x}),
\end{equation}
$U$ is an expensive black-box function, with no available derivative information. Bayesian optimization (BO) is a sample-efficient approach for optimizing black-box functions that are expensive to evaluate, and has been applied across domains as
varied as hyperparameter tuning, robotics, and drug discovery \citep{brochu_tutorial_2010a, shahriari_taking_2016, frazier_tutorial_2018}. At each iteration, a probabilistic surrogate, typically a Gaussian process (GP) \citep{rasmussen_gaussian_2006}, models the unknown objective from past observations, and an acquisition function balances exploration and exploitation to select the next candidate to evaluate. 
Under budgets of only a few tens of experiments, learning an informative surrogate can be difficult even at moderate dimensionality. Although this difficulty has traditionally been attributed to the curse of dimensionality \citep{wang_bayesian_2013,kandasamy_high_2015}, recent evidence identifies prior-induced model complexity as an important and potentially avoidable cause \citep{hvarfner_vanilla_2024}.

We consider the setting where a mechanistic simulator approximating the behavior of the system being optimized is available.
We propose to use the simulator to learn a restriction of the search space. Offline, we train a neural network that amortizes the solution of the simulator-defined optimization problem into a single forward pass; its image $\mathcal{M}\subseteq\mathcal{D}$ is the subset of simulator-optimal candidates. Online, BO searches over the network inputs, using a Gaussian process kernel pulled back from $\mathcal{D}$ to preserve the geometry of the original problem. 
Because the simulator is used offline only, the online optimization loop can be conducted under any form of feedback, including scalar evaluations, preferences, or other types of observations that may not directly correspond to the simulator output.


Personalized drug dosing, our motivating application, is a concrete instance of this setting. Here the simulator is a pharmacokinetic-pharmacodynamic (PK/PD) model, describing how a dosing regimen translates into drug concentration across body compartments (pharmacokinetics) and into a clinical effect (pharmacodynamics). These models are estimated on clinical-trial data, with inter-individual variability captured by random effects on the parameters. The PD model comes with a known target level of efficacy, so the dosing problem can be solved inside the simulator for any patient parameters.
In this setting, the image of the amortized network is a low-dimensional subspace of pharmacologically coherent regimens, so the online search is restricted to regimens that are sensible by construction, which
matters both for patient safety and for keeping the number of trials small. Figure \ref{fig:overview} illustrates our methodology applied to this use-case.

Our contributions can be summarized as follows:
\begin{itemize}
  \itemsep2pt \parskip0pt
  \item We introduce BOASS (Bayesian Optimization on Amortized Search Space), a method where a model is trained to amortize the solution of the simulator-defined optimization problem, and its image defines a low-dimensional set of simulator-optimal candidates over which BO runs online. Within BOASS, we define the pullback kernel for the GP surrogate to make the model aware of amortized map geometry;
  \item We derive regret bounds that highlight the benefits and limitations of this method, decomposing the contribution of the simulator misspecification, the amortization error, and the online optimization error in the regret. We derive an intrinsic-dimension-dependent online optimization regret rate under  additional geometric and Gaussian Process assumptions;
  \item We evaluate BOASS on synthetic experiments and on a dose personalization problem of Levodopa-Carbidopa for Parkinson's disease;
\end{itemize}
 
The remainder of the paper is organized as follows: Section \ref{sec:related_work} discusses related works, Section \ref{sec:method} presents BOASS in a general setting,
Section \ref{sec:regret} contains the regret analysis, and Section \ref{sec:synthetic}-\ref{sec:levodopa} presents experimental results. We conclude with a discussion in Section \ref{sec:discussion}

\section{Related Work}
\label{sec:related_work}

\paragraph{Bayesian optimization, high dimensions and learned search spaces.}
A recurring strategy for high-dimensional or structured domains is to run BO on a lower-dimensional
representation of the raw design space. REMBO \citep{wang_bayesian_2016} optimizes inside a random linear
embedding, assuming that the objective varies along a low-dimensional subspace only; SAASBO
\citep{eriksson21a_sassbo} places a sparsity-inducing prior on the kernel lengthscales, which restricts it
to axis-aligned structure; TuRBO \citep{eriksson19a_turbo} keeps the full space and confines the search to
local trust regions. These methods scale to hundreds of dimensions, but the representation is either drawn
at random or inferred from the observations, requiring a large iteration budget and several restarts, which is not applicable in contexts such as precision medicine.
Alternatively, latent-space BO learns a generative model of the inputs, typically a variational autoencoder, and optimizes in its latent space \citep{gomez-bombarelli_automatic_2018, tripp_sampleefficient_2020, maus_local_2022, stanton_accelerating_2022}; amortized optimization instead trains a network offline to map a problem instance to its solution, replacing per-instance optimization with a forward pass \citep{amos_tutorial_2023}. 
Our approach combines these two ideas: the dosing network is an amortized optimizer trained offline mapping simulator parameters to the system-optimal design, and its image defines the low-dimensional space over which online BO runs.


\paragraph{Leveraging simulators in Bayesian optimization.} Several BO methods have been proposed to exploit mechanistic simulators.
One line of work builds the GP kernel on a latent representation of simulated trajectories, so that configurations with similar simulated behaviour are treated as similar by the surrogate \citep{rai_using_2019, antonova_bayesian_2020}; another uses the simulator to bias the acquisition function, the surrogate mean, or the presumed location of the optimum with an informative prior \citep{hvarfner_pbo_2022, hvarfner_general_2024, boltz_leveraging_2025}. Multi-fidelity BO \citep{kandasamy_gaussian_2016, moss_gibbon_2021} is the most standard way to combine a simulator with real-world evaluations, treating the simulator as a  cheap, low-fidelity approximation of the expensive, high-fidelity real-world objective. This assumption does not necessarily hold in our setting: for example, the PK/PD model predicts the plasma concentration $\mathbf{c}(t)$ or the treatment effect $r(t)$, whereas the feedback we can realistically collect from a patient (a symptom score, a wearable-derived signal, or a pairwise preference), is not a noisier or cheaper version of that same quantity but a genuinely different observation. We instead use the simulator purely offline to restrict the search space (Section~\ref{sec:method}), and let the online BO loop operate directly on whatever feedback is actually available.
Closest to our approach, \citet{granley_humanintheloop_2023} personalize the encoding of visual stimuli into electrical stimulation in visual prostheses, by first training a neural network encoder against a static perception model acting as a decoder, and then fine-tuning a low-dimensional set of patient-specific parameters online with Preferential Bayesian optimization \citep{brochu_bayesian_2010, fauvel_humanintheloop_2022}.
We build on the same idea, with three main differences: we frame the method in a general BO setting, benchmarking it on multiple scenarios and providing regret bounds.  Moreover, we introduce a pullback kernel which allows leveraging the original problem space geometry. 
Finally in our use case, the encoder now amortizes a dosing optimization problem, and the decoder is an ODE-based model rather than a static perception model: the offline objective~\eqref{eq:ddn-loss} is defined on the simulated trajectory, so training backpropagates through the ODE solution (Section~\ref{sec:offline}).

\section{BOASS method}
\label{sec:method}

Let $\Phi\subseteq\mathbb{R}^p$ be the parameter space of the simulator. We assume that the user has a prior knowledge of the simulator parameters in the form of a distribution $p(\phi)$ over it.
For a design $\mathbf{x}\in\mathcal{D}$ and parameters $\phi$, the simulator returns an output $\mathbf{s}(\mathbf{x};\phi): \mathcal{D}\times\Phi \to \mathbb{R}^k$ that predicts the behavior of the system and is differentiable in $\mathbf{x}$ and $\phi$. 
By assumption, the desired system output $\mathbf{s}^\star_\phi$ at parameter $\phi$ is known. 
This turns the simulator into an objective $f_{\mathrm{sim}}(\cdot\,;\phi):\mathcal{D}\to\mathbb{R}$ that scores a design by the proximity of its simulated output to the target, $f_{\mathrm{sim}}(\mathbf{x};\phi) = d\left(\mathbf{s}(\mathbf{x};\phi)-\mathbf{s}^\star_\phi\right)$, where $d$ is some relevant distance on $\mathbb{R}^k$. The BOASS methodology decomposes in the following two steps.

\paragraph{Amortization step.} 
We denote the simulator-optimal design by $\mathcal{E}^\star(\phi) = \argmin_{\mathbf{x}\in\mathcal{D}} f_{\mathrm{sim}}(\mathbf{x};\phi)$, and we assume it exists for every $\phi\in\Phi$.
Computing $\mathcal{E}^\star(\phi)$ anew for each $\phi$ is expensive, so we amortize it offline into a neural network $\mathcal{E}_w:\Phi\to\mathcal{D}$ with weights $w$, trained to minimize the simulator objective in expectation:
\begin{equation}\label{eq:amortization-general}
\min_w\; \mathbb{E}_{\phi\sim p(\phi)}\big[\,f_{\mathrm{sim}}\big(\mathcal{E}_w(\phi);\phi\big)\,\big].
\end{equation}
Its image $\mathcal{M} = \mathcal{E}_w(\Phi) = \{\mathcal{E}_w(\phi):\phi\in\Phi\}\subseteq\mathcal{D}$
is the \emph{amortized search space}, the designs that are approximately simulator-optimal for some $\phi$.

\paragraph{Online optimization step.}
To solve the online optimization problem \eqref{eq:setup}, we optimize the target objective over $\Phi$ instead of $\mathcal{D}$,
\begin{equation}\label{eq:reparam-general}
\phi^\star = \argmax_{\phi\in\Phi}\; U\big(\mathcal{E}_w(\phi)\big),
\end{equation}
and return the design $\mathcal{E}_w(\phi^\star)\in\mathcal{M}$. 
At iteration $n$, we observe a feedback $y_n = U\big(\mathcal{E}_w(\phi_n)\big) + \epsilon_n$ with ${\epsilon_n\overset{\text{iid}}{\sim}\mathcal{N}(0,\sigma_\epsilon^2)}$. A Gaussian process surrogate with kernel $k$ on $\Phi$ is fit to the history $\mathcal{H}_n=\{(\phi_i,y_i)\}_{i\le n}$ of candidates and feedback, and an acquisition function $\alpha$ proposes the next candidate
$\phi_{n+1} = \argmax_{\phi\in\Phi}\; \alpha\big(\phi;\mathcal{H}_n\big)$,
whose design $\mathcal{E}_w(\phi_{n+1})$ is evaluated to return the feedback $y_{n+1}$. 
The kernel $k$ of the surrogate GP can be defined directly on the search coordinates, $k_\Phi(\phi,\phi')$, or on the regimens they produce, $k_\mathcal{D}\big(\mathcal{E}_w(\phi),\mathcal{E}_w(\phi')\big)$, i.e., the pullback of a kernel on $\mathcal{D}$
through the deterministic map $\mathcal{E}_w$. The latter is a valid positive-definite kernel for any positive-definite $k_\mathcal{D}$, being its pullback through a deterministic map. 

BOASS restricts the search space rather than assuming a particular observation model, so nothing in the construction of $\mathcal{M}$ depends on the Gaussian-noise feedback assumed above. 
The proposed method directly applies to other BO settings, including preference feedback, which we use in Section~\ref{sec:levodopa_experiments}, as well as multi-objective optimization and other non-Gaussian observation models.
Restricting the search to $\mathcal{M}$ reduces the optimization dimension from $D=\dim(\mathcal{D})$ to the intrinsic dimensionality of $\mathcal{M}$, which is often much lower, at the cost of a bias whenever $\mathbf{x}^\star\notin\mathcal{M}$. We analyze this trade-off in terms of the regret in Section~\ref{sec:regret}.


\section{Regret analysis under BOASS}
\label{sec:regret}

Let $\widehat{\mathbf{x}}_n = \mathrm{argmax}_{i\leq n}U(\mathbf{x}_i)$ denote the solution to the optimization problem reported after $n$ iterations. 
Let $\mathbf x^\star$ be the maximizer of $U$ over $\mathcal D$, $\mathbf{x}^\star\in\argmax_{\mathbf{x}\in\mathcal{D}}U(\mathbf{x})$, and $\mathbf x^\star_{\mathcal{M}}$ its maximizer over the learned search space $\mathcal{M}$, $\mathbf{x}^\star_{\mathcal{M}}\in\argmax_{\mathbf{x}\in\mathcal{M}}U(\mathbf{x})$. The simple regret after $n$ evaluations decomposes as
$r_n=\Delta_{\mathcal{M}} + r_n^{\mathcal{M}}$, with search-space bias $\Delta_{\mathcal{M}}=U(\mathbf x^\star)-U(\mathbf x^\star_{\mathcal{M}})$, and online optimization error $r_n^{\mathcal{M}}=U(\mathbf x^\star_{\mathcal{M}})-U(\widehat{\mathbf{x}}_n)$.   The following proposition provides a decomposition of the regret bound into online optimization error and regret due to simulator misspecification and  amortization error.

\begin{proposition}
For each $\phi\in\Phi$, let $\mathcal E^\star(\phi)$ be an exact optimizer of the simulator-defined optimization problem. Let $\mathcal M^\star=\mathcal E^\star(\Phi)$ be the simulator-induced search space and $\mathcal M=\mathcal E_w(\Phi)$ the learned search space. 
Assume the following : 
\begin{enumerate}
\itemsep0pt \parskip0pt 
    \renewcommand{\theenumi}{A\arabic{enumi}}
    \renewcommand{\labelenumi}{(\theenumi)}
    \item $U$ is $L_U$-Lipschitz with respect to the design metric $\delta$;
    \item  $ \exists\,\phi^\star_{\mathcal{M}^\star}
    \in \argmax_{\phi\in\Phi} U(\mathcal E^\star(\phi))$ such that, for some $ \varepsilon_{\mathrm{am}} \geq 0$, $\delta\!\left(
        \mathcal E_w(\phi^\star_{\mathcal{M}^\star}),
        \mathcal E^\star(\phi^\star_{\mathcal{M}^\star})
    \right)
    \leq \varepsilon_{\mathrm{am}}$.
\end{enumerate} 
Then
\begin{equation}
    \Delta_{\mathcal M}
    \leq r_n
    \leq
    \underbrace{\Delta_{\mathrm{model}}}_{\text{simulator misspecification}}
    +
    \underbrace{L_U\varepsilon_{\mathrm{am}}}_{\text{amortization error}}
    +   
    \underbrace{
        r_n^{\mathcal{M}}
    }_{\text{online optimization error}},
    \label{eq:theory-overall}
\end{equation}
where $\Delta_{\mathrm{model}}  = U(\mathbf{x}^\star)  - \max_{\mathbf{x}\in\mathcal M^\star} U(\mathbf{x})$ is the simulator misspecification gap, that is, the loss in the best achievable utility from restricting the search to $\mathcal M^\star$, even with perfect amortization.  

\label{prop:regret-decomposition}
\end{proposition}
The proof is given in Appendix~\ref{app:amortization_error}.   
The following proposition provides a regret bound on the online optimization error when using the pullback kernel.

\begin{proposition}
 Let \(\mathcal{E}_w:\Phi\rightarrow\mathcal D\), and let \(\mathcal M=\mathcal{E}_w(\Phi)\). Define the pullback kernel
\[
k_{\mathrm{pb}}(\phi,\phi')
=
k_{\mathcal D}\!\left(\mathcal{E}_w(\phi),\mathcal{E}_w(\phi')\right).
\]
Suppose that: 
\begin{enumerate}
\renewcommand{\theenumi}{A\arabic{enumi}}
\renewcommand{\labelenumi}{(\theenumi)}
\setcounter{enumi}{2}
\itemsep0pt \parskip0pt 
    \item   \label{ass:domain} \(\Phi\subset\mathbb{R}^{p}\) is compact and convex;
    \item $g=U\circ \mathcal{E}_w$ is drawn from  $\mathcal{GP}(0,k_{\mathrm{pb}})$ and satisfies the derivative-tail condition of \cite{srinivas_gaussian_2010};
    \item \label{ass:noise} Observations have the form \(y_t=g(\phi_t)+\epsilon_t\), with independent noise \(\epsilon_t\sim\mathcal N(0,\sigma_\epsilon^2)\). 
\end{enumerate}

Under (\ref{ass:domain})--(\ref{ass:noise}), GP-UCB with the exploration sequence $\beta_t$ of \citet{srinivas_gaussian_2010} satisfies, with high probability,
\begin{equation}
     r_n^{\mathcal M} \leq \frac{R_n^{\mathcal M}}{n}
    = \frac{1}{n}\sum_{t=1}^n
        \left[U(\mathbf x^\star_{\mathcal M})-g(\phi_t)\right]
    = \widetilde O\!\left(
        \sqrt{\frac{\beta_n\gamma_n(\mathcal M;k_{\mathcal D})}{n}}
    \right),
    \label{eq:online-cumulative-regret}
\end{equation}
where $\gamma_n(\mathcal M;k_{\mathcal D})$ is the maximum information gain for a sequence of $n$ designs restricted to $\mathcal{M}$.

To derive a bound on $\gamma_n(\mathcal M;k_{\mathcal D})$ that makes explicit the dependence on the intrinsic dimension of the search space, we use the additional assumptions: 
\begin{enumerate}
\itemsep0pt \parskip0pt 
\renewcommand{\theenumi}{A\arabic{enumi}}
    \renewcommand{\labelenumi}{(\theenumi)}
    \setcounter{enumi}{5}
    \item     \label{ass:manifold} $\mathcal{M}\subset\mathbb R^D$ is a smooth, compact, $\rho$-dimensional embedded submanifold without boundary; 
    \item $k_{\mathcal D}$ is a Matérn-$\nu$ kernel restricted to $\mathcal{M}$;
    \item \label{ass:eigenfunctions} The Mercer eigenfunctions of \(k_{\mathcal D}\) restricted to \(\mathcal M\) satisfy the uniform boundedness condition used in the information-gain bound of \citet{vakili_information_2021}.
\end{enumerate}

If (\ref{ass:manifold})--(\ref{ass:eigenfunctions}) also
hold, the simple regret of the best evaluated design satisfies: 
\begin{equation}
    r_n^{\mathcal M}
    \leq \frac{R_n^{\mathcal M}}{n}
    = \widetilde O\!\left(n^{-\frac{\nu}{2\nu+\rho}}\right),
    \label{eq:online-intrinsic-simple-regret}
\end{equation}
compared with $\widetilde O(n^{-\nu/(2\nu+D)})$ on the full $D$-dimensional input space. 
\label{prop:online-regret}
\end{proposition}

The proof is detailed in Appendix~\ref{app:intrinsic-information-gain}.

By combining Proposition \ref{prop:regret-decomposition} and Proposition \ref{prop:online-regret},  Under the hypotheses of Proposition~\ref{prop:online-regret}, we get the following bounds on the simple regret: 
\begin{equation}
    \Delta_{\mathcal M}
    \leq r_n
    \leq
    \underbrace{\Delta_{\mathrm{model}}}_{\text{simulator misspecification}}
    +
    \underbrace{L_U\varepsilon_{\mathrm{am}}}_{\text{amortization error}}
    +   
    \underbrace{
        \widetilde O\!\left(n^{-\frac{\nu}{2\nu+\rho}}\right)
    }_{\text{online optimization error}}. 
\end{equation}
\label{eq:regret-decomposition-dimension-dependent}

Equation \ref{eq:regret-decomposition-dimension-dependent}  makes explicit the central trade-off of amortized search-space restriction: the simulator and its amortization can introduce bias, while restricting online BO to a subspace of lower intrinsic dimension can accelerate optimization.

Kernel $k_\Phi$ defined directly on the $p$-dimensional simulator parameters has the generic Matérn rate $\widetilde O(n^{p/(2\nu+p)})$, with $\rho \leq p$. However, the pullback kernel is geometry-aware. In particular, two points may be far away in the simulator parameter space,  $ \|\phi - \phi' \| \gg 0$, but produce almost identical regimens, $\|\mathcal{E}_w(\phi) - \mathcal{E}_w(\phi')\| \approx 0 $. While a Matérn kernel defined on $\Phi$ may regard these as nearly independent, the pullback kernel captures their similarity : $k_{\mathcal{D}}(\mathcal{E}_w(\phi),\mathcal{E}_w(\phi')) \approx k_{\mathcal{D}}(\mathcal{E}_w(\phi),\mathcal{E}_w(\phi))$.


\section{Synthetic experiments} \label{sec:synthetic}

We illustrate the trade-off of Section~\ref{sec:regret} on a problem where the dimension $D$ of the design space and the search-space bias can be set independently. We embed the Branin function in $\mathcal{D}=[-1,1]^D$ through the exact map $\mathcal{E}^\star(\phi)=(\phi,m(\phi))$, $\phi\in\Phi=[-1,1]^2$, with a custom $m:\Phi \rightarrow\mathbb{R}^{D-2}$ (see \eqref{eq:fixed-geometry-features}, Appendix \ref{app:synthetic_experiments}). We minimize $f:\mathcal{D}\rightarrow \mathbb{R}$ defined as $f_\delta(\mathbf{x})=\widetilde{B}(\mathbf{x}_{1:2})+{\|\mathbf{x}_{3:D}-m(\mathbf{x}_{1:2})-\delta\,\mathbf{e}_1\|_2^2}$, with $\widetilde{B}$ the Branin function rescaled to $\Phi$ and shifted to minimum $0$. 
Since $\min_{\mathcal{D}} f_\delta=0$ and $f_\delta(\mathcal{E}^\star(\phi))=\widetilde{B}(\phi)+\delta^2$, the search-space bias is $\Delta_{\mathcal{M}}=\delta^2$. We focus on BOASS with the kernel $k_\Phi$ (BOASS-$k_\Phi$), and compare it with random search on $\Phi$ and with three methods searching $\mathcal{D}$: standard BO (BO-on-$\mathcal{D}$), SAASBO \citep{eriksson21a_sassbo} and TuRBO-1 \citep{eriksson19a_turbo}. All methods are initialised in $\mathcal{M}=\mathcal{E}(\Phi)$.

Table~\ref{tab:regret-delta-D_main} reports the simple regret after $60$ evaluations, which for BOASS-$k_\Phi$ does not depend on $D$ by construction. Without bias, BOASS-$k_\Phi$ has the lowest regret for all $D$. 
With a bias, BOASS-$k_\Phi$ cannot go below $\Delta_\mathcal{M} = \delta^2$ and the methods searching $\mathcal{D}$ gradually do better at smaller $D$ as the bias increases. Restricting the search to $\mathcal{M}$ is thus beneficial as long as the bias is below the regret that a search over $\mathcal{D}$ reaches within the budget, and this regret grows with $D$.
We report results with additional values of $\delta$ in Appendix~\ref{app:synthetic_experiments}, together with further details and extra experiments on the amortization error and the comparison between the kernels $k_\mathcal{D}$ and $k_\Phi$.

\begin{table}[t]
\caption{BO benchmark for the Branin function optimization, across search-space biases and ambient dimension. Simple regret at iteration $60$, in units of $10^{-3}$: mean $\pm$ s.e.m. over $25$ seeds. $\Delta_{\mathcal{M}}=\delta^{2}$ (in units of $10^{-3}$) is the smallest regret attainable by a method restricted to $\Phi$, and is $0$ for the methods that search $\mathcal{D}$ directly. Within each block, \textbf{bold} marks, for each $D$ column, the lowest mean together with every method whose $\hat m \pm \mathrm{s.e.m.}$ interval overlaps that of the lowest.}
\label{tab:regret-delta-D_main}
\centering
\footnotesize
\setlength{\tabcolsep}{2pt}
\begin{tabular}{clcrrrr}
\toprule
$\delta$ & Method & \multicolumn{1}{c}{$\Delta_{\mathcal{M}}$} & \multicolumn{1}{c}{$D=6$} & \multicolumn{1}{c}{$D=8$} & \multicolumn{1}{c}{$D=16$} & \multicolumn{1}{c}{$D=32$} \\
\midrule
 & BOASS-$k_\Phi$ & $0$ & $\bm{6.3 \pm 2.5}$ & $\bm{6.3 \pm 2.5}$ & $\bm{6.3 \pm 2.5}$ & $\bm{6.3 \pm 2.5}$ \\
\rowcolor[gray]{0.92}\cellcolor{white} & BO-on-$\mathcal{D}$ & - & $37.2 \pm 7.1$ & $74.7 \pm 18.0$ & $149.1 \pm 26.0$ & $687.5 \pm 120.0$ \\
 & SAASBO & - & $\bm{9.0 \pm 2.0}$ & $21.6 \pm 3.7$ & $177.0 \pm 36.0$ & $489.7 \pm 72.0$ \\
\rowcolor[gray]{0.92}\cellcolor{white} & TuRBO-1 & - & $\bm{7.7 \pm 1.6}$ & $31.2 \pm 7.6$ & $275.2 \pm 48.0$ & $609.8 \pm 99.0$ \\
\multirow{-5}{*}{$0$} & Random search on $\Phi$ & $0$ & $12.1 \pm 2.3$ & $12.1 \pm 2.3$ & $12.1 \pm 2.3$ & $12.1 \pm 2.3$ \\
\midrule
 & BOASS-$k_\Phi$ & $62.5$ & $68.8 \pm 2.5$ & $68.8 \pm 2.5$ & $\bm{68.8 \pm 2.5}$ & $\bm{68.8 \pm 2.5}$ \\
\rowcolor[gray]{0.92}\cellcolor{white} & BO-on-$\mathcal{D}$ & - & $41.5 \pm 11.0$ & $81.0 \pm 20.0$ & $154.7 \pm 22.0$ & $761.8 \pm 110.0$ \\
 & SAASBO & - & $\bm{11.4 \pm 2.3}$ & $\bm{19.0 \pm 3.0}$ & $108.3 \pm 17.0$ & $534.0 \pm 59.0$ \\
\rowcolor[gray]{0.92}\cellcolor{white} & TuRBO-1 & - & $\bm{8.3 \pm 1.6}$ & $28.5 \pm 6.2$ & $292.2 \pm 54.0$ & $636.7 \pm 110.0$ \\
\multirow{-5}{*}{$0.25$} & Random search on $\Phi$ & $62.5$ & $74.6 \pm 2.3$ & $74.6 \pm 2.3$ & $74.6 \pm 2.3$ & $74.6 \pm 2.3$ \\
\midrule
 & BOASS-$k_\Phi$ & $250$ & $256.3 \pm 2.5$ & $256.3 \pm 2.5$ & $256.3 \pm 2.5$ & $\bm{256.3 \pm 2.5}$ \\
\rowcolor[gray]{0.92}\cellcolor{white} & BO-on-$\mathcal{D}$ & - & $48.1 \pm 8.2$ & $69.7 \pm 14.0$ & $\bm{184.1 \pm 31.0}$ & $853.8 \pm 92.0$ \\
 & SAASBO & - & $\bm{8.8 \pm 2.0}$ & $\bm{22.2 \pm 4.2}$ & $\bm{140.1 \pm 24.0}$ & $637.2 \pm 57.0$ \\
\rowcolor[gray]{0.92}\cellcolor{white} & TuRBO-1  & - & $\bm{7.3 \pm 1.2}$ & $40.3 \pm 8.0$ & $308.3 \pm 51.0$ & $764.7 \pm 110.0$ \\
\multirow{-5}{*}{$0.5$} & Random search on $\Phi$ & $250$ & $262.1 \pm 2.3$ & $262.1 \pm 2.3$ & $262.1 \pm 2.3$ & $262.1 \pm 2.3$ \\
\bottomrule
\end{tabular}

\end{table}

\section{Use-case: personalized Levodopa dosing for Parkinson's disease}
\label{sec:levodopa}

As a concrete case study, we consider personalised dosing of Levodopa-Carbidopa , the most common treatment for Parkinson's disease \citep{dhall_advances_2016,
houghton_treatment_2019}. As Parkinson's progresses, the therapeutic window (the safe and effective range of a drug's dose) narrows \citep{pezzoli_levodopa_2010}, and advanced patients increasingly struggle to find a dosage that controls symptoms without triggering side effects, see Appendix \ref{app:parkinson} for further details.
Traditionally, clinicians optimize dosing through a time-consuming trial-and-error titration process, guided by protocols and experience. We tackle here the problem of personalisation with Bayesian optimisation.

\paragraph{Problem setup.}
We fix the maximum number of daily drug intakes $M=7$ and denote by $\mathbf{d} = [d_1, \dots, d_M]^\intercal$ and $\mathbf{t}=[t_1, \dots, t_M]^\intercal$ the dose amounts and the intake times.
For a given patient, we aim at solving $(\mathbf{d}^\star, \mathbf{t}^\star) = \argmax_{(\mathbf{d},\mathbf{t})\in\mathcal{D}}
U(\mathbf{d},\mathbf{t})$, 
where $U$ is the black-box patient utility, a proxy for treatment efficacy. The observations collected from the patient, which we call \emph{feedback}, can take different forms, such as symptoms rating or pairwise preference between two regimens $(\mathbf{d}_k, \mathbf{t}_k) \succ (\mathbf{d}_j, \mathbf{t}_j)$.  
We represent the space of admissible dosing regimens as $\mathcal{D}=\{(\mathbf{d},\mathbf{t}) \mid 0\leq d_i \leq
d_{\max};\; 0\leq t_1,\; t_{i}<t_{i+1},\; t_M \leq T\}$. Reflecting common clinical constraints for this treatment, we require intake times to be at least $\Delta=60\,\text{min}$ apart and to lie within the intake window $[t_0, t_f]$, and dose amounts to lie in $[0, d_{\max}]$, with $d_{\max}=250\,\text{mg}$. 
Microtablets are available in increments of $\Delta_\text{tab}=5$~mg, so we round each dose to the nearest multiple of $\Delta_\text{tab}$ before administering the regimen to the patient.
We optimize dosing regimen sequentially, assuming washout of the effect of the previous regimen between one day and the next. 
 
\subsection{Mechanistic PK/PD models for dosing}
\label{sec:pkpd}

As a differentiable mechanistic simulator, we use a PK/PD model, which describes how a dosing regimen translates into drug concentration over time (pharmacokinetics, PK) and, in turn, into clinical effect (pharmacodynamics, PD).
Let $\phi\in\Phi\subset\mathbb{R}^p$ denote a vector of patient-specific PK/PD parameters (absorption and elimination rates, clearance, volumes of distribution, etc.) in the parameter space $\Phi$. 
The PK model describes the concentration trajectories $\mathbf{c}(t; \phi)$ for $t>0$ across the $K$ compartments (e.g., plasma, effect site, \ldots) as the solution of a system of  ODEs:
\begin{equation}
\label{eq:pkpd-ode}
\begin{cases}
c_1(0) = g(d;\phi), \quad c_i(0) = c_{i0}, \quad i=2,\dots,K \\
\dot{\mathbf{c}}(t) = f_\text{PK}\big(\mathbf{c}(t); \phi\big),
\end{cases}
\end{equation}

where $c_1(t)$ is the concentration in the compartment where the dose is administered, $c_i(t)$ are the concentrations in the other compartments. 
The initial condition $g(d;\phi)$ maps the dose amount to an initial concentration in the dosing compartment. 
The PD output $r(t;\phi) = f_\text{PD}\big(c_e(t;\phi); \phi\big)$ maps the concentration in the effect compartment $e$, denoted by by $c_e(t)$, to a clinical quantity such as a symptom or treatment-response score.
These models are estimated on clinical-trial data, and inter-individual variability in the parameters is modeled through random effects, whose distribution define $p(\phi)$ (Appendix~\ref{app:pkpd_levodopa}).
For our experiments, we combine the PK model of \citet{senek_population_2018} with the PD model of \citet{westin_pharmacokineticpharmacodynamic_2011}, whose output is the treatment response scale (TRS, ranging from $r=-3$ for parkinsonism to $r=+3$ for dyskinesia, with $r^\star=0$ the optimal response).

Let $c^\star_\phi$ be the lowest drug concentration that achieves the target clinical state of the patient $r^\star$ under the patient's pharmacodynamics, $c^\star_\phi=\min \{c>0: f_\text{PD}\big(c;\phi\big)=r^\star\}$. 
We assume that $c^\star_\phi$ always exists, meaning that the patient can be efficiently treated with the drug.
We define the model-optimal regimen as:
\begin{equation}
\label{eq:model-optimal}
(\mathbf{d}^\star_\phi, \mathbf{t}^\star_\phi) \;=\; \argmin_{(\mathbf{d},\mathbf{t})\in\mathcal{D}} \;\int_0^T \lvert
c_{e}(s;\phi\mid (\mathbf{d},\mathbf{t})) - c^\star_\phi \rvert^2\,ds,
\end{equation}
where $c_{e}(s;\phi\mid (\mathbf{d},\mathbf{t}))$ is the concentration in $e$ at time $s$ induced by the regimen $(\mathbf{d},\mathbf{t})$. In practice, it is computed by integrating the ODE system piecewise between successive intakes, updating the initial conditions at each dosing time, see Appendix \ref{app:ode_intakes} for the computation details.

\subsection{Offline training of the dosing network}
\label{sec:offline}

We start by training a \emph{dosing neural network} that maps a patient's PK/PD parameters $\phi$ to the regimen the PK/PD model predicts to be optimal for that patient (panel~A Figure~\ref{fig:overview} ).  
We define the training loss at $\phi$ as the average squared deviation of the induced concentration from the target $c^\star_\phi$ over the treatment horizon $[0,T]$, and approximate it by trapezoidal quadrature over a fine grid of $J+1$ time points $[0=\tau_0,\dots,\tau_J=T]$:
\begin{equation}
\label{eq:ddn-loss}
\mathcal{L}(w;\phi) \;=\; \frac{1}{T}\int_0^T \lvert c_e\left(s;\phi\mid \mathcal{E}_w(\phi)\right) - c^\star_\phi
\rvert^2\,ds \approx \frac{1}{T} \sum_{j=1}^J \frac{\epsilon(\tau_j;\phi) + \epsilon(\tau_{j-1};\phi)}{2} (\tau_j - \tau_{j-1}),
\end{equation}
with $\epsilon(\tau_j;\phi) = \lvert c_e\left(\tau_j;\phi\mid \mathcal{E}_w(\phi)\right) - c^\star_\phi \rvert^2$.
We train $\mathcal{E}_w$ on a training set of virtual patients $\{\phi_i\}_{i=1}^N$ sampled from the population distribution $p(\phi)$, minimizing the Monte Carlo estimate of the average loss over the population $\mathcal{E}_{\phi\sim p(\phi)} \left[\mathcal{L}(w;\phi)\right] \;\approx\; \frac{1}{N}\sum_{i=1}^N \mathcal{L}(w;\phi_i)$.
Since the data is generated synthetically, we can train the network with arbitrarily large $N$. Panel A in Figure~\ref{fig:overview} gives a schematic of the training procedure.

Training the network requires backpropagating the gradient of the loss \eqref{eq:ddn-loss} through the ODE solution. The network output enters the initial conditions of each integration interval (see \eqref{eq:pkpd-ode-piecewise}, Appendix \ref{app:ode_intakes}). 
Following the recommendations of \citet{kidger_neural_2021, kidger_diffrax_2022}, we take the discretize-then-optimize approach, which backpropagates through the discretized ODE solution produced by the numerical solver.


The network takes the PK/PD parameters $\phi$ as input and returns $2M$ values: the dose amounts $\{d_i\}_{i=1}^M$, the lags between consecutive intakes $\{t_i - t_{i-1} - \Delta\}_{i=2}^M$, and the slack $t_f - t_M$ left between the last intake and the end of the intake window.
It is a feedforward network, trained on $20{,}000$ samples from $p(\phi)$ ($80/20$ train/validation split). We provide more details on the architecture and the training procedure in Appendix~\ref{app:dosingnet}.

\subsection{Online Bayesian optimization}
\label{sec:online}

In this use case, the dose regimens have
$D=13$ free coordinates, and $p=\dim(\Phi)=14$ PK/PD parameters. However, $\mathcal{M}$ is low dimensional due to structural non-identifiability of the PK/PD parameters from the optimal regimen. We provide provide a detailed analysis in Appendix~\ref{app:intrinsic_dimension}, together with a study of the Jacobian of $\mathcal{E}_w$ which gives us an estimate the intrinsic dimension of $\mathcal{M}$ between $3$ and $5$, with a bound $\dim(\mathcal{M})\leq7$. This makes this use case a relevant example of
optimization over a low-dimensional amortized search space.

For the optimization, we add two dosage constraints of clinical relevance: a minimum daily dose of $D_{\min}=320$~mg, and a variation of the total daily dose of at most $40\%$ with respect to the previous regimen. The first constraint restricts $\mathcal{D}$ but not $\mathcal{M}$, as it is already satisfied by the regimens the PK/PD model deems optimal. The second one limits the search to total-dose neighborhoods of the previous regimen, as a form of clinical prudence relevant to realistic applications. We enforce both constraints for all methods by restricting the optimization of the acquisition function at each step (Appendix~\ref{app:bo}).

\subsection{Experiment settings and baselines}
\label{sec:levodopa_experiments}
\begin{table}[t]
\caption{The four evaluation settings differ in the query submitted to the patient, a pair of regimens in the
\emph{preference feedback} setting and a single regimen in the other three, and in the utility that generates the
feedback, either PK/PD or misspecified. In the \emph{preference feedback} setting, $\rmU$ and $\rmU'$ denote the
utilities of the two regimens in the query.
}
\label{tab:settings}
\centering
\small
\setlength{\tabcolsep}{4pt}
\begin{tabular}{@{}lc@{\hskip 14pt}ccc@{}}
\toprule
& preference & concentration & misspecified & fully misspecified \\
\midrule
Query & $\{(\mathbf{d},\mathbf{t}),(\mathbf{d}',\mathbf{t}')\}$ & $(\mathbf{d},\mathbf{t})$ & $(\mathbf{d},\mathbf{t})$ & $(\mathbf{d},\mathbf{t})$ \\
Feedback &  $y \sim B\left[\Psi\left((\rmU+\epsilon-\rmU'-\epsilon')/\nu\right)\right]$& $y = \rmU+\epsilon$& $y = \rmU+\epsilon$& $y = \rmU+\epsilon$\\
Noise & $\epsilon,\epsilon'\overset{\text{iid}}{\sim}\mathcal{N}(0,0.05)$ & $\epsilon\sim\mathcal{N}(0,0.1)$ & $\epsilon\sim\mathcal{N}(0,0.2)$ & $\epsilon\sim\mathcal{N}(0,0.2)$ \\
Utility $\rmU$ & $\rmU_\text{pkpd}$ & $\rmU_\text{pkpd}$ & $\rmU_\text{oracle}$ & $\rmU_\text{oracle}$ \\
Oracle regimen & -- & -- & perturbed $\mathcal{E}_w(\phi)$ & random in $\mathcal{D}$ \\
Maximizer of $\rmU$ & in $\mathcal{M}$ & in $\mathcal{M}$ & outside $\mathcal{M}$ & outside $\mathcal{M}$ \\
\bottomrule
\end{tabular}

\end{table}

We evaluate the online optimization on $100$ held-out patients, under two well-specified settings and two misspecified settings, summarized in Table~\ref{tab:settings}. The settings differ in the query submitted to the patient, in the feedback returned, and in the latent utility generating it. In the well-specified settings, the utility $\rmU_\text{pkpd}$ is the negative concentration error induced by the regimen under the chosen PK/PD model, and 
in the misspecified settings is $\rmU_\text{oracle}$ the negative
distance to an oracle regimen $(\mathbf{d}^\text{or},\mathbf{t}^\text{or})$:
\begin{equation}
\label{eq:utility}
\begin{aligned}
\rmU_\text{pkpd}(\mathbf{d},\mathbf{t},\phi) &= -\frac{1}{T}\int_0^T \lvert c_e\left(s;\phi\mid (\mathbf{d},\mathbf{t})\right) - c^\star_\phi
\rvert^2\,ds, \\
\rmU_\text{oracle}(\mathbf{d},\mathbf{t}) &= -\delta\big[(\mathbf{d},\mathbf{t}),(\mathbf{d}^\text{or},\mathbf{t}^\text{or})\big],
\end{aligned}
\end{equation}
where $\phi$ is the patient's true PK/PD parameter vector, and
${\delta^2\big[(\mathbf{d},\mathbf{t}),(\mathbf{d}',\mathbf{t}')\big] \;=\; \sum_{q=1}^Q \big(\sum_{i=1}^M d_i
\mathbf{1}_{t_i\leq \lambda_q} - \sum_{i=1}^M d'_i \mathbf{1}_{t'_i\leq \lambda_q}\big)^2}$
measures how much two cumulative dose profiles differ on an equispaced grid of time points
$0=\lambda_0<\lambda_1<\dots<\lambda_Q=T$, with $Q=20$.
The four settings are the following:

\begin{enumerate}
\itemsep0pt \parskip0pt
    \item \textbf{concentration feedback}: the feedback is a noisy observation of $\rmU_\text{pkpd}$;
    \item \textbf{preference feedback}: the feedback is a pairwise comparison between two regimens, drawn from a  probit model on $\rmU_\text{pkpd}$. At iteration $n$, the comparison is between the regimens tested at iterations $n$ and $n-1$. That is, $y_n\sim B(p_n)$ with $p_n = P((\mathbf{d}_n, \mathbf{t}_n) \succ (\mathbf{d}_{n-1}, \mathbf{t}_{n-1})) = \Psi\left((U(\mathbf{d}_{n}, \mathbf{t}_{n}) - U(\mathbf{d}_{n-1}, \mathbf{t}_{n-1}))/\nu\right) $, where $\Psi$ the normal c.d.f. , and $\nu = 1$ ;
    \item \textbf{misspecified feedback}: the feedback is a noisy observation of $\rmU_\text{oracle}$,
    with a biased oracle regimen obtained by perturbing the doses of $\mathcal{E}_w(\phi)$, the regimen the network
    returns on the patient's true $\phi$ (Appendix~\ref{app:settings}).
    By construction, the oracle dosage is outside, but close, to $\mathcal{M}$.
    \item \textbf{fully misspecified feedback}: the feedback is as in the previous setting, but the
    oracle regimen is now sampled at random in
    $\{(\mathbf{d},\mathbf{t})\in\mathcal{D}: \sum_{i=1}^M d_i \geq D_{\min}\}$.
\end{enumerate}

The optimization loop starts with the output of the dosing network on the population-average PK/PD parameters, and two random samples in $\mathcal{M}$ lying in the $40\%$-total-dose neighborhood of that regimen. 
The optimization then runs for a total of 50 iterations, using log noisy EI as acquisition function \citep{ament_unexpected_2023} in all settings except
\emph{preference feedback}, where we use EUBO \citep{lin_preference_2022}.

We compare the two kernel variants of our method to several baselines.
We consider the following methods:

\begin{itemize}
\itemsep0pt \parskip0pt
\item \textbf{BOASS-$k_\mathcal{D}$}, \textbf{BOASS-$k_\Phi$}: BO over $\Phi$ with, respectively, the pullback kernel $k_\mathcal{D}\big(\mathcal{E}_w(\phi),\mathcal{E}_w(\phi')\big)$ and the kernel $k_\Phi(\phi,\phi')$ on the PK/PD parameters;
\item \textbf{BOASS-random-$k_\mathcal{D}$}: BO over $\Phi$ with the pullback kernel $k_\mathcal{D}$ and random acquisition;
\item \textbf{BO-on-$\mathcal{D}$}, \textbf{BO-on-$\mathcal{D}$-DSP}: BO over $\mathcal{D}$ with the kernel $k_\mathcal{D}$, respectively with a lengthscale prior $\Gamma(2, 2)$ and a dimension-scaled prior (DSP) $\mathcal{LN}\big(\sqrt{2}+\frac{1}{2}\ln{D}, \sqrt{3} \big)$, following \cite{hvarfner_vanilla_2024};
\item \textbf{$\pi$BOASS-on-$\mathcal{D}$} and \textbf{$\pi$BOASS-on-$\mathcal{D}$-strong}:
as a soft counterpart to the amortized search space, we optimize over $\mathcal{D}$ with a $\pi$BO prior over the location of the optimum \citep{hvarfner_pbo_2022}, informed by the dosing network.
For each patient, the prior $\pi$ is a Gaussian mixture model fitted to samples $\mathcal{E}(\phi_s)$ of dosing regimens, with $\phi_s\sim p(\phi)$. At iteration $n$, the acquisition function is $\alpha(x)\bigl(\pi(x)+\epsilon_\pi\bigr)^{\gamma_n}$,
with $\gamma_n=\beta/n$ and $\epsilon_\pi=10^{-12}$. We use $\beta=10$, following \citet{hvarfner_pbo_2022}, for the strong prior (\textbf{-strong}), and $\beta=0.808$, set by our own calibration procedure, see Appendix~\ref{app:pibo} for details. This variant can be seen as a robustified version of BOASS, which we expect would be less sensitive to search space bias;
\item \textbf{SAASBO-on-$\mathcal{D}$}, \textbf{TuRBO-on-$\mathcal{D}$}: as state-of-the-art high-dimensional-BO methods that do not include any PKPD information, we have included SAASBO (\cite{eriksson21a_sassbo}) and TuRBO (\cite{eriksson19a_turbo}). For TuRBO, we have used TuRBO-1 with tolerances $(\tau_{\text{succ}}, \tau_{\text{fail}}) = (3, 8)$: lower $\tau_{\text{fail}}$ than the recommended $\tau_{\text{fail}} = 14$, to compensate for the low iteration budget.
\item \textbf{PCA-BO-on-$\mathcal{D}$}: BO over the leading principal components of the regimens returned by the dosing network. This is a linear counterpart to the amortized search space, adapted from PCA-BO \citep{raponi2020pcabo}. See Appendix~\ref{app:pcabo} for details.
\item \textbf{MIPD-EIG}: Model-Informed Precision Dosing (MIPD) with Expected Information Gain (EIG). The dose is optimized based on the PK/PD model whose posterior distribution of the  parameters is inferred using patient feedback or measurements. In this approach, the PK/PD model is combined with an observation model to define a likelihood function, and the prior is defined as the distribution of the parameters in the population. The next candidate dosing regimen is the one with the largest expected information gain about the parameters. See Appendix \ref{app:mipd_eig} for details. 
\item \textbf{Oracle parameters}: the parameters $\mathbf{x}$ returned by directly optimizing $\mathbf{x}$ on the patient-specific simulator with parameters $\phi$, using L-BFGS-B optimizer.
In the two misspecified settings, this is the biased oracle regimen.
\end{itemize}
For all methods, the search space, either $\mathcal{D}$ or $\Phi$, is rescaled to the unit box (Appendix~\ref{app:bo}), and $k_\mathcal{D}$ and $k_\Phi$ are Matérn 5/2 kernels with automatic relevance determination, over the regimen coordinates and over the PK/PD parameters.
$k_\mathcal{D}$ is defined on the non-discretized output of the dosing network, although the patient feedback depends on the discretized doses. Differentiability is required for optimizing the acquisition function and fitting the GP hyperparameters.
Because the discretization interval is small relative to the range of clinically relevant dose variation, we expect the resulting kernel approximation to be minor. All reported utilities and regrets are evaluated after rounding. GP hyperparameters are re-estimated at each iteration by maximizing the marginal likelihood.

\subsection{Results}
\label{sec:results}

\begin{figure}
\centering
\includegraphics[width=\linewidth]{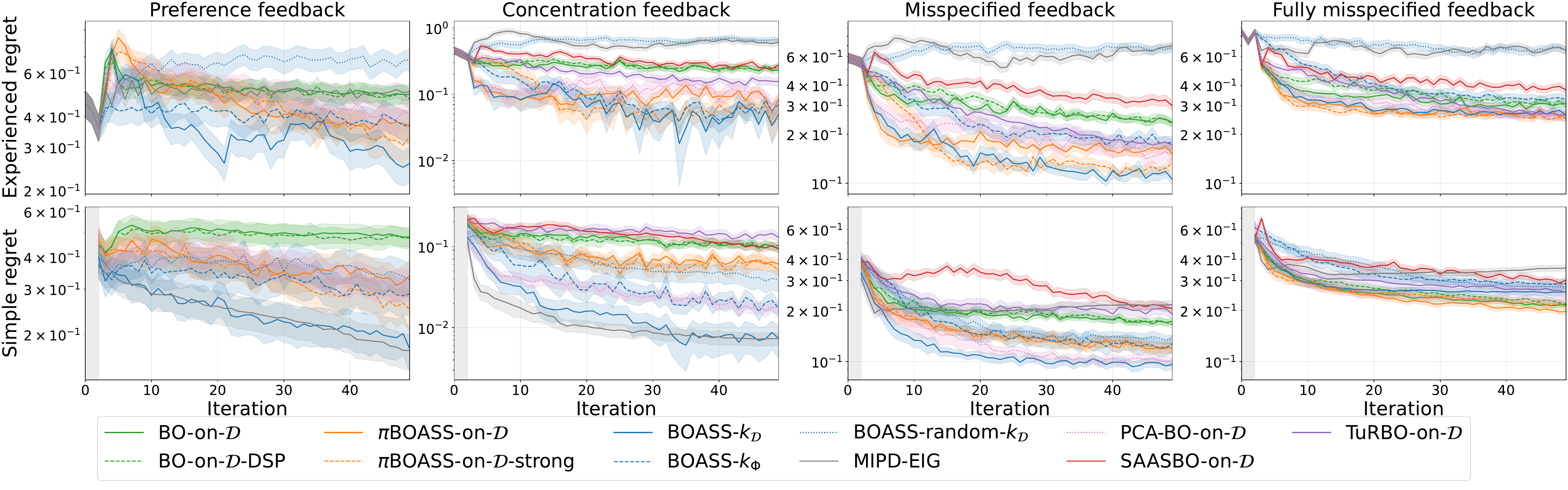}
\caption{
Results from the Levodopa-Carbidopa experiments (evaluation settings of Table~\ref{tab:settings}).
Top row: experienced regret to the oracle regimen.
Bottom row: simple regret for the posterior mean maximizer at each iteration computed after the initialisation, represented by the grey band.
Averages are over 100 simulated patients. Shaded bands: 10th-90th percentile of the estimated mean, assuming Gaussian regret.}
\label{fig:regret}
\end{figure}

The dosing amortized optimization yields results nearly indistinguishable from direct optimization using L-BFGS-B, at a fraction of the computational cost (see analysis in Appendix \ref{app:amortization_results}).

In our use-case, two forms of regret are relevant, both measured at each iteration as a utility gap to the oracle regimen: the \emph{experienced regret} refers to the regimen prescribed to the patient, the \emph{simple regret} to the estimated optimal regimen, i.e.\ the posterior mean maximizer. The posterior mean is maximized under the constraint that the daily dose is greater than the minimum daily dose $D_\text{min}$.
Figure~\ref{fig:regret} reports the two regrets over the $100$ held-out patients.
In the two well-specified settings, \emph{concentration feedback} and \emph{preference feedback}, optimizing over the simulator-induced search space (methods \emph{BOASS-*}) brings a clear advantage on both regrets over all baselines, except MIPD-EIG which has comparable simple regret, but much worse experienced regret. The \emph{preference feedback} setting gives the worst experienced regret trajectories for every method, which is expected since the surrogate is harder to learn from pairwise comparisons than from direct observations of the utility.

The two misspecified settings quantify the cost of restricting the search to $\mathcal{M}$.
In the \emph{misspecified feedback} setting, where the optimum lies outside $\mathcal{M}$ by construction, BOASS variants are still the best performing. Here the search-space bias is small compared to the gain in sample efficiency obtained by searching a lower-dimensional space.
The \emph{fully misspecified feedback} setting introduces a stronger bias: BOASS-
$k_\mathcal{D}$ and  BOASS-$k_\mathcal{\Phi}$ converge to a suboptimal regimen and are outperformed by baselines optimizing over the full space $\mathcal{D}$. The experienced regret of BOASS, however, matches that of the best baselines. $\pi$BOASS is more robust to strong misspecification, as its search and the posterior mean maximizer are not restricted to $\mathcal{M}$.
Between the two kernel options, \emph{BOASS-$k_\mathcal{D}$} is consistently better than \emph{BOASS-$k_\Phi$}. The mapping is not injective, and the pullback kernel $k_\mathcal{D}$ accounts for this property whereas the kernel on $\Phi$ does not. 

All high-dimensional BO methods are consistently outperformed by the BOASS methods and by BO-on-$\mathcal{D}$. TuRBO is designed for settings with a large evaluation budget split across several trust-region restarts. Our setting is the opposite, since the observation budget is very small. SAASBO is the worst-performing baseline. A likely explanation is its sparsity-inducing prior on the lengthscales, which assumes that only a few input coordinates are relevant. In our problem, every coordinate is a dose intake, so all of them are relevant. Finally, the dimension-scaled lengthscale prior (BO-on-$\mathcal{D}$-DSP) performs comparably to the $\Gamma(2, 2)$ prior, with no significant difference between the two.


PCA-BO-on-$\mathcal{D}$ improves markedly on BO-on-$\mathcal{D}$ in the well-specified and misspecified settings.
It nevertheless remains behind BOASS-$k_\mathcal{D}$ in the two well-specified settings. Like BOASS, it loses its advantage under full misspecification, where the optimum lies far from the search subspace.
MIPD-EIG shows very little robustness to misspecification, which can be explained by the fact that it's aiming at maximizing information gain about the parameters of the misspecified model, rather than maximizing the patient's utility. This is confirmed by supplementary experiments (Figure \ref{fig:mipd_comparison}) in which we misspecified the observation model.

\section{Discussion}
\label{sec:discussion}

In this paper, we introduced BOASS, a BO approach that leverages mechanistic simulators to effectively reduce the search to a lower dimensional subspace, while preserving the initial problem geometry for learning GP surrogates. We show through theoretical analysis, synthetic experiments and a concrete case study of individualized drug dosing, that BOASS and $\pi$BOASS improve sample-efficiency of BO, outperforming all baselines.
Performance gains can be explained by the fact that the maximum information gain is dependent on the amortized search space intrinsic dimension.  However, we note that the assumptions we used for deriving the online regret rate using established GP posterior contraction rates rely on geometric and Gaussian Process assumptions that remain hard to justify in practice.
Sharper regret bounds could be obtained when the simulator-optimal parameters locally vary along only fewer independent directions than the number of mechanistic model input parameters. We leave the theoretical analysis of these sharper bounds for future work.

However, because the search is confined to the subspace, a strongly misspecified model can limit the search to a subset excluding the true optimum.  The method thus trades-off search space bias with search space dimensionality. 
We showed it remains relatively robust to misspecification of the simulator, the advantage of searching a lower-dimensional space over a finite time horizon being able to compensate for the potential bias introduced. We also introduce a variant, $\pi\text{BOASS}$, that is more robust to such misspecification, and is recommended if there is not enough evidence that the simulator can be trusted.

BOASS is especially relevant in a setting of moderate dimensionality with very small evaluation budget, for which standard high-dimensional BO methods are not suited. Moreover, the use of such an amortized search space could be generalized beyond BO, e.g. for evolutionary optimization, in multi-arm bandits or in Reinforcement Learning.
In the case of Parkinson’s disease, most previous work on algorithmic dosing and prescription is retrospective or in silico. However, a performance gap may occur when going from simulation to real-life. A logical next step is a clinical trial of our dosing schedule recommendation system with physician oversight, with a control group following a dosing schedule designed by physician’s standard practice in real-life patients. 


\section{AI use statement}
LLM-based coding assistants were used for the development and review of the code, and for deriving the intrinsic-dimension-dependent information gain. Additionally, commercially available LLMs were used for manuscript proofreading, for refining the design of the synthetic experiments, for complementing our literature review, and for providing feedback on the manuscript. LLMs were used for drafting some sections of the appendix, before careful editing by the authors.  
The authors reviewed all AI-assisted work and take full responsibility for the final content of this work.

\section{Ethics Statement}

This work focuses on leveraging differentiable mechanistic simulators for more efficient global optimization of black-box objective functions. In particular, it presents an application of the proposed method for personalizing medication dosing, with the potential benefit of reducing the burden and duration of empirical dose titration. 
The present study uses only simulated patients generated from previously published aggregate PK/PD models. It involves no human participants, identifiable patient information, or individual-level clinical records, and therefore did not require informed consent or institutional review board approval.  

The proposed method is evaluated exclusively in silico and is not validated for clinical use. Errors in the mechanistic model, noisy or biased patient feedback, distribution shifts, and optimization outside the model’s domain of validity could all lead to inappropriate recommendations, potentially resulting in ineffective treatment or adverse effects.

In particular, the population PK/PD models used in this work were estimated from particular study cohorts and may not represent all patient groups equally. The safety constraints considered in this study are not comprehensive, and should therefore not be interpreted as clinical safety guarantees.

Any prospective use would require independent  evaluation on representative patient populations, approval by the relevant ethics and regulatory bodies, and a carefully controlled clinical study with informed consent. Recommendations should remain subject to qualified clinician review rather than being administered autonomously.

\section{Reproducibility statement}

All empirical results are based on synthetic experiments or simulated patients. No proprietary data, individual-level clinical records, or human participants are used. Virtual patients are generated from the published aggregate pharmacokinetic and pharmacodynamic models cited in the paper. The manuscript specifies the model equations and parameter distributions, BOASS algorithm, neural-network architecture and training procedure, data-generation and pruning rules, search-space specification, numerical ODE solver and tolerances, acquisition-function optimization, Gaussian process fitting, baselines, evaluation protocols, sample sizes, and uncertainty summaries. These details are provided in Sections 3–5 and Appendices A and C–H. The implementation uses PyTorch, JAX, diffrax, BoTorch, and GPyTorch. 

We have documented in details the instantiation of the Levodopa-Carbidopa use-case in Appendix \ref{app:algorithm}-\ref{app:dosingnet}-\ref{app:training}-\ref{app:bo}-\ref{app:pkpd_levodopa}-\ref{app:pibo} — architecture, training procedure, solver settings, search space, baselines, and evaluation protocol, and for the synthetic experiments in Appendix \ref{app:synthetic_experiments}. The code to reproduce all experiments is submitted with the manuscript and will be released in a public repository upon acceptance, specifying the exact package versions and random seeds.

\bibliography{bibtex_references}
\bibliographystyle{plainnat}

\clearpage
\begin{appendices}

\setcounter{figure}{0}
\renewcommand{\thefigure}{S\arabic{figure}}

\setcounter{table}{0}
\renewcommand{\thetable}{S\arabic{table}}

\startcontents[appendices] 
\printcontents[appendices]{}{1}{}
 
\section{Method summary}
\label{app:algorithm}
Algorithm~\ref{alg:method} summarizes the two stages of the BOASS method in the precision drug dosing context. Offline, the dosing network $\mathcal{E}_w$ is
trained against the PK/PD model (Section~\ref{sec:offline}). Online, Bayesian optimization searches over the network inputs $\phi$, using
$\mathcal{E}_w$ to turn each candidate into a regimen administered to the patient (Section~\ref{sec:online}).

\begin{algorithm}[htbp]
\caption{Bayesian optimization on amortized search-space for precision dosing}
\label{alg:method}
\begin{algorithmic}[1]
\REQUIRE population distribution $p(\phi)$; differentiable simulator $(f_\text{PK},h,f_\text{PD})$; target effect
$r^\star$; training size $N$; step size $\eta$; GP kernel $k$; acquisition function $\alpha$; online budget
$T_\text{BO}$
\STATE \textbf{\emph{Offline: train the dosing network}}
\STATE sample virtual patients $\{\phi_i\}_{i=1}^N \sim p(\phi)$
\FOR{$i=1,\dots,N$}
    \STATE $c^\star_{\phi_i} \gets \min\{c>0 : f_\text{PD}(c;\phi_i)=r^\star\}$ \COMMENT{target concentration}
\ENDFOR
\STATE initialize network weights $w$
\REPEAT
    \STATE sample a minibatch $B \subseteq \{1,\dots,N\}$
    \FOR{$i \in B$}
        \STATE $(\mathbf{d},\mathbf{t}) \gets \mathcal{E}_w(\phi_i)$; integrate the PK/PD
ODEs~\eqref{eq:pkpd-ode-piecewise} and evaluate the loss $\mathcal{L}(w;\phi_i)$~\eqref{eq:ddn-loss}
    \ENDFOR
    \STATE $w \gets w - \eta\,\nabla_w \tfrac{1}{|B|}\sum_{i\in B}\mathcal{L}(w;\phi_i)$ \COMMENT{backpropagate through
the ODE solver}
\UNTIL{convergence}
\STATE the trained network defines the search space
$\mathcal{M}=\{\mathcal{E}_w(\phi):\phi\in\Phi\}$
\STATE \textbf{\emph{Online: personalize the regimen for a patient}}
\STATE initialize the history $\mathcal{H} \gets \emptyset$ \COMMENT{optionally seed with the population mean
$\bar\phi$}
\FOR{$t=0,1,\dots,T_\text{BO}-1$}
    \STATE fit the GP surrogate with kernel $k$ to $\mathcal{H}$
    \STATE $\phi_{t+1} \gets \argmax_{\phi\in\Phi}\alpha(\phi;\mathcal{H})$ \COMMENT{acquisition step}
    \STATE administer the regimen $(\mathbf{d}_{t+1},\mathbf{t}_{t+1})=\mathcal{E}_w(\phi_{t+1})$ to the patient
    \STATE observe feedback $y_{t+1}$ \COMMENT{symptom score, wearable-derived signal, preference, \dots}
    \STATE $\mathcal{H} \gets \mathcal{H}\cup\{(\phi_{t+1},y_{t+1})\}$
\ENDFOR
\ENSURE recommended regimen $\mathcal{E}_w(\phi^\star)$, with $\phi^\star$ the best candidate in $\mathcal{H}$
\end{algorithmic}
\end{algorithm}

\begin{figure}[htbp]
    \centering
    \includegraphics[width=1\linewidth]{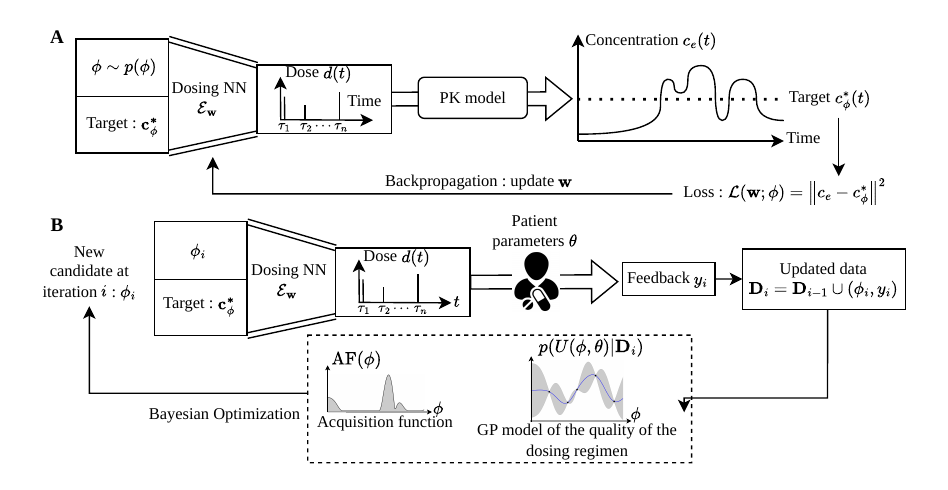}
        \caption{Illustration of the BOASS method in an application to individualized drug dosing. \textbf{(A)} The dosing network $\mathcal{E}_w$ is trained
offline against the PK/PD model; \textbf{(B)} Online,
BO searches over patient parameters $\phi$.}
    \label{fig:overview}
\end{figure}

The method is illustrated in the context of individualized patient-in-the-loop drug dosing in Figure \ref{fig:overview}.

\section{Regret bound for Bayesian optimization on Amortized Search Space}
\label{app:theory}
\subsection{Regret decomposition}

Restricting Bayesian optimization to the image of the simulator-inverting network $\mathcal{E}_w:\Phi\to\mathcal{D}$ reduces the set of candidate parameters that must be explored, but may exclude the parameters that maximize the true utility. We make this trade-off explicit below.

Let $\mathbf{x} \in\mathcal{D}$ denote the system's parameters being optimized, and let $U(\mathbf{x})$ be the unknown patient-specific utility. Let $\mathcal{M}=\{\mathcal{E}_w(\phi):\phi\in\Phi\}\subseteq\mathcal{D}$ denote the search space induced by the simulator inversion neural network. This is the set over which the online BO stage operates. 
Let $\mathbf{x}^\star\in\argmax_{\mathbf{x}\in\mathcal{D}}U(\mathbf{x})$ and $\mathbf{x}^\star_{\mathcal{M}}\in\argmax_{\mathbf{x}\in\mathcal{M}}U(\mathbf{x})$ be respectively the globally optimal system parameters and the best system parameters contained in the learned search space. We define the search-space bias as :
\begin{equation}
    \Delta_{\mathcal{M}}
    = U(\mathbf{x}^\star)
    -
    U(\mathbf{x}^\star_{\mathcal{M}})
    \geq 0,
    \label{eq:search-space-bias}
\end{equation}
with equality when $\mathbf{x}^\star\in\mathcal{M}$.

Suppose that after $n$ online evaluations, BO has tested regimens $\mathbf{x}_1,\ldots,\mathbf{x}_n\in\mathcal{M}$. 
In the noise-free scenario, the reported solution is typically $\max_{i\leq n}U(\mathbf{x}_i)$. Alternatively, in practice, one may report instead a maximum of the posterior mean $\widehat{\mathbf{x}}_n  \in \argmax_{\mathbf{x}\in\mathcal{M}}
    \mu_n(\mathbf{x})$, where $\mu_n$ is the posterior mean of the GP surrogate. The corresponding regret is a distinct quantity and is not directly controlled by the standard cumulative-regret argument used below. Our theoretical results therefore concern the best evaluated point.

Its global simple regret admits the decomposition : 
\begin{equation}
\begin{aligned}
    r_n
    &=
    U(\mathbf{x}^\star)
    -
    \max_{i\leq n}U(\mathbf{x}_i)
    \\
    &=
    \underbrace{
        U(\mathbf{x}^\star)
        -
        U(\mathbf{x}^\star_{\mathcal{M}})
    }_{\text{search-space bias}}
    +
    \underbrace{
        U(\mathbf{x}^\star_{\mathcal{M}})
        -
        U(\widehat{\mathbf{x}}_n)
    }_{\text{optimization error on }\mathcal{M}}
    \\
    &=
    \Delta_{\mathcal{M}}
    +
    r_n^{\mathcal{M}}.
\end{aligned}
\label{eq:regret-decomposition}
\end{equation}
 
Equation~\eqref{eq:regret-decomposition} captures the central motivation for the method: amortized search-space restriction may introduce an irreducible bias $\Delta_{\mathcal{M}}$, but it can simultaneously make the remaining optimization problem substantially easier.

\subsection{Reduced information gain}

\citet{srinivas_gaussian_2010}  established regret bounds for the GP-UCB acquisition function with iid Gaussian noise, and showed that regret is  controlled by the maximum information gain. 
For a domain $A$ and kernel $k$, the maximum information gain is defined as:
\begin{equation}
    \gamma_n(A;k) =    \max_{\mathbf{x}_{1:n}\in A^n}
    I(y_{1:n};U_{1:n}),
    \label{eq:max-information-gain}
\end{equation}
where $U_{1:n}=(U(\mathbf{x}_1),\ldots,U(\mathbf{x}_n))$, and $y_{1:n}$ denotes noisy observations. Here, mutual information is computed under a GP prior with covariance kernel $k$ and independent Gaussian observation noise.

\begin{lemma}[Pullback kernel and information gain]
\label{rev:prop:pullback}
Let $\mathcal{E}_w:\Phi\to\mathcal D$ be deterministic, $\mathcal{M}=\mathcal{E}_w(\Phi)$, and $k_{\mathcal D}$ be a bounded positive-semidefinite kernel on $\mathcal D$. Then
\begin{equation}
  k_{\mathrm{pb}}(\phi,\phi')
  = k_{\mathcal D}\bigl(\mathcal{E}_w(\phi),\mathcal{E}_w(\phi')\bigr)
  \label{rev:eq:pullback}
\end{equation}
is a positive-semidefinite kernel. For a common independent Gaussian observation-noise variance $\sigma_\epsilon^2>0$, the maximum information gain, defined by the supremum over $n$-tuples with repetitions allowed,
satisfies
\begin{equation}
  \gamma_n(\Phi;k_{\mathrm{pb}})
  =\gamma_n(\mathcal{M};k_{\mathcal D})
  \leq\gamma_n(\mathcal D;k_{\mathcal D}).
  \label{rev:eq:information-identity}
\end{equation}
 
\end{lemma}
\begin{proof} 
For any $\phi_1,\ldots,\phi_n\in\Phi$ and $c\in\mathbb R^n$,
\begin{equation}
  \sum_{i,j=1}^n c_i c_j k_{\mathrm{pb}}(\phi_i,\phi_j)
  =\sum_{i,j=1}^n c_i c_j
    k_{\mathcal D}\bigl(E_w(\phi_i),E_w(\phi_j)\bigr)
  \geq0.
\end{equation}
Thus $k_{\mathrm{pb}}$ is positive semidefinite.

Since $\mathcal{M}=\mathcal{E}_w(\Phi)$ and $k_{\mathrm{pb}}(\phi,\phi')
= k_{\mathcal{D}}(\mathcal{E}_w(\phi),\mathcal{E}_w(\phi'))$, the two domains generate exactly the same set of kernel Gram matrices.
Hence,
\begin{equation}
    \gamma_n(\Phi;k_{\mathrm{pb}})
    =
    \gamma_n(\mathcal{M};k_{\mathcal{D}}).
    \label{eq:pullback-information-gain}
\end{equation}
Because $\mathcal{M}\subseteq\mathcal{D}$, the set of admissible $n$-point designs in $\mathcal{M}$ is a subset of those in $\mathcal{D}$. Therefore,
\begin{equation}
    \gamma_n(\mathcal{M};k_{\mathcal{D}})
    \leq
    \gamma_n(\mathcal{D};k_{\mathcal{D}}).
    \label{eq:information-gain-subset}
\end{equation}
\end{proof}

\citet{srinivas_gaussian_2010}  showed that for a general compact and convex domain, for $f\sim GP(0, k(\mathbf{x},\mathbf{x'}))$ (under some derivative-tail condition) : $R_n= \widetilde{O}(\sqrt{n\beta_n\gamma_n})$. 

$\mathcal{M}$ is not guaranteed to be convex. Therefore, we instead define the objective in the simulator-induced coordinates as $g(\phi)=U(\mathcal{E}_w(\phi))$ and use the pullback kernel $k_{\mathrm{pb}}(\phi,\phi') =k_{\mathcal{D}}(\mathcal{E}_w(\phi),\mathcal{E}_w(\phi'))$. 
Throughout this analysis, the decoder and kernel hyperparameters are fixed. We assume that $g\sim\mathcal{GP}(0,k_{\mathrm{pb}})$.
We assume that $\Phi\subset[0,r]^p$ is compact and convex, where $p=\dim(\Phi)$, and
that observations are of the form $y_t=g(\phi_t)+\epsilon_t$, with independent Gaussian noise $\epsilon_t\sim\mathcal{N}(0,\sigma^2)$.


We further assume that the pullback GP satisfies the derivative-tail condition of \citet{srinivas_gaussian_2010}: for some constants $a,b>0$, and all $L>0$,
\begin{equation}
    \Pr\left(
        \sup_{\phi\in\Phi}
        \left|
        \frac{\partial g(\phi)}{\partial \phi_j}
        \right| > L
    \right)
    \leq
    a\exp\left(-\left(\frac{L}{b}\right)^2\right),
    \qquad j=1,\ldots,p.
    \label{eq:derivative-tail}
\end{equation}

Under these assumptions, Theorem~2 of \citet{srinivas_gaussian_2010} implies that, for an appropriate exploration sequence $\beta_n$, GP-UCB gives a regret bound of $\widetilde{O} \left(
        \sqrt{
            n\beta_n
            \gamma_n(\Phi;k_{\mathrm{pb}})
        }
    \right)$ (here and below, the $\widetilde{O}$ notation is a variant of $O$, where log factors are suppressed).
 
Therefore, using Lemma \ref{rev:prop:pullback},  cumulative regret satisfies :
\begin{equation}
    R_n^{\mathcal{M}}
    =
   \widetilde{O} \left(
        \sqrt{
            n\beta_n
            \gamma_n(\mathcal{M};k_{\mathcal{D}})
        }
    \right),
    \label{eq:cumulative-regret}
\end{equation}
and the simple regret of the best evaluated point satisfies : 
\begin{equation}
    r_n^{\mathcal{M}}
    \leq
    \frac{R_n^{\mathcal{M}}}{n}
    =
    \widetilde{O}\left(
        \sqrt{
            \frac{
                \beta_n
                \gamma_n(\mathcal{M};k_{\mathcal{D}})
            }{n}
        }
    \right).
    \label{eq:restricted-simple-regret}
\end{equation}
Combining Equations~\eqref{eq:regret-decomposition} and~\eqref{eq:restricted-simple-regret} yields : 
\begin{equation}
    r_n
    \leq
    \Delta_{\mathcal{M}}
    +
    \widetilde{O}\left(
        \sqrt{
            \frac{
                \beta_n
                \gamma_n(\mathcal{M};k_{\mathcal{D}})
            }{n}
        }
    \right).
    \label{eq:bias-complexity-bound}
\end{equation}

Equation~\eqref{eq:bias-complexity-bound} makes the bias-complexity trade-off explicit: restricting the search can reduce the information-gain term, at the cost of introducing search-space bias. 

Note that when BO is run over $\Phi$ with the pullback kernel, its statistical complexity is that of the simulator-induced set of system parameters  rather than that of the raw parameterization $\Phi$. 
This also explains why the pullback kernel is insensitive to redundant coordinates: distinct $\phi,\phi'$ satisfying $\mathcal{E}_w(\phi)=\mathcal{E}_w(\phi')$ are identical from the perspective of the GP.

\subsection{Error introduced by amortization}
\label{app:amortization_error}

The previous argument treats $\mathcal{M}$ as given. In practice, it is itself learned by amortizing the simulator-defined optimization problem in Equation~3.

\begin{lemma}[Simulator and amortization bias]
Let $\mathcal{E}^\star(\phi)$ denote an exact solution of the simulator optimization problem, and define the corresponding ideal simulator-induced search space as $\mathcal{M}^\star=\{\mathcal{E}^\star(\phi):\phi\in\Phi\}$. The dosing network $\mathcal{E}_w$ approximates $\mathcal{E}^\star$, producing $\mathcal{M}=\mathcal{E}_w(\Phi)$.  
Suppose that $U$ is $L_U$-Lipschitz with respect to a regimen distance $\delta$, so that $|U(\mathbf{x})-U(\mathbf{x}')|\leq L_U\delta(\mathbf{x},\mathbf{x}')$, and that for some $\phi^\star_{\mathcal{M}^\star}\in\argmax_{\phi\in\Phi}
U(\mathcal{E}^\star(\phi))$,
\begin{equation}
    \delta\!\left(
        \mathcal{E}_w(\phi^\star_{\mathcal{M}^\star}),
        \mathcal{E}^\star(\phi^\star_{\mathcal{M}^\star})
    \right)
    \leq \varepsilon_{\mathrm{am}}.
    \label{eq:amortization-error}
\end{equation}
 With the error due solely to simulator misspecification $\Delta_{\mathrm{model}} =
    U(\mathbf{x}^\star)  - \max_{\mathbf{x}\in\mathcal{M}^\star}   U(\mathbf{x})$,  
\begin{equation}
    \Delta_{\mathcal{M}}
    \leq
    \Delta_{\mathrm{model}} + L_U\varepsilon_{\mathrm{am}}.
    \label{eq:search-space-bias-bound}
\end{equation}
\end{lemma}
\begin{proof} 
 Indeed, Lipschitz continuity implies : 
\begin{equation}
\begin{aligned}
L_U\varepsilon_{\mathrm{am}} &\geq L_U\delta(\mathcal{E}^\star(\phi^\star_{\mathcal{M}^\star}),\mathcal{E}_w(\phi^\star_{\mathcal{M}^\star})) \\
&\geq
|U\left(\mathcal{E}^\star(\phi^\star_{\mathcal{M}^\star})\right) - U\left(\mathcal{E}_w(\phi^\star_{\mathcal{M}^\star})\right)| \\
&\geq  U\left(\mathcal{E}^\star(\phi^\star_{\mathcal{M}^\star})\right) - U\left(\mathcal{E}_w(\phi^\star_{\mathcal{M}^\star})\right)  
\end{aligned}
\end{equation}
By definition of $\phi^\star_{\mathcal{M}^\star}$, $ U\left(\mathcal{E}^\star(\phi^\star_{\mathcal{M}^\star})\right) = \max_{\phi\in\Phi} U(\mathcal{E}^\star(\phi)) $.  Moreover, $\max_{\phi\in\Phi} U(\mathcal{E}^\star(\phi)) = \max_{\mathbf{x}\in\mathcal{M}^\star}U(\mathbf{x}) $.  Therefore :
\begin{equation}
\begin{aligned}
L_U\varepsilon_{\mathrm{am}} 
&\geq  \max_{\mathbf{x}\in\mathcal{M}^\star}U(\mathbf{x})  - U\left(\mathcal{E}_w(\phi^\star_{\mathcal{M}^\star})\right)
\end{aligned}
\end{equation}

Since $\mathcal{E}_w(\phi^\star_{\mathcal{M}^\star})\in\mathcal{M}$ and $\mathbf{x}^\star_{\mathcal{M}}$ maximizes $U$ over $\mathcal{M}$,
\[
    U(\mathbf{x}^\star_{\mathcal{M}})
    \geq U\!\left(\mathcal{E}_w(\phi^\star_{\mathcal{M}^\star})\right).
\]
Therefore
\begin{align*}
\Delta_{\mathcal{M}}
    &= U(\mathbf{x}^\star)-U(\mathbf{x}^\star_{\mathcal{M}}) \\
    &= \Delta_{\mathrm{model}}
       + \max_{\mathbf{x}\in\mathcal{M}^\star} U(\mathbf{x})
       - U(\mathbf{x}^\star_{\mathcal{M}}) \\
    &\leq \Delta_{\mathrm{model}}
       + \max_{\mathbf{x}\in\mathcal{M}^\star} U(\mathbf{x})
       - U\!\left(\mathcal{E}_w(\phi^\star_{\mathcal{M}^\star})\right) \\
    &\leq \Delta_{\mathrm{model}}+L_U\varepsilon_{\mathrm{am}},
\end{align*}
which proves Equation~\eqref{eq:search-space-bias-bound}.
\end{proof}

Combining Equations~\eqref{eq:bias-complexity-bound} and~\eqref{eq:search-space-bias-bound} gives the overall bound
\begin{equation}
\begin{aligned}
    r_n
    \leq\;&
    \underbrace{
        \Delta_{\mathrm{model}}
    }_{\substack{\text{simulator}\\\text{misspecification}}}
    +
    \underbrace{
        L_U\varepsilon_{\mathrm{am}}
    }_{\substack{\text{amortization}\\\text{error}}}
    +
    \underbrace{
        \widetilde{O}\left(
            \sqrt{
                \frac{
                    \beta_n
                    \gamma_n(\mathcal{M};k_{\mathcal{D}})
                }{n}
            }
        \right)
    }_{\substack{\text{online}\\\text{optimization error}}}.
\end{aligned}
\label{eq:overall-regret-bound}
\end{equation} 
Equation~\eqref{eq:overall-regret-bound} separates the three sources of error in our method. A more accurate mechanistic model reduces $\Delta_{\mathrm{model}}$, a better-trained dosing network reduces $\varepsilon_{\mathrm{am}}$, and a more structured, lower-complexity search space reduces the number of online evaluations required by Bayesian optimization.

\subsection{Intrinsic-dimension-dependent information gain}
\label{app:intrinsic-information-gain}
 
The set-inclusion argument in Equation~\eqref{eq:information-gain-subset} guarantees that restricting the search to $\mathcal{M}$ cannot increase the maximum information gain. A sharper characterization can be obtained from the spectral decay of the kernel on the restricted search space.

Let $\{\lambda_j\}_{j\geq 1}$ denote the Mercer eigenvalues of $k_{\mathcal{D}}$ restricted to $\mathcal{M}$ with respect to the  Riemannian volume measure on $\mathcal{M}$, ordered such that $\lambda_1\geq\lambda_2\geq\cdots$.

Assume additionally that the Mercer eigenfunctions satisfy the boundedness condition required by \citet{vakili_information_2021}.
If, for some $\beta>1$, the eigenvalues satisfy the polynomial decay $\lambda_j\leq Cj^{-\beta}$, then
\begin{equation}
    \gamma_n(\mathcal{M};k_{\mathcal{D}})
    =
    O\left(
        n^{1/\beta}
        (\log n)^{1-1/\beta}
    \right).
    \label{eq:information-gain-eigendecay}
\end{equation}
Thus, the information-gain rate is controlled by the spectral complexity of the kernel on the simulator-induced search space. In particular, if restriction to $\mathcal{M}$ induces faster eigenvalue decay than on $\mathcal{D}$, Equation~\eqref{eq:information-gain-eigendecay} gives a sharper rate.

A useful sufficient condition for such faster decay can be obtained for Matérn kernels when the simulator-induced search space has additional geometric regularity. 
\citet{vakili_information_2021} use the fact that for a Matérn kernel with smoothness parameter $\nu > \frac{1}{2}$, $\lambda_m = O\left(m^{-\frac{2\nu+D}{D}}\right)$, and for the Squared Exponential kernel, $\lambda_m = O\left(\exp(-m^{\frac{1}{D}})\right)$. Substituting into Equation~\eqref{eq:information-gain-eigendecay}  leads to bounds on the maximum information gain  $O\left(
        n^{\frac{D}{2\nu+D}}
        (\log n)^{\frac{2\nu}{2\nu+D}}
    \right)$ and $O\left( \log^{D+1}(n) \right)$, respectively.

Now, suppose that $\mathcal{M}\subset\mathcal{D}\subset\mathbb R^D$ is a compact, smooth $\rho$-dimensional embedded submanifold without boundary, and that $k_{\mathcal{D}}$ is a Matérn kernel with smoothness parameter $\nu$. 
Proposition~27 of \citet{rosa_posterior_2023} establishes that the RKHS $\mathcal{H}_k$ of the restricted extrinsic Matérn kernel is norm-equivalent to the Sobolev space $H^{\nu+\rho/2}(\mathcal{M})$. The proof of Corollary~1.2 of \citet{rosa_$l^2$posterior_2025} then implies that the eigenvalues satisfy:
\begin{equation}
    \lambda_j
    \asymp
    j^{-(1+2\nu/\rho)}
    =
    j^{-(2\nu+\rho)/\rho}.
\end{equation}
 
In particular, $\lambda_j=O(j^{-(2\nu+\rho)/\rho})$.

Combined with the information-gain results of \citet{vakili_information_2021}, this gives an information-gain rate depending on \(\rho\) rather than on the dimension \(D\) of the full design space. Indeed, substituting $\beta=(2\nu+\rho)/\rho$ into Equation~\eqref{eq:information-gain-eigendecay} gives
\begin{equation}
    \gamma_n(\mathcal{M};k_{\mathcal{D}})
    =
    O\left(
        n^{\frac{\rho}{2\nu+\rho}}
        (\log n)^{\frac{2\nu}{2\nu+\rho}}
    \right).
    \label{eq:intrinsic-information-gain}
\end{equation}

Under the GP-UCB assumptions stated above, this results in $ r_n^{\mathcal{M}} = \widetilde O\left(n^{-\frac{\nu}{2\nu+\rho}} \right)$ and $r_n^{\mathcal{D}}
=\widetilde O(n^{-\nu/(2\nu+D)})$.  
Since $\rho<D$, $ \frac{\rho}{2\nu+\rho}  < \frac{D}{2\nu+D}$, restricting optimization to a $\rho$-dimensional simulator-induced search space improves the polynomial dependence of the maximum information gain on the evaluation budget $n$, and therefore improves the simple regret exponent, which is more favorable on $\mathcal{M}$. 

 Finally, combining the optimization rate with the simulator and
amortization errors gives
\begin{equation}
\begin{aligned}
    r_n
    \leq\;&
    \underbrace{
        \Delta_{\mathrm{model}}
    }_{\text{simulator misspecification}}
    +
    \underbrace{
        L_U\varepsilon_{\mathrm{am}}
    }_{\text{amortization error}}
    +
    \underbrace{
        \widetilde O
        \left(
            n^{-\frac{\nu}{2\nu+\rho}}
        \right)
    }_{\text{online optimization error}}.
\end{aligned}
\label{eq:overall-intrinsic-regret}
\end{equation}

Equation~\eqref{eq:overall-intrinsic-regret} makes explicit the trade-off induced by restricting the search space. A more accurate mechanistic model reduces $\Delta_{\mathrm{model}}$, better amortization reduces $\varepsilon_{\mathrm{am}}$, while a lower-dimensional simulator-induced solution set improves the rate at which the remaining online optimization error decreases.

Moreover, it follows immediately from \ref{eq:regret-decomposition} that
\begin{equation} 
     \Delta_\mathcal{M}\leq\ r_n .
\end{equation}

The Mercer-eigenvalue information-gain result of \cite{vakili_information_2021} has been widely used. However, it  assumes uniformly bounded eigenfunctions. \citet{janz_maximum_2026} recently noted that this assumption has not been verified even for the Matérn or squared-exponential kernels and recognize it as a genuine gap in the information-gain literature. Our result is conditional on this eigenfunction bound and on the geometric and GP assumptions of Proposition \ref{prop:online-regret}, which are not established for our PK/PD experiments.

\subsection{Pullback kernel vs kernel defined in the simulator parameters space}
\label{sec:pullback-vs-standard-kernel}

The pullback kernel defines similarity between two search coordinates $\phi,\phi'\in\Phi$ through the regimens they produce, $ k_{\mathrm{pb}}(\phi,\phi')
    =
    k_{\mathcal D}
    \left(
        \mathcal E_w(\phi),
        \mathcal E_w(\phi')
    \right)$.
For an isotropic Matérn kernel $k_{\mathcal D}$, similarity is therefore determined by the regimen-space distance $\|\mathcal E_w(\phi)-\mathcal E_w(\phi')\|$.

For comparison, suppose that BO instead uses a Matérn-$\nu$ kernel
$k_\Phi$ directly on the search coordinates $\Phi$, which measures similarity using Euclidean distance $\|\phi-\phi'\|$.

Under the corresponding Matérn information-gain assumptions,
\begin{equation}
    \gamma_n(\Phi;k_\Phi)
    =
    \widetilde O
    \left(
        n^{\frac{p}{2\nu+p}}
    \right), 
\end{equation}
with $p = \dim(\Phi)$.

In general, $p = \dim(\Phi)\geq\dim(\widehat{\mathcal M})=\rho$, so the pullback kernel can have a more favorable polynomial information gain rate. 

Note that if $\mathcal E_w:\Phi\to\widehat{\mathcal M}$ is an embedding, $\dim(\Phi)=\dim(\widehat{\mathcal M})=\rho$.  In general, the two kernels nevertheless encode different geometries. 

The pullback kernel  inherits the geometry induced by the map $\mathcal E_w$, and therefore accounts for directions in $\Phi$ that have different effects on the produced regimen. 
In particular, two points may be far away in the simulator parameter space,  $ \|\phi - \phi' \| \gg 0$, but produce almost identical regimens, $\|\mathcal{E}_w(\phi) - \mathcal{E}_w(\phi')\| \approx 0 $. While a Matérn kernel defined on $\Phi$ may regard these as nearly independent, the pullback kernel captures their similarity : $k_\text{pb}(\phi, \phi') \approx k_{\mathcal{D}}(x,x)$, where $x = \mathcal{E}_w(\phi)$.

This distinction is consistent with recent results showing that manifold geometry can affect GP-bandit regret beyond the dimension-dependent rate.
\citet{dorn_manifoldaware_2026} derive regret bounds for Matérn GP bandits on compact Riemannian manifolds in which geometric quantities such as the Riemannian volume, injectivity radius, and curvature explicitly enter the bound \citep{dorn_manifoldaware_2026}. Their setting differs from ours as they focused on intrinsic Matérn kernels, but it supports the observation that two kernels with the same
dimensional asymptotic exponent may nevertheless exhibit different regret due the geometry encoded by the kernel.

\subsection{Consequences}

The proof of Proposition \ref{prop:regret-decomposition}  shows that the amortization term depends on the regimen distance between the learned and exact simulator-optimal designs at \(\phi^\star_{\mathcal M^\star}\),  $\delta\!\left(\mathcal{E}_w(\phi^\star_{\mathcal{M}^\star}),         \mathcal{E} \star(\phi^\star_{\mathcal{M}^\star})  \right)$ ( \(\phi^\star_{\mathcal M^\star}\) is the simulator parameter whose exact simulator-optimal design achieves the highest true utility over the ideal search space). 
 
This suggests fitting the inverse map directly to cached approximate simulator-optimal designs \(\widetilde{\mathcal E}(\phi_i)\) by minimizing $\mathcal L(w)=\frac{1}{N}\sum_{i=1}^{N}
\delta\!\left(\mathcal E_w(\phi_i),\widetilde{\mathcal E}(\phi_i)\right)$.
This has an important implication: the training pipeline can therefore use a nondifferentiable simulator if the targets \(\widetilde{\mathcal E}(\phi_i)\) are generated with a derivative-free solver.
The regret bound depends on how well the network matches the target and how accurate that target is at \(\phi^\star_{\mathcal M^\star}\). 
Appendix K.1.2 already uses direct regression against the ideal map in the synthetic experiment.  

\section{Challenges in Levodopa treatment for Parkinson's disease}
\label{app:parkinson}

Four aspects make the dosing especially challenging in Parkinson's disease. First, the response to a regimen is highly patient-specific due to inter-individual physiological differences influencing drug absorption, metabolism, and distribution \citep{hammarlund-udenaes_activesite_2010, delange_understanding_2022, chan_importance_2005, titova_personalized_2017}, and drug response \citep{jankovic_parkinsons_2020}.
Second, the optimization objective is only weakly observed: relevant real-time biomarkers of treatment response are rarely available, and clinicians often rely instead on patient- and clinician-reported outcome measures such as symptom diaries \citep{hauser_parkinsons_2004, hauser_prevalence_2019}, which are subjective, sparse, and noisy. Third, even if a mechanistic simulator of the drug is often available in the form of a pharmacokinetic/pharmacodynamic (PK/PD) model, it predicts a quantity, drug concentration in the effective body compartment or a pharmacological effect, that cannot be readily measured in routine care. Fourth, the optimization problem is typically moderate to high dimensional, with potentially complex dosing schedules. 

\section{Dosing network architecture and training details for Levodopa use-case}
\label{app:dosingnet}

\paragraph{Architecture.} The network takes as input the PK/PD parameters $\phi$ of
Appendix~\ref{app:pkpd_levodopa} together with the target concentration $c^\star_\phi$, itself a function of the PD parameters. All inputs are standardized with the mean and standard deviation of the sampled population. It is a feedforward network with $6$ fully connected layers of width $256$ and ReLU activations, followed by a linear layer returning $z\in\mathbb{R}^{2M}$.

\paragraph{Output layer.} The vector $z$ is turned into a regimen by the deterministic layer
\begin{equation}
\label{eq:output-layer}
d_i \;=\; d_{\max}\,\frac{1+\tanh(z_i)}{2}, \qquad
(\eta_1,\dots,\eta_M) \;=\; W\,\softmax(z_{M+1},\dots,z_{2M}),
\end{equation}
for $i=1,\dots,M$, where $W = (t_f-t_0)-(M-1)\Delta$ is the free time to allocate between intakes. The administration times follow as $t_1 = t_0$ and $t_i = t_{i-1} + \Delta + \eta_{i-1}$ for $i=2,\dots,M$.
The $\tanh$ keeps each dose in $[0,d_{\max}]$, and the softmax makes the lags positive and constrains them to sum to
$W$, so that consecutive intakes are at least $\Delta$ apart and all of them fall within $[t_0,t_f]$. The last lag $\eta_M$ enters no administration time: it is the slack between the last intake and the end of the window, and it lets the network leave part of the window unused. Layer \eqref{eq:output-layer} is the decoding map \eqref{eq:decode} evaluated at $u_i=(1+\tanh(z_i))/2$ and $v_i=\exp(z_{M+i})$, up to the positive rescaling of $\mathbf{v}$ that leaves
\eqref{eq:decode} unchanged. The image $\mathcal{M}=\mathcal{E}_w(\Phi)$ is therefore contained in $\mathcal{D}$ by
construction.

\paragraph{Smooth dose discretization.} Microtablets come in multiples of $\Delta_\text{tab}=5$~mg, so the admissible
doses are the $K=d_{\max}/\Delta_\text{tab}+1=51$ values $\{0,\Delta_\text{tab},\dots,d_{\max}\}$. Rounding $d_i$ to the
nearest of them is piecewise constant, hence has zero gradient almost everywhere and cannot be trained through. During training we replace it by the smooth surrogate
\begin{equation}
\label{eq:smooth-round}
\rho_\kappa(d) \;=\; \frac{\Delta_\text{tab}}{2}\sum_{k=0}^{K-2}\left[1+\tanh\left(\kappa\left(d-k\,\Delta_\text{tab}-\frac{\Delta_\text{tab}}{2}\right)\right)\right],
\end{equation}
a sum of $K-1$ smooth steps of height $\Delta_\text{tab}$, each centred at the midpoint between two consecutive
admissible doses. The steepness $\kappa$ interpolates between the two regimes of interest: for
$\kappa\,\Delta_\text{tab}\ll1$ the map is close to the identity away from the endpoints of the dose range and leaves
the continuous doses essentially untouched; as $\kappa\to\infty$, it converges to exact rounding, with vanishing
gradients. We anneal it during training: discretization is disabled for the first half of the epochs, then $\kappa$
increases linearly from $0.1$ to $1$ over the remaining ones. The validation loss and the regimens actually
administered always use exact rounding.

We note that although in our application, the dosing resolution is 5mg, this is much smaller than the typical range of dose variation so that we could still use a GP defined on a continuous dosing space.

\paragraph{ODE integration.} The PK/PD system is integrated with the Tsit5 solver of \textit{diffrax}
\citep{kidger_diffrax_2022}, restarting at each intake with the updated initial conditions
of~\eqref{eq:pkpd-ode-piecewise}, using an adaptive step size (relative tolerance $10^{-7}$, absolute tolerance
$10^{-9}$) and $200$ saved points per inter-intake segment. These points form the grid of the trapezoidal rule
in~\eqref{eq:ddn-loss}. The horizon $[0,T]$ starts $30$~min before the first intake and ends $60$~min after $t_f$, so
that the loss also covers the elimination phase following the last intake. The loss is computed on the concentration scale, as in \eqref{eq:ddn-loss}.

\paragraph{Optimization.} We train for $25$ epochs with AdamW \citep{loshchilov_decoupled_2019}, batch size $64$, learning
rate $2\cdot10^{-3}$ decayed by a factor $0.2633$ every $4$ epochs, weight decay $4.13\cdot10^{-5}$, and gradient-norm
clipping at $10$. The hidden layers use dropout with probability $0.07$. We retain the checkpoint with the lowest
validation loss. The training hyperparameters were selected with Optuna \citep{akiba_optuna_2019}.

\paragraph{Training cost.}
We trained the network on 16{,}000 synthetic patients (80\% of a simulated population of 20{,}000; the remaining 20\% were held out for
testing). Each training step simulates the predicted posology through the
PK/PD ODE and backpropagates through the solver. Training took $7985$\,s ($\approx 2.2$\,h, about $5.3$\,min per
epoch) on the CPU of a laptop (Apple M4, 10 cores, 24\,GB RAM), without a GPU. This time covers one JIT compilation of
the solver and the test-set evaluation after every epoch, but not the simulation of the synthetic population. The cost is paid once, offline, and is then shared by every patient the network doses.

\section{Training data}\label{app:training}

PK/PD studies typically report only the marginal of the estimated distribution of each parameter, not their joint. Sampling the parameters independently from these marginals can produce combinations that are physiologically implausible, which inflates $\mathcal{M}$ relative to what the true joint distribution would give. When possible, it is therefore advisable to prune implausible parameter samples from the training set using clinical knowledge.
A simple approach that we found effective in the dosing context is to discard the samples whose peak
effect-compartment concentration is too high under a low dose $d_l$, or too low under a high dose $d_h$, relative to
the target concentration, that is
\begin{equation}
\label{eq:pruning} 
\text{training set} = \{\phi \sim p(\phi) : C_{\text{max}}(d_{l};\phi) < c^\star_\phi < C_{\text{max}}(d_{h};\phi) \}  
\end{equation}
with $C_{\text{max}}(d;\phi) = \max_{t\in[0,\tau^*]}c_e(t;\phi \mid d)$, where the values $d_l$, $d_h$ and $\tau^*$ are chosen based on the clinical context.

For the dosing network, we sample the training data and prune the parameters as in \eqref{eq:pruning}, with $\tau^*=120\,\text{min}$, $d_l=80\,\text{mg}$ and $d_h=600\,\text{mg}$, values chosen for the clinical context of Levodopa treatment in advanced Parkinson's patients.

\section{Amortization analysis for Levodopa-Carbidopa dosing use-case}
\label{app:amortization_results}

We compare the concentration-tracking error of the regimen obtained with a forward pass of the dosing network against a direct, per-patient optimization of the same concentration error using L-BFGS-B (4 restarts: 1 warm-start at the dosing network forward pass, 3 random Sobol starts), evaluated on 500 held-out virtual patients. For the direct optimization, we report both the continuous-dose solution ("Continuous-direct") and its rounding to the nearest 5 mg increment ("Discrete-direct", $\Delta=5$ mg). Table~\ref{tab:amortization_error_timing} shows that the three approaches achieve essentially indistinguishable concentration errors. The table also reports the average time needed to obtain each solution: for the dosing network this is a single forward pass, excluding the one-time offline training cost; for direct optimization it is roughly five orders of magnitude slower, since each optimization step requires solving the ODE and backpropagating through it to compute gradients.

\begin{table}[ht]
\caption{Per-patient induced-concentration error and compute time to obtain one patient's dosage ($N=500$ patients), for the amortized, continuous-direct (per-patient optimization with L-BFGS-B), and discrete-direct solutions (per-patient optimization with L-BFGS-B rounded to the nearest 5 mg increment). Error is the trapezoid-integrated squared error between the induced effect-site concentration and the patient's target. Time is the mean wall-clock time for a single-patient call (not batched). $^\dagger$discrete-direct's doses are continuous-direct's solution rounded to the dose grid; the rounding itself is negligible and was not separately timed, so its reported time is continuous-direct's. Reported times excludes the one-off JIT compilation of the ODE solver}
\label{tab:amortization_error_timing}
\centering
\small
\begin{tabular}{lcccc}
\toprule
& \multicolumn{3}{c}{Time-normalized squared conc. error ((mg/L)$^2$)} & \multirow{2}{*}{Mean time / patient} \\
\cmidrule(lr){2-4}
Method & $q_{10}$ & Median & $q_{90}$ & \\
\midrule
Amortized (dosing net) & 0.0558 & 0.2163 & 1.2013 & 0.116\,ms \\
Continuous-direct (optimized) & 0.0543 & 0.2142 & 1.1993 & 11.715\,s \\
Discrete-direct (5\,mg grid) & 0.0544 & 0.2143 & 1.1993 & 11.715\,s$^\dagger$ \\
\bottomrule
\end{tabular}
\end{table}

\begin{figure}
    \centering
    \includegraphics[width=\linewidth]{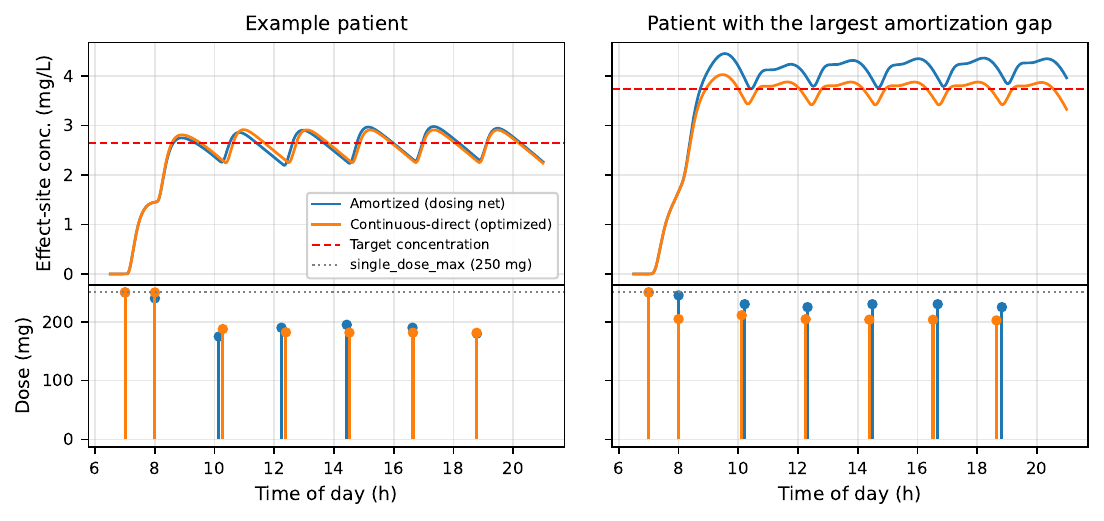}
    \caption{Two examples of optimal dosing and induced PK trajectory. On the right, we select the virtual patient with the largest discrepancy in concentration error between the amortized and the continuous-direct dosing solutions.}
    \label{fig:amortization_example}
\end{figure}

Figure~\ref{fig:amortization_example} compares the induced concentration trajectories and prescribed regimens of the amortized and continuous-direct solutions for two example patients. The left panel shows the results for a typical patient, for which the two solutions remain nearly indistinguishable. The right panel shows the worst case among the 500 held-out virtual patients: the amortized solution prescribes slightly higher maintenance doses than the numerically optimized one, pushing its induced concentration slightly further above the target. Even in this worst case, the discrepancy stays small, and the amortized regimen still offers a fast, high-quality solution to the PK/PD dosing problem.

\begin{table}[t]
\caption{Mean wall-clock time of one optimization of the acquisition function of BOASS-$k_\mathcal{D}$ with concentration feedback after $n$ observations (average over 10 patients). Amortized time is measured (median of 5 repeats per patient). $^\dagger$Continuous-direct timing is estimated as the number of distinct points the optimizer queries (1990, 2089, 1638 on average) times the mean time of one continuous-direct solve (11.7\,s, excluding the one-off JIT compilation of the ODE solver). Whole run is the time of all the acquisition optimizations of one 50-iterations BO run: the mean pooled over every patient and $N$, times the 47 days chosen by the acquisition function (the first 3 days are the shared initialization). All times exclude the GP fit: only the optimization of the acquisition function is timed.}
\label{tab:acquisition_timing}
\centering
\small
\begin{tabular}{lcccc}
\toprule
& \multicolumn{3}{c}{One optimization} & \multirow{2}{*}{Whole run} \\
\cmidrule(lr){2-4}
Method & $n=10$ & $n=20$ & $n=30$ & \\
\midrule
Amortized (dosing net) & 1.138\,s & 3.027\,s & 2.635\,s & 106.539\,s \\
Continuous-direct (estimate) & 6.48\,h$^\dagger$ & 6.80\,h$^\dagger$ & 5.33\,h$^\dagger$ & 291.49\,h$^\dagger$ \\
\bottomrule
\end{tabular}
\end{table}

The benefit of amortization is even clearer for the optimization of the acquisition function. The optimizer evaluates the acquisition function at many points of the search space, and each evaluation needs the dose regimen of the corresponding point: one forward pass of the dosing network, or one direct optimization without it.
Table~\ref{tab:acquisition_timing} reports the time of one acquisition optimization after $n = 10$, $20$ and $30$ observations. It also reports an estimate for a whole Bayesian optimization run of 50 iterations: the average time over the three snapshots, multiplied by the 47 iterations in which the acquisition function is optimized (the first three are the shared initialization). Running the direct optimization would take too long, so we only estimate its
time. We count the distinct points the optimizer queries when it uses the dosing network (1990, 2089 and 1638 on average for $n = 10$, $20$ and $30$). We then multiply these counts by the mean time of one direct optimization
(Table~\ref{tab:amortization_error_timing}). This estimate is conservative, because it leaves out further costs, such
as computing gradients through the direct optimization. Gradient-free optimizers, such as evolutionary algorithms,
usually need even more queries.
The computational time is measured on the CPU of a laptop (Apple M4, 10 cores, 24\,GB RAM), without a GPU.

Let's denote $n$ the number of BO iterations.  The compute costs for whole BO runs with $n$ iterations are:
\[
T_{\mathrm{amortized}}(n)
= T_{\mathrm{offline}} + n\,t_{\mathrm{am}},
\qquad
T_{\mathrm{direct}}(n)
= n\,t_{\mathrm{direct}}.
\]
where $T_{\mathrm{offline}}$ is the offline training cost,  $t_{\mathrm{am}}$ the typical time it takes to perform one acquisition using amortized optimization, and $t_{\mathrm{direct}}$ the typical time it takes to perform one acquisition using direct optimization. Amortization saves compute when:
\[
n >
\frac{T_{\mathrm{offline}}}
     {t_{\mathrm{direct}}-t_{\mathrm{am}}},
\qquad t_{\mathrm{direct}}>t_{\mathrm{am}}.
\]
In our PK/PD experiment, $T_{\mathrm{offline}} \approx 2.2 \text{ hours}$, so $T_{\mathrm{offline}} /({t_{\mathrm{direct}}-t_{\mathrm{am}}}) < 1$. This means that despite the offline training cost, the extremely high cost of optimizing the acquisition function when using direct optimization of the dosing regimen implies that amortization is always beneficial.

\section{Out-of-distribution patients}
\label{app:ood}
The experiments of Section~\ref{sec:results} draw the patients from the population $p(\phi)$, the same which the dosing network was trained on, and from which the search space $\Phi$ is built. 
Here we evaluate the methods on patients drawn from shifted populations, while the dosing network, the search space $\Phi$ and the $\pi$BOASS prior remain those of the nominal population.
This measures the robustness of the methods to a misspecified population distribution.

\paragraph{Shifted populations.}
Each PK/PD parameter is a transformation of a standard random effect $z \sim \mathcal{N}(0, 1)$
(Appendix~\ref{app:pkpd_levodopa}).
We define the shifted populations by replacing, for some specific parameters, the random effect $z$ by \begin{equation}
    z' = s\,P + \tfrac{1}{2}\, z, \qquad \text{so that } z' \sim \mathcal{N}\big(s\,P,\ 0.5^2\big),
    \label{eq:ood_shift}
\end{equation}
where $P \geq 0$ is the size of the shift, in standard deviations of the nominal population, and $s \in \{-1, +1\}$ its direction, fixed for each parameter.
The variance of the random effect is reduced to concentrate the patients around the shifted mean.
Since $\Phi$ spans the $0.025$ to $0.975$ quantiles of the nominal marginals, that is $|z| \leq 1.96$, the shifted parameters of most patients lie inside $\Phi$ for $P = 1.5$ and outside $\Phi$ for $P = 3$. We consider four types of shift, chosen for their clinical interpretation (Table~\ref{tab:ood_types}), at $P \in \{0, 1.5, 3\}$. At $P = 0$ only the spread is reduced, this level serves as an in-distribution reference.

\begin{table}[htbp]
\centering
\caption{The four types of distribution shift. Arrows give the direction $s$ in which the parameter value moves.
The drug-free baseline $\mathrm{BASE}$ is raised towards $0$ in all types: lowering it makes
$E_{\max} + \mathrm{BASE} \leq 0$ for many patients, for whom no dose reaches the target response.}
\label{tab:ood_types}
\small
\begin{tabular}{lll}
\toprule
Type & Shifted parameter & Effect on the patient \\
\midrule
Fast absorption & \quad\quad $\mathrm{fa}^{\mathrm{LD}} \uparrow$  & larger share of the dose absorbed through the fast route \\
Slow absorption & \quad\quad $\mathrm{fa}^{\mathrm{LD}} \downarrow$ & larger share of the dose absorbed through the slow route \\
High sensitivity & \quad\quad $\mathrm{EC}_{50} \downarrow$ & lower target concentration $c^\star_\phi$ \\
Low sensitivity & \quad\quad $\mathrm{EC}_{50} \uparrow$ & higher target concentration $c^\star_\phi$ \\
\bottomrule
\end{tabular}
\end{table}

\paragraph{Matched populations.}
The populations share their patients. 
We draw the patients once from the nominal population, keeping those that pass the pruning rule of equation~\ref{eq:pruning}, and obtain each shifted population by applying equation~\ref{eq:ood_shift} to these same patients. 
A given patient therefore has the same covariates, the same values of all the parameters that are not shifted, and the same random effect $z$ in equation~\ref{eq:ood_shift} across the four types and the three shift sizes.
The seeds of the feedback noise and of the initialization are also shared, so that the populations differ only by the shifted parameters.
The shifted version of a patient is kept even if it no longer satisfies equation~\ref{eq:pruning}; we call such patients \emph{untreatable}, since their target concentration is either not reached by a single dose of $d_h$ or already exceeded by a single dose of $d_l$, and we report their proportion, and optimize dosing for them as well. 

\paragraph{Protocol.}
We use $60$ patients per population and the \emph{concentration feedback} setting of Section~\ref{sec:levodopa_experiments}, with the same initialization, budget of $50$ iterations, constraints and hyperparameters.
We compare BOASS-$k_\mathcal{D}$, BOASS-$k_\Phi$, BOASS-random-$k_\mathcal{D}$, BO-on-$\mathcal{D}$, $\pi$BOASS-on-$\mathcal{D}$, $\pi$BOASS-on-$\mathcal{D}$-strong, and PCA-BO-on-$\mathcal{D}$ .
We have omitted the comparison with the other methods benchmarked in Figure \ref{fig:regret} due to computational constraints.
Regret is measured against the oracle regimen of each shifted patient, obtained by direct optimization of the utility.

\begin{figure}[htbp]
\centering
\includegraphics[width=\linewidth]{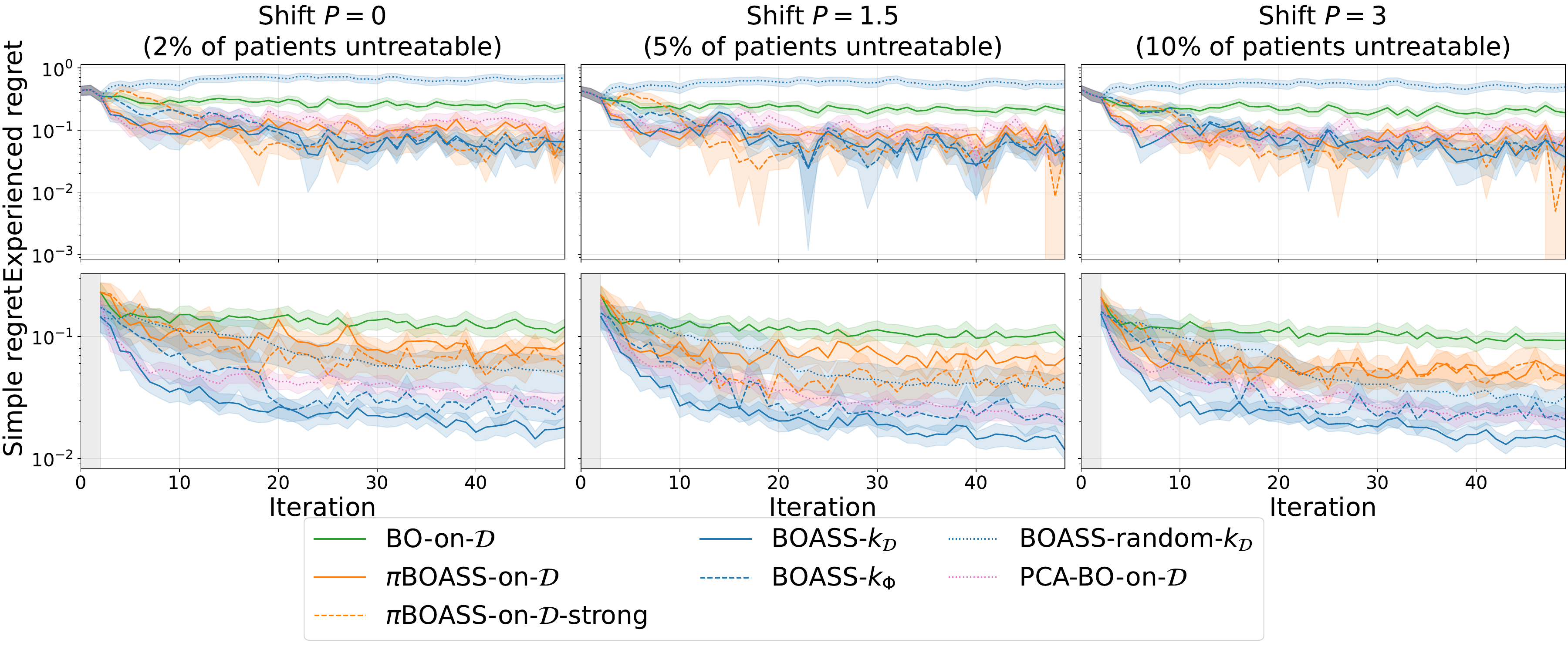}
\caption{Out-of-distribution patients, fast absorption. Top row: experienced regret. Bottom row: simple regret of the
posterior mean maximizer, computed after the initialization (grey band). Columns: size $P$ of the shift, with the
proportion of untreatable patients. Averages over $60$ patients, the same in the three columns. Shaded bands:
10th--90th percentile of the estimated mean, assuming Gaussian regret.}
\label{fig:ood_fastabs}
\end{figure}

\begin{figure}[htbp]
\centering
\includegraphics[width=\linewidth]{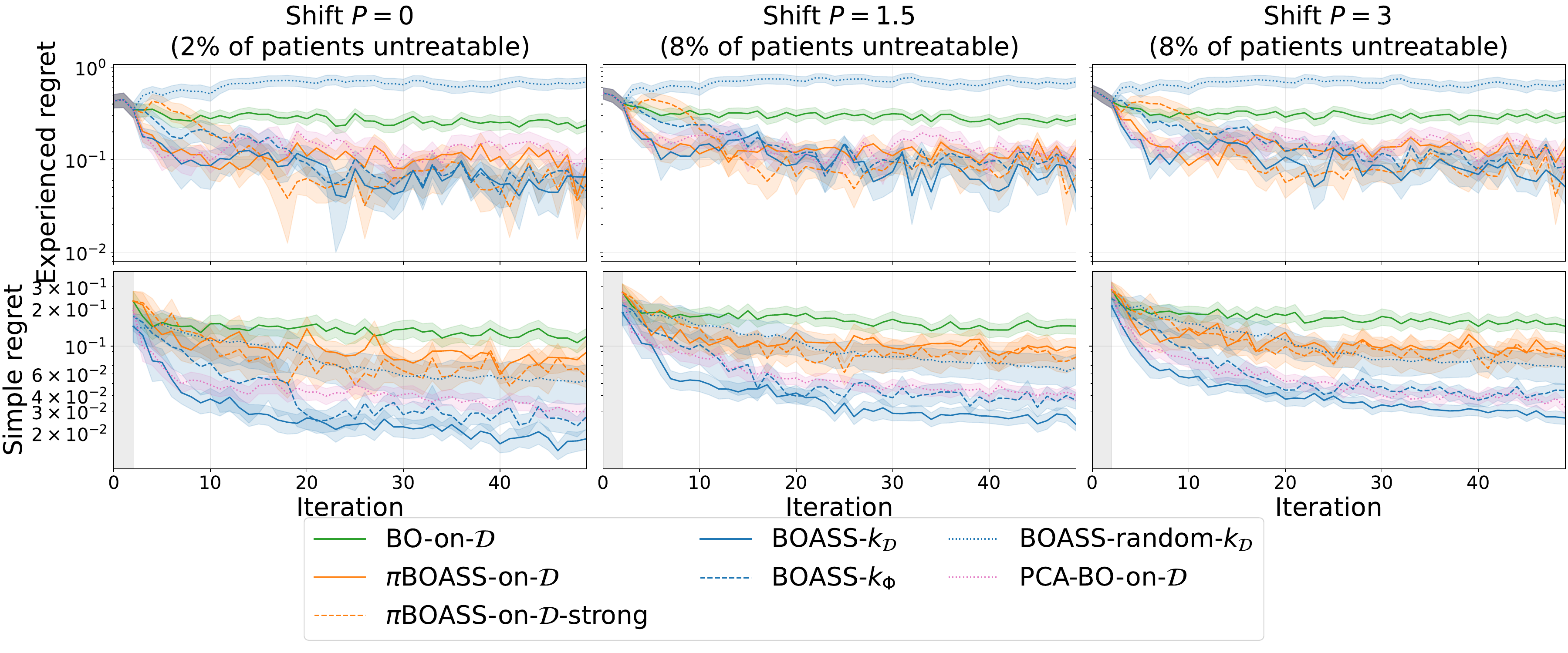}
\caption{Out-of-distribution patients, slow absorption. Same layout as Figure~\ref{fig:ood_fastabs}.}
\label{fig:ood_slowabs}
\end{figure}

\begin{figure}[htbp]
\centering
\includegraphics[width=\linewidth]{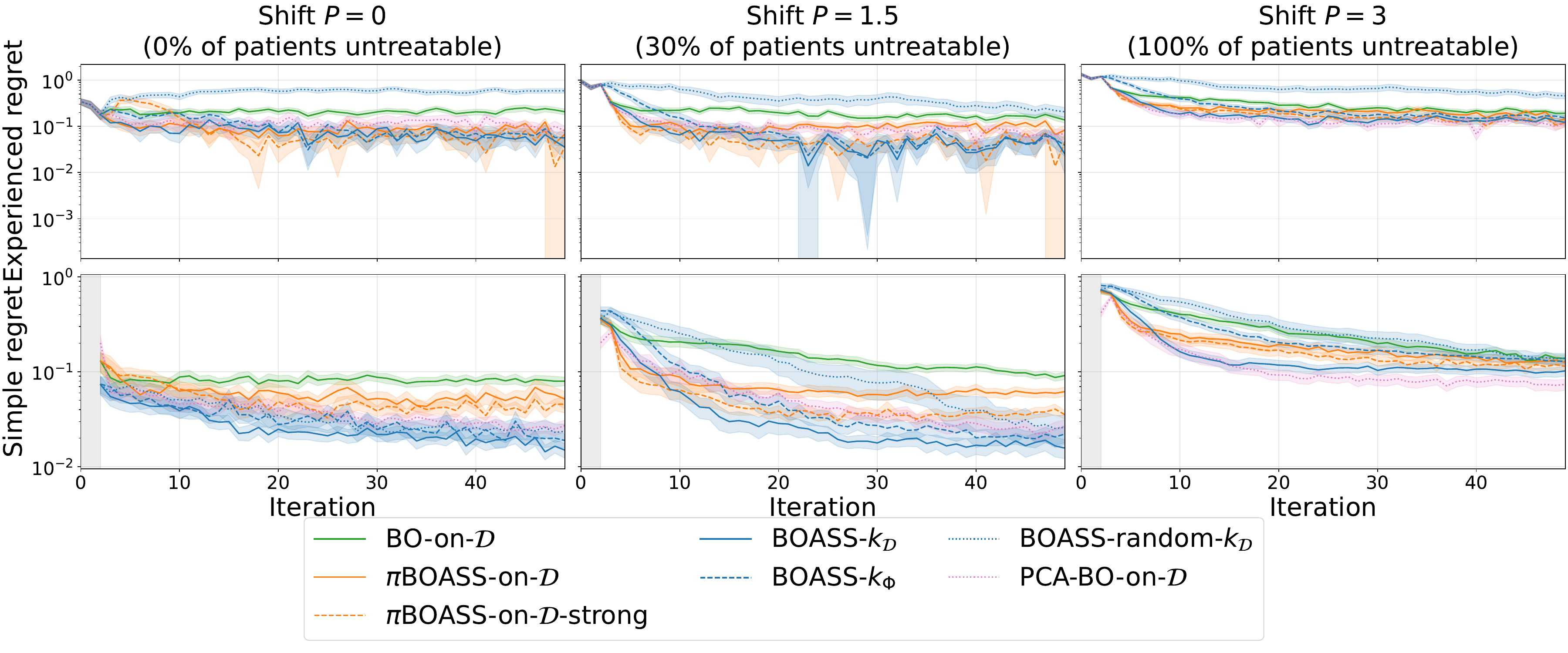}
\caption{Out-of-distribution patients, high sensitivity. Same layout as Figure~\ref{fig:ood_fastabs}.}
\label{fig:ood_highsens}
\end{figure}

\begin{figure}[htbp]
\centering
\includegraphics[width=\linewidth]{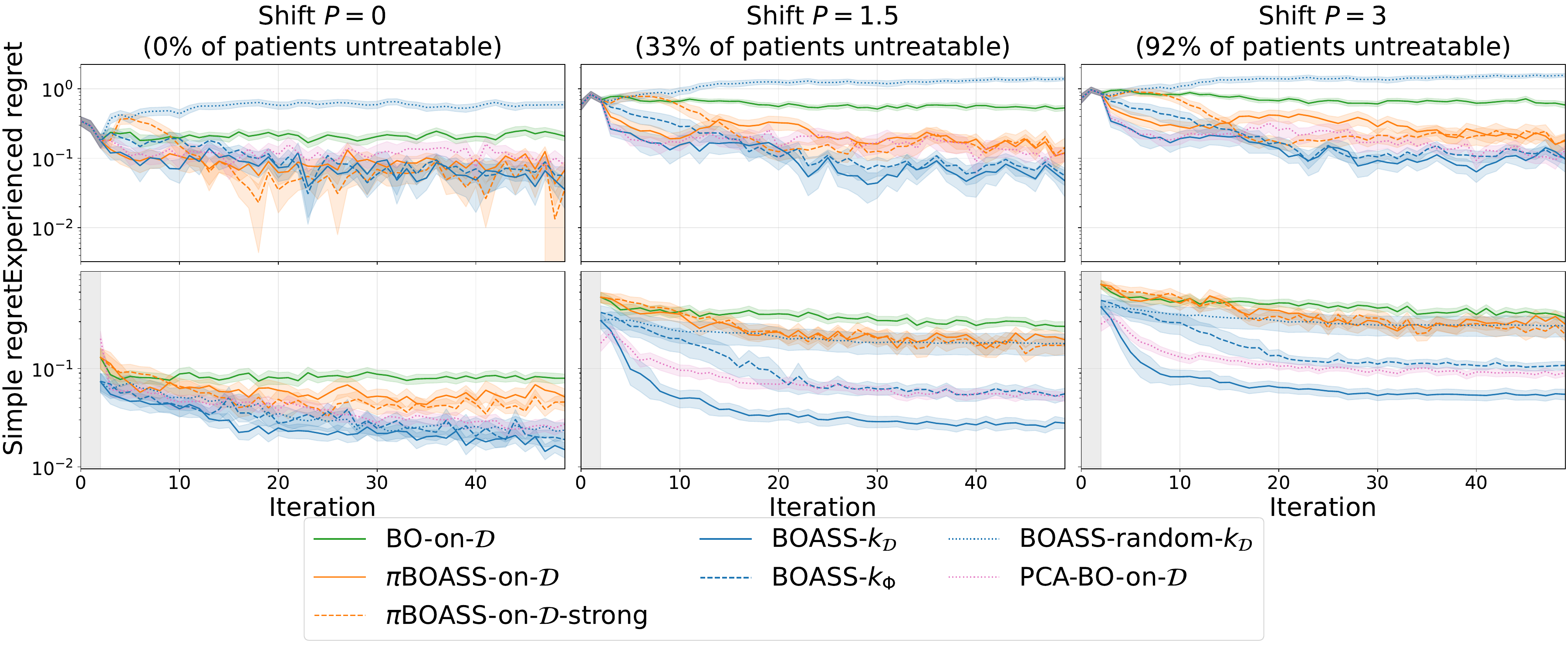}
\caption{Out-of-distribution patients, low sensitivity. Same layout as Figure~\ref{fig:ood_fastabs}.}
\label{fig:ood_lowsens}
\end{figure}

\begin{figure}[htbp]
\centering
\includegraphics[width=\linewidth]{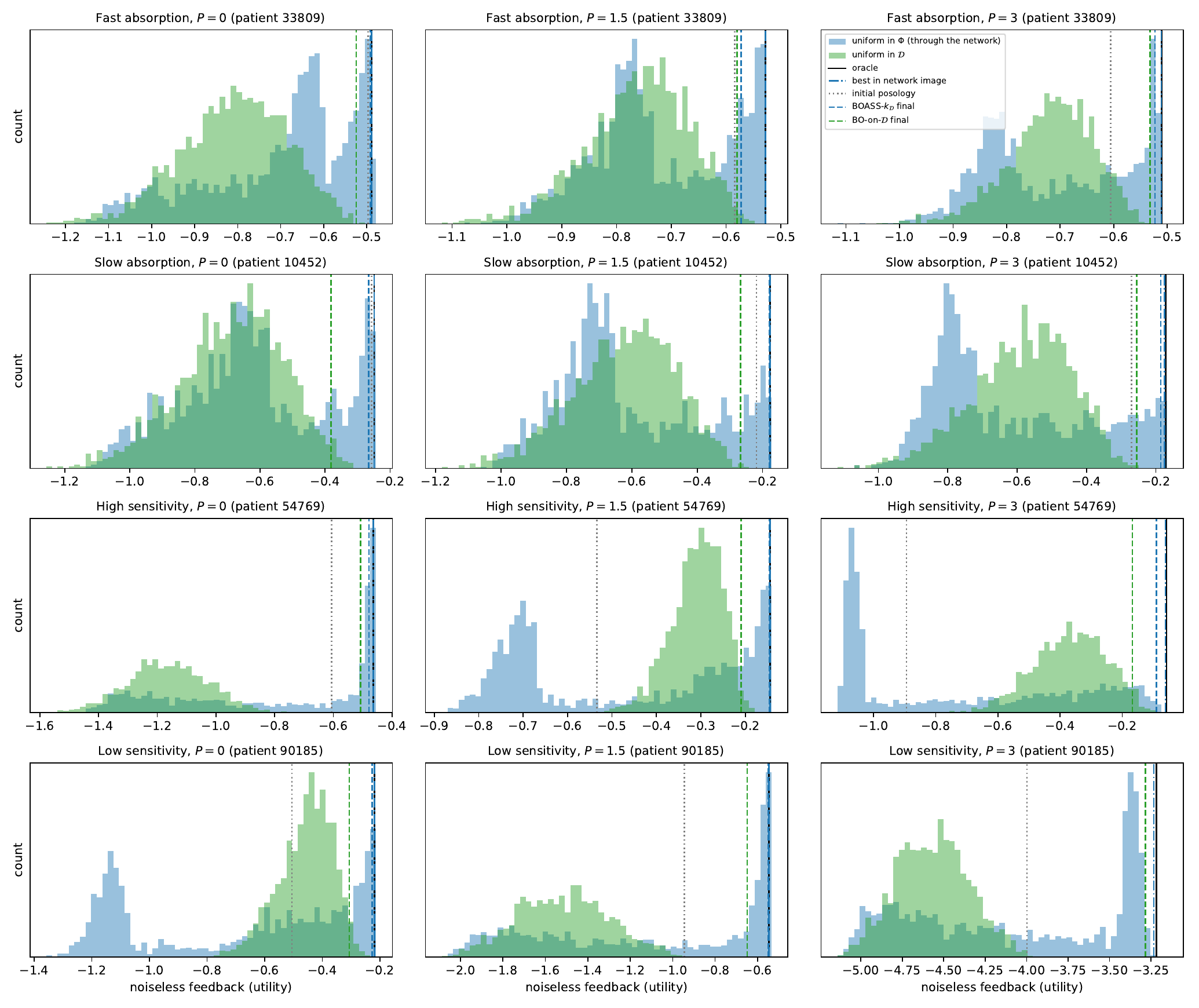}
\caption{Noiseless utility of regimens sampled uniformly in $\Phi$ (through the dosing network, blue) and in
$\mathcal{D}$ (green), for one randomly drawn patient per type of shift (rows) at each shift size (columns). Vertical
lines: oracle regimen, best regimen of $\mathcal{M}$, initial regimen, and final posterior mean maximizers of
BOASS-$k_\mathcal{D}$ and BO-on-$\mathcal{D}$.}
\label{fig:ood_histograms}
\end{figure}

\paragraph{Results.} Figures \ref{fig:ood_fastabs}-\ref{fig:ood_slowabs}-\ref{fig:ood_highsens}-\ref{fig:ood_lowsens} show the results for all the scenarios listed in Table \ref{tab:ood_types}. All BOASS methods show robustness to population shift, including the largest level (P=3).
PCA-BO-on-$\mathcal{D}$, which also
searches a low-dimensional set built from the simulator, is the closest competitor, beating BOASS-$k_{\phi}$ only on the simple regret of the high-sensitivity setting with P=3, in which all simulated patients are \textit{untreatable}.
Nonetheless, BOASS still remains the best performing method (either alone or tied) across the rest of scenarios.
The robustness of BOASS follows from the fact that a patient whose parameters lie outside $\Phi$ does not necessarily have an optimal regimen outside $\mathcal{M}$: the map $\mathcal{E}_w$ is not injective (Appendix~\ref{app:intrinsic_dimension}), so other points of $\Phi$ can produce the regimen that the shifted patient needs. 
To illustrate this, we draw one patient at random for each type of shift and compare, at each shift size, the noiseless utility of $2000$ regimens obtained from points drawn uniformly in the unit box of $\Phi$ and mapped through $\mathcal{E}_w$, and of $2000$ regimens drawn uniformly in the unit box of $\mathcal{D}$ (Figure~\ref{fig:ood_histograms}).
We also report the best regimen of $\mathcal{M}$, obtained by maximizing the patient's utility over $\Phi$ with L-BFGS-B. 
In every panel, the best regimen of $\mathcal{M}$ is indistinguishable from the oracle, so the search-space bias $\Delta_{\mathcal{M}}$ of Section~\ref{sec:regret} remains negligible under these shifts. 
Moreover, a visible fraction of the regimens sampled through $\Phi$ lies close to the oracle, whereas the regimens sampled in $\mathcal{D}$ stop well short of it. 
The shift therefore changes which region of $\Phi$ is optimal, but not the fact that near-optimal regimens are easier to find in $\Phi$ than in $\mathcal{D}$.

\section{Bayesian optimization details}
\label{app:bo}

\subsection{Search space and constraints}
\paragraph{Dosing space.} We define the dosing space $\mathcal{D}$ of Section~\ref{sec:levodopa} through a map $[0,1]^{2M} \to \mathcal{D}$, built so that every point of the unit box corresponds to an admissible regimen. Let $W = (t_f - t_0) - (M-1)\,\Delta$ be the free time to allocate between intakes where $\Delta$ is the minimum time between two intakes.
For $\mathbf{x}=(\mathbf{u},\mathbf{v}) \in [0,1]^{M} \times [0,1]^{M}$ and $i=1,\dots,M$, the corresponding dose $d_i$ and intake time $t_i$ are
\begin{equation}
\label{eq:decode}
\begin{cases}
d_i = u_i\, d_{\max}, \\[2pt]
t_i = t_0 + (i-1)\,\Delta + \sum_{k=1}^{i-1} \frac{v_k}{\sum_{j=1}^{M} v_j}\, W.
\end{cases}
\end{equation}

\paragraph{Search space rescaling.} For all methods, the BO search space is rescaled to the unit box. When
the search space is $\mathcal{D}$, the rescaling is the right inverse of \eqref{eq:decode} that normalizes
$\mathbf{v}$ to sum to one; this fixes the one redundant degree of freedom of \eqref{eq:decode}, whose
decoded regimen is invariant to a positive rescaling of $\mathbf{v}$. When it is $\Phi$, each parameter is
rescaled affinely in the scale in which its random effect is Gaussian (logarithmic for the log-normal
parameters, and reflected for those, such as the PD baseline, defined on the negative scale), centred on
that patient's covariate-adjusted typical value, so that $x=0$ and $x=1$ are the $0.025$ and $0.975$
quantiles of the patient's own marginal of $p(\phi)$. For our method with the pullback kernel $k_\mathcal{D}$, the kernel is
evaluated on the regimens $\mathcal{E}_w(\phi)$, mapped to the unit box with the same right inverse of
\eqref{eq:decode}, so that both search spaces present the kernel with identical coordinates.

\paragraph{Constraint handling.} Dosing regimens are usually subject to practical constraints, for instance a minimum and maximum intake dose, a maximum total daily dose, and a discrete set of available tablet strengths. These constraints can often be enforced directly in the output layer of $\mathcal{E}_w$, rather than handled during the online Bayesian optimization step.
The constraint on the variation of the total daily dose is a form of
transition-constrained BO, which can be treated as a Markov decision process \citep[see, e.g.,][]{paulson_efficient_2022,folch_transition_2024}. Here we take a simpler approach, restricting the optimization of the acquisition function at each step: in practice, we assign an arbitrary low value to the points violating the constraints. This simplification does not affect the purpose of the comparison, as we use the same
approach for all methods. Discarding infeasible points in this way makes the acquisition landscape more challenging to optimize, as it introduces flat regions and sharp transitions; we counterbalance it with a large number of restarts in the acquisition function optimization.

\paragraph{Constrained acquisition function.} Let $\mathbf{x}$ be a point of the unit-box search space (of
$\mathcal{D}$ or of $\Phi$), $(\mathbf{d}(\mathbf{x}),\mathbf{t}(\mathbf{x}))$ the regimen it encodes, and
$D(\mathbf{x}) = \sum_{i=1}^M d_i(\mathbf{x})$ its total daily dose. Writing $D_t$ for the total daily dose
administered at iteration $t$, the feasible set at that iteration is
\begin{equation}
\label{eq:feasible-set}
\mathcal{C}_t \;=\; \big\{\mathbf{x} : D(\mathbf{x}) \geq D_{\min}\big\} \;\cap\;
\big\{\mathbf{x} : \lvert D(\mathbf{x}) - D_t\rvert \leq \rho\, D_t \big\},
\end{equation}
with $D_{\min}=320$~mg and $\rho=0.4$ as in Section~\ref{sec:levodopa}, and the acquisition function we
actually optimize is
\begin{equation}
\label{eq:limited-acq}
\tilde{\alpha}\big(\mathbf{x}\,;\,\mathcal{H}_t\big) \;=\;
\begin{cases}
\alpha\big(\mathbf{x}\,;\,\mathcal{H}_t\big), & \mathbf{x}\in\mathcal{C}_t,\\[2pt]
\alpha_\emptyset, & \text{otherwise}.
\end{cases}
\end{equation}
We set $\alpha_\emptyset=0$ for EUBO, which is its infimum, and $\alpha_\emptyset=-10^{9}$ for log expected
improvement, far below any value it attains. The total dose $D(\mathbf{x})$ is evaluated on the
undiscretized regimen, before rounding to the tablet grid, so that $\mathbf{x}\mapsto D(\mathbf{x})$ remains
differentiable inside the acquisition optimization: on $\mathcal{D}$ it is linear in $\mathbf{x}$, and on
$\Phi$ it is the total dose returned by the dosing network. In the rare case where every restart of the
optimizer returns an infeasible point, we fall back to a candidate drawn uniformly from $\mathcal{C}_t$, by
rejection sampling on a Sobol batch of $4096$ points; the random acquisition baseline draws its candidates
in the same way. The reported best guess maximizes the GP posterior mean over
$\{\mathbf{x}: D(\mathbf{x})\geq D_{\min}\}$ instead: the minimum daily dose applies to it, but the
$\rho$-band does not, since the best guess is a recommended destination rather than a step along the path.

\subsection{Implementation details}
\paragraph{Acquisition function optimization.} We maximize \eqref{eq:limited-acq} with BoTorch's multi-start
routine, returning one candidate per iteration: $100$ restarts, selected by its initialization heuristic
from $256$ raw Sobol samples, each run with L-BFGS-B for at most $500$ iterations and $150$ line search
steps per iteration. Log expected improvement is estimated with $100$ quasi-Monte Carlo samples, over a
baseline pruned to the observed points that can still be optimal; the posterior mean maximized for the best
guess is estimated with $1024$ samples.

\paragraph{GP surrogate.} In the three settings with a scalar observed utility (concentration feedback and the two misspecified settings), the outcomes are standardized before the GP is fitted: they are centred and
scaled to unit variance with their empirical mean and standard deviation over the observations available at that iteration, and the posterior is mapped back to the original scale. The utility scales differ across the three settings --- an unnormalized squared concentration error in \eqref{eq:utility} against a
cumulative dose error in \eqref{eq:utility} --- so this keeps one set of GP priors appropriate for
all of them. No such transformation is needed in the preference setting, where the GP is fitted to
comparisons and the scale of the latent utility is set by the likelihood. Kernel hyperparameters
(lengthscales and output scale, under $\Gamma(2,2)$ priors) are re-estimated at every iteration by
maximizing the marginal likelihood, starting from the first acquisition iteration; if the fit does not
converge, the previous iteration's values are kept.

\paragraph{Software.} The online optimization is implemented with BoTorch \citep{balandat_botorch_2020} and GPyTorch
\citep{gardner_gpytorch_2018}. The dosing network is implemented in PyTorch \citep{paszke_pytorch_2019}, combined with JAX
\citep{bradbury_jax_2018} and diffrax \citep{kidger_diffrax_2022} for the ODE solver.

\subsection{Evaluation settings}
\label{app:settings}

In the \emph{misspecified feedback} setting, the biased oracle regimen is obtained by a forward pass of the dosing network on the patient's true $\phi$, followed by a perturbation of the dose amount at each intake with an additive
shift drawn from $\{s\,\Delta_\text{tab}\}_{s=-S}^S$ with weights $\{(S+1-|s|)/(S+1)^2\}_{s=-S}^S$ and $S=5$.
The first dose alone is perturbed with a negative shift, large enough to bring it below the largest of the subsequent
doses.
The first dose is lowered below the largest of the subsequent doses, which places the oracle outside $\mathcal{M}$: under the selected PK/PD model the model-optimal regimens have a first intake larger than or equal to the subsequent ones.

\section{PK/PD model for Levodopa-Carbidopa in Parkinson's disease}
\label{app:pkpd_levodopa}
The use-case is the Levodopa-Carbidopa combination therapy, the standard oral treatment for Parkinson's disease.
Levodopa is the active principle: a dopamine precursor that crosses the blood-brain barrier, where it is converted to dopamine. Carbidopa is a peripheral decarboxylase inhibitor that blocks this conversion outside the brain, so that more Levodopa reaches the central nervous system and peripheral side effects are reduced. Because Carbidopa inhibits the peripheral metabolism of Levodopa, the two drugs are pharmacokinetically coupled, and the model accounts for this interaction.

We combine the population pharmacokinetic model of \citet{senek_population_2018} with the pharmacodynamic model of \citet{westin_pharmacokineticpharmacodynamic_2011}, whose clinical effect is the treatment response scale (TRS), a score ranging from $-3$ (\emph{off}, parkinsonism) to $+3$ (dyskinesia), with $0$ indicating the optimal response \citep{nyholm_duodenal_2005}. 
Together they define the simulator $(f_\text{PK},f_\text{PD})$ and the population distribution $p(\phi)$ of Section~\ref{sec:pkpd}. 
The vector $\phi_i$ collects the patient-specific PK/PD parameters. In this application, the concentration entering the PD model is the central Levodopa plasma concentration, $c_e(t)=c_{\rm LD}(t)=A_C^{\rm LD}(t)/V_C^{\rm LD}$.

Throughout, the subscript $i$ indexes the patient and the superscripts LD and CD refer to Levodopa and Carbidopa. We
write $A_\bullet$ for the amount of drug in a compartment; these amounts are the state variables of the ODE system, and
the plasma concentration driving the pharmacodynamics is the central Levodopa amount divided by its volume of
distribution. To keep the notation light, we suppress the patient subscript $i$ on the PK/PD parameters inside the
differential equations and restore it in the parameter models that follow.

\subsection{Pharmacokinetic (PK) model}
Levodopa follows a two-compartment disposition model (a central and a peripheral compartment) and Carbidopa a
one-compartment disposition model (a central compartment only). For each drug, the oral dose reaches the central
compartment through two parallel absorption routes, a fast and a slow transit chain. A transit chain is a series of
compartments through which the dose passes before entering the circulation, and it delays and spreads the absorption. We
write $n_1^\bullet$ and $n_2^\bullet$ for the number of compartments in the fast and slow chains, $k_1^\bullet$ and
$k_2^\bullet$ for their transit rate constants, $d_\text{LD}$ and $d_\text{CD}$ for the administered doses, $F$ for the
bioavailability, and $\text{fa}^\bullet$ for the fraction of the dose entering the fast chain.

\paragraph{Levodopa fast-transit chain.} The fast chain has $n_1^\text{LD}$ compartments
$A^\text{LD}_{1,1},\dots,A^\text{LD}_{1,n_1^\text{LD}}$, in which the fast fraction of the dose enters the first
compartment,
\begin{equation}\label{eq:ld-fast}
\begin{cases}
    \dot{A}^\text{LD}_{1,1} = -k_1^\text{LD} A^\text{LD}_{1,1}, & A^\text{LD}_{1,1}(0) = F_i\,\text{fa}^\text{LD}_i\,
d_\text{LD}, \\
    \dot{A}^\text{LD}_{1,j} = k_1^\text{LD} A^\text{LD}_{1,j-1} - k_1^\text{LD} A^\text{LD}_{1,j}, &
j=2,\dots,n_1^\text{LD}.
\end{cases}
\end{equation}

\paragraph{Levodopa slow-transit chain.} The slow chain has $n_2^\text{LD}$ compartments
$A^\text{LD}_{2,1},\dots,A^\text{LD}_{2,n_2^\text{LD}}$, which receive the complementary fraction of the dose,
\begin{equation}\label{eq:ld-slow}
\begin{cases}
    \dot{A}^\text{LD}_{2,1} = -k_2^\text{LD} A^\text{LD}_{2,1}, & A^\text{LD}_{2,1}(0) =
F_i\,(1-\text{fa}^\text{LD}_i)\, d_\text{LD}, \\
    \dot{A}^\text{LD}_{2,j} = k_2^\text{LD} A^\text{LD}_{2,j-1} - k_2^\text{LD} A^\text{LD}_{2,j}, &
j=2,\dots,n_2^\text{LD}.
\end{cases}
\end{equation}

\paragraph{Levodopa central and peripheral compartments.} The central compartment $A^\text{LD}_C$ (volume
$V^\text{LD}_C$) collects the output of both chains and exchanges with the peripheral compartment $A^\text{LD}_P$
(volume $V^\text{LD}_P$) at intercompartmental clearance $Q^\text{LD}$, while Levodopa is eliminated at clearance
$\text{CL}^\text{LD}$,
\begin{equation}\label{eq:ld-disp}
\begin{aligned}
    \dot{A}^\text{LD}_C &= k_1^\text{LD} A^\text{LD}_{1,n_1^\text{LD}} + k_2^\text{LD} A^\text{LD}_{2,n_2^\text{LD}} +
\frac{Q^\text{LD}}{V^\text{LD}_P} A^\text{LD}_P - \frac{Q^\text{LD} + \text{CL}^\text{LD}}{V^\text{LD}_C} A^\text{LD}_C,
\\
    \dot{A}^\text{LD}_P &= -\frac{Q^\text{LD}}{V^\text{LD}_P} A^\text{LD}_P + \frac{Q^\text{LD}}{V^\text{LD}_C}
A^\text{LD}_C.
\end{aligned}
\end{equation}
The Levodopa plasma concentration driving the pharmacodynamics is $c^\text{LD} = A^\text{LD}_C / V^\text{LD}_C$.

\paragraph{Carbidopa transit chains and central compartment.} Carbidopa is absorbed through its own fast and slow
chains, $A^\text{CD}_{1,1},\dots,A^\text{CD}_{1,n_1^\text{CD}}$ and
$A^\text{CD}_{2,1},\dots,A^\text{CD}_{2,n_2^\text{CD}}$, with the same structure as for Levodopa,
\begin{equation}\label{eq:cd-chains}
\begin{cases}
    \dot{A}^\text{CD}_{1,1} = -k_1^\text{CD} A^\text{CD}_{1,1}, & A^\text{CD}_{1,1}(0) = F_i\,\text{fa}^\text{CD}_i\,
d_\text{CD}, \\
    \dot{A}^\text{CD}_{1,j} = k_1^\text{CD} A^\text{CD}_{1,j-1} - k_1^\text{CD} A^\text{CD}_{1,j}, &
j=2,\dots,n_1^\text{CD}, \\[4pt]
    \dot{A}^\text{CD}_{2,1} = -k_2^\text{CD} A^\text{CD}_{2,1}, & A^\text{CD}_{2,1}(0) =
F_i\,(1-\text{fa}^\text{CD}_i)\, d_\text{CD}, \\
    \dot{A}^\text{CD}_{2,j} = k_2^\text{CD} A^\text{CD}_{2,j-1} - k_2^\text{CD} A^\text{CD}_{2,j}, &
j=2,\dots,n_2^\text{CD}.
\end{cases}
\end{equation}
The single (central) Carbidopa compartment $A^\text{CD}_C$ (volume $V^\text{CD}_C$) collects the output of both chains
and eliminates Carbidopa at clearance $\text{CL}^\text{CD}$,
\begin{equation}\label{eq:cd-central}
    \dot{A}^\text{CD}_C = k_1^\text{CD} A^\text{CD}_{1,n_1^\text{CD}} + k_2^\text{CD} A^\text{CD}_{2,n_2^\text{CD}} -
\frac{\text{CL}^\text{CD}}{V^\text{CD}_C} A^\text{CD}_C.
\end{equation}

\paragraph{Disposition parameters and covariates.} The disposition parameters (clearances and volumes) depend on patient
covariates through an allometric weight term and, for the clearances, additional covariates: the Levodopa clearance
depends on the Hoehn-Yahr stage $\text{HY}_i$ and, through the metabolic interaction, on the Carbidopa dose
$d_\text{CD}$, while the Carbidopa clearance depends on the patient age $\text{AGE}_i$,
\begin{equation}\label{eq:disp-params}
\begin{aligned}
    \text{CL}^\text{LD}_i &= \exp\!\left(\log \text{CL}^\text{LD}_0 + \alpha\log\tfrac{W_i}{70} +
\beta_\text{HY}(\text{HY}_i-\overline{\text{HY}}) - \log\!\left(1 + \tfrac{d_\text{CD}}{\beta_\text{int}}\right) +
\eta^{\text{CL}^\text{LD}}_i\right), \\
    Q^\text{LD}_i &= \exp\!\left(\log Q^\text{LD}_0 + \alpha\log\tfrac{W_i}{70} + \eta^{Q^\text{LD}}_i\right), \\
    V^\text{LD}_{C,i} &= \exp\!\left(\log V^\text{LD}_{C,0} + \alpha\log\tfrac{W_i}{70} + \eta^{V^\text{LD}_C}_i\right),
\\
    V^\text{LD}_{P,i} &= \exp\!\left(\log V^\text{LD}_{P,0} + \alpha\log\tfrac{W_i}{70} + \eta^{V^\text{LD}_P}_i\right),
\\
    \text{CL}^\text{CD}_i &= \exp\!\left(\log \text{CL}^\text{CD}_0 + \alpha\log\tfrac{W_i}{70} +
\beta_\text{AGE}(\text{AGE}_i-\overline{\text{AGE}}) + \eta^{\text{CL}^\text{CD}}_i\right), \\
    V^\text{CD}_{C,i} &= \exp\!\left(\log V^\text{CD}_{C,0} + \alpha\log\tfrac{W_i}{70} + \eta^{V^\text{CD}_C}_i\right),
\end{aligned}
\end{equation}
where a subscript $0$ denotes the typical (population) value, $\alpha$ is the allometric scaling exponent,
$\beta_\text{HY}$ and $\beta_\text{AGE}$ are covariate coefficients, and $70$~kg is the reference body weight. The term
$-\log\!\left(1 + d_\text{CD}/\beta_\text{int}\right)$ encodes the reduction in Levodopa clearance caused by Carbidopa,
with $\beta_\text{int}$ controlling the strength of the interaction.

\paragraph{Absorption parameters.} The fast-chain fractions are bounded in $(0,1)$ and modeled
on the logit scale, while the transit rate constants are parametrized through the mean transit times
$\text{MTT}^\bullet$ on the log scale,
\begin{equation}\label{eq:abs-params}
\begin{aligned}
    \text{fa}^\text{LD}_i &= \text{expit}\!\left(\text{logit}(\text{fa}^\text{LD}_0) +
\eta^{\text{fa}^\text{LD}}_i\right), \\
    \text{fa}^\text{CD}_i &= \text{expit}\!\left(\text{logit}(\text{fa}^\text{CD}_0) +
\eta^{\text{fa}^\text{CD}}_i\right), \\
    k^\bullet_{r,i} &= 1/\text{MTT}^\bullet_{r,i}, \quad \text{MTT}^\bullet_{r,i} = \exp\!\left(\log
\text{MTT}^\bullet_{r,0} + \eta^{\text{MTT}^\bullet_r}_i\right), \quad r\in\{1,2\},\ \bullet\in\{\text{LD},\text{CD}\}.
\end{aligned}
\end{equation}
It was not possible to reconstruct how the bioavailability $F_i$ was parametrised in terms of random effects in the original model, therefore we consider here it fixed at its typical value $F_0$.

\paragraph{Random effects and inter-patient variability.} Each PK/PD parameter carries a patient-specific random effect
$\eta^\bullet_i \sim \mathcal{N}(0,\omega^2_\bullet)$ that captures inter-patient variability. Following common practice, unconstrained parameters (the PD baseline and
maximum effect below) use additive normal random effects; positive parameters (clearances, volumes, mean transit times)
use log-normal effects, i.e.\ a normal effect on the log scale; and parameters bounded in $(0,1)$ (bioavailability and
fast-chain fractions) use logit-normal effects, i.e.\ a normal effect on the logit scale.

Pharmacometric reports usually summarize each $\omega_\bullet$ as a percentage coefficient of variation,
$\text{CV}\%_\bullet = 100\,\operatorname{Std}[\cdot]/\operatorname{E}[\cdot]$, the ratio of the standard deviation to
the mean of the parameter distribution. For a log-normal parameter this has the closed form $\text{CV}\%_\bullet =
100\sqrt{\exp(\omega^2_\bullet) - 1}$; for a logit-normal parameter no closed form exists and the two moments are
evaluated numerically. We take the fixed-effect and random-effect estimates from \citet{senek_population_2018} for the
PK model and \citet{westin_pharmacokineticpharmacodynamic_2011} for the PD model.

\subsection{Pharmacodynamic (PD) model}

The PD model maps the central Levodopa plasma concentration $c_{\rm LD}$ to the TRS through a sigmoidal $E_{\max}$ model with a baseline and a Hill coefficient,
\begin{equation}
    \mathrm{TRS}_i(c_{\rm LD})
    =
    \mathrm{BASE}_i
    +
    \frac{E_{\max,i}\,c_{\rm LD}^{\gamma}}
         {c_{\rm LD}^{\gamma}+EC_{50,i}^{\gamma}},
\label{eq:pd-trs}
\end{equation}
where $\text{BASE}_i$ is the drug-free baseline, $E_{\max,i}$ the maximum drug effect, $\text{EC}_{50,i}$ the
concentration producing half of $E_{\max,i}$, and $\gamma$ the Hill coefficient controlling the steepness of the
concentration-effect curve. In the notation of Section~\ref{sec:pkpd}, this map is the pharmacodynamic model $f_\text{PD}$ and $\text{TRS}_i$ is the effect $r(t;\phi)$. All parameters
carry random effects,
\begin{equation}\label{eq:pd-re}
\begin{cases}
    \text{BASE}_i = \text{BASE}_0 + \eta_i^{\text{BASE}}, \\
    E_{\max,i} = E_{\max,0} + \eta_i^{E_{\max}}, \\
    \text{EC}_{50,i} = \exp\!\left(\log \text{EC}_{50,0} + \eta_i^{\text{EC}_{50}}\right), \\
    \gamma_i = \exp\!\left(\log \gamma_0 + \eta_i^{\gamma}\right)
\end{cases}
\end{equation}
The fixed-effect and random-effect estimates are taken from \citet{westin_pharmacokineticpharmacodynamic_2011}.

\subsection{PKPD solution under multiple intakes}
\label{app:ode_intakes}

Let $c_{e}(s;\phi\mid (\mathbf{d},\mathbf{t}))$ be the concentration in the effect compartment at time $s$ induced by the regimen $(\mathbf{d},\mathbf{t})$. In practice, it is computed by integrating the ODE system piecewise between successive intakes, updating the initial conditions at each dosing time, that is, for $t\in(t_i, t_{i+1})$:
\begin{equation}
\label{eq:pkpd-ode-piecewise}
\begin{aligned}
\mathbf{c}(t_i) &= \mathbf{c}(t_i^-) + g(d_i;\phi)\,\mathbf{e}_1, \text{ and }
c_e(t) &= \Big[\, \mathbf{c}(t_i) + \int_{t_i}^{t} f_\text{PK}\big(\mathbf{c}(s);\phi\big)\,ds \,\Big]_e ,
\end{aligned}
\end{equation}
where $\mathbf{e}_1$ is the first canonical basis vector of $\mathbb{R}^K$, so that the intake $d_i$ enters the dosing compartment while the other compartments carry over their concentrations at $t_i^-$, and $[\,\cdot\,]_e$ denotes the component corresponding to the effect compartment.

\section{Intrinsic dimension of the amortized search space}
\label{app:intrinsic_dimension}

The model-optimal regimen $\mathcal{E}^\star(\phi)$ of Section~\ref{sec:pkpd} does not depend on all coordinates of $\phi$ independently. This follows from two observations which we can make directly analysing the PK/PD equations.

First, Carbidopa acts on Levodopa only through its dose amount
$d_\text{CD}$ in the clearance of Levodopa \eqref{eq:disp-params}, and this dose is fixed by the
Levodopa regimen due to the pill format. Therefore, the  PK parameters of Carbidopa are nuisance parameters for the optimal dosage.
Second, the PD parameters $\text{BASE}$, $E_{\max}$, $\text{EC}_{50}$ and $\gamma$ enter the dosing objective only through the scalar target concentration $c^\star_\phi$. The TRS \eqref{eq:pd-trs} is monotone, so the target concentration is unique, which gives only one degree of freedom. Therefore, those 4 parameters are not fully identifiable from the optimal dosage. This gives us the bound $\text{dim}(\mathcal{M})\leq 7$.

Additional structural non-idenfiability could come from the complex relationship between the PK parameters and the optimal dosing solution, which cannot be ruled out analytically. 
We therefore study the effective dimension of $\mathcal{M}=\mathcal{E}_w(\Phi)$ looking at the Jacobian of the dosing network, $J_{\mathcal{E}_w}(\phi)=\partial\mathcal{E}_w(\phi)/\partial\phi$. The rank of
$J_{\mathcal{E}_w}(\phi)$ is the number of independent directions in which the regimen changes when $\phi$ is
perturbed, that is, the local dimension of $\mathcal{M}$ around $\mathcal{E}_w(\phi)$. In practice, the singular values
of $J_{\mathcal{E}_w}(\phi)$ are rarely exactly zero, but several of them can be orders of magnitude smaller than the
largest one. Along the corresponding directions the regimen barely changes, so we look at how fast the singular values
decay rather than at the exact rank.

\paragraph{Setup.} We use the coordinates of the online optimization (Appendix~\ref{app:bo}): $\phi$ is expressed in
the unit box of the $p=14$ PK/PD parameters searched by BO, and the regimen $\mathcal{E}_w(\phi)$ in the unit-box
coordinates of $\mathcal{D}$ given by the right inverse of \eqref{eq:decode}. Since this right inverse normalizes
$\mathbf{v}$ to sum to one, we drop $v_M$, so that the regimen has $D=2M-1=13$ coordinates and
$J_{\mathcal{E}_w}(\phi)\in\mathbb{R}^{13\times 14}$. We compute
$J_{\mathcal{E}_w}(\phi)$ by automatic differentiation.

\paragraph{Experiment.} We draw $10$ virtual patients. For each patient, we draw $10^5$ parameter vectors $\phi$ from $p(\phi)$ given the
patient's covariates. For each of the
$10^6$ draws, we compute the singular values $\sigma_1(\phi)\geq\dots\geq\sigma_{13}(\phi)$ of
$J_{\mathcal{E}_w}(\phi)$, i.e.\ the square roots of the eigenvalues of
$J_{\mathcal{E}_w}(\phi)J_{\mathcal{E}_w}(\phi)^\intercal$, and divide them by the largest one, $\sigma_1(\phi)$, to
remove the overall scale of the map.

\begin{figure}[t]
    \centering
    \includegraphics[width=0.6\linewidth]{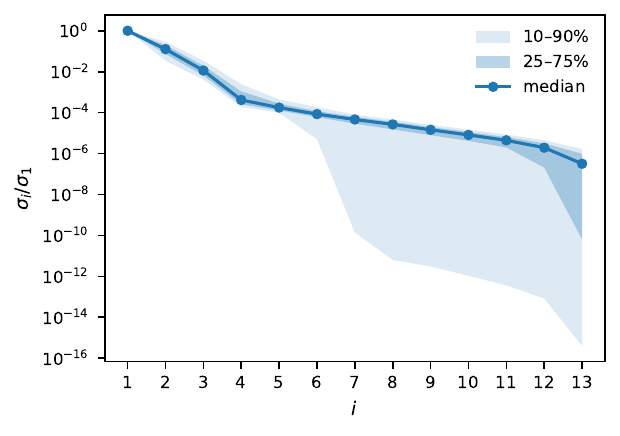}
    \caption{Singular values $\sigma_i$ of the Jacobian $J_{\mathcal{E}_w}(\phi)$ of the dosing network, divided by the
    largest one. Median, $25$--$75\%$ and
    $10$--$90\%$ quantiles; log-scale for the y-axis.}
    \label{fig:jacobian_spectrum}
\end{figure}

\paragraph{Results.} Figure~\ref{fig:jacobian_spectrum} shows the distribution of $\sigma_i(\phi)/\sigma_1(\phi)$ over
the draws. The median ratios are $0.13$ for $i=2$, $1.1\cdot10^{-2}$ for $i=3$ and $4.2\cdot10^{-4}$ for $i=4$, and
decrease slowly from $1.8\cdot10^{-4}$ to $3.2\cdot10^{-7}$ for $i=5,\dots,13$. The largest drop between two
consecutive median ratios is between $i=3$ and $i=4$, by a factor $28$. With a threshold of $10^{-3}$,
inside this range, the number of ratios above the threshold is $3$ for $73\%$ of the draws and $4$ for $26\%$. We
therefore estimate the intrinsic dimension of $\mathcal{M}$ between $3$ and $5$.

\section{Bayesian optimization with \texorpdfstring{$\pi$BOASS}{piBOASS}}
\label{app:pibo}

$\pi$BOASS uses the same amortized map $\mathcal{E}_w$ as BOASS, but does not restrict the search to its image $\mathcal{M}$. The optimization runs on $\mathcal{D}$, and $\mathcal{E}_w$ is used instead to build a
simulator-based prior $\pi$ over the location of the optimum, combined with the acquisition function as in $\pi$BO \citep{hvarfner_pbo_2022}. The simulator thus enters the online loop as a prior that decays as observations
accumulate, rather than as a hard restriction of the search space. In this sense $\pi$BOASS is a soft
variant of our method, that we expect to be less sensitive to search-space bias.

\paragraph{Prior.} The prior is fitted once per patient and cached across iterations. We draw $N=10^4$
samples of the optimized PK/PD parameters from their population random-effect distribution and map each draw $\phi_s$
through the dosing network to a regimen $(\mathbf{d},\mathbf{t}) = \mathcal{E}_w(\phi_s)$. This sample is
the pushforward of the population distribution $p(\phi)$ through $\mathcal{E}_w$, that is, the distribution
of the regimens the network prescribes across the population compatible with the patient's covariates. 

\paragraph{Transformation.} The pushforward is supported on a bounded set with atoms at its boundary: a
fraction of the doses saturate at the single-dose cap $d_{\max}$, and a fraction of the intervals between
consecutive intakes sit at the minimum separation $\Delta$. Each regimen $(\mathbf{d}, \mathbf{t})$ is
therefore first mapped to unbounded coordinates
\begin{equation}
  \mathbf{v}
  = \Big(
      \underbrace{\operatorname{logit}\!\big(\epsilon_d + (1-2\epsilon_d)\, d_i / d_{\max}\big)}_{i = 1, \dots, M},\;
      \underbrace{\log\!\big(t_{j+1} - t_j - \Delta + s\big)}_{j = 1, \dots, M - 1}
    \Big) \in \mathbb{R}^{2 M - 1},
  \label{eq:pibo-transform}
\end{equation}
with $\epsilon_d = \Delta_\text{tab} / d_{\max}$ a squashing margin of one tablet increment and $s=1$~min a
fixed shift. The first intake carries no information, since $t_1 = t_0$ for every regimen, which leaves
$2M-1=13$ free coordinates.

\paragraph{Mixture fit.} We standardize the transformed samples per coordinate, $\mathbf{z} = (\mathbf{v} -
\hat{\boldsymbol{\mu}}) / \hat{\boldsymbol{\sigma}}$, and fit by expectation-maximization a Gaussian
mixture with $K$ full-covariance components,
\begin{equation}
  \pi(\mathbf{z}) = \sum_{k=1}^{K} w_k\, \mathcal{N}\!\big(\mathbf{z} \,;\, \boldsymbol{\mu}_k,\,
  \boldsymbol{\Sigma}_k + \varrho\, \mathbf{I}\big),
  \label{eq:pibo-prior}
\end{equation}
keeping the best of $10$ restarts. The two parameters $\varrho$ and $K$ are fixed for all patients and selected offline, before the
experiments. 
The floor $\varrho$, added to the covariance of every
component during fitting, bounds how narrow a component can become: we set $\varrho = 10^{-2}$, which corresponds to a standard deviation of one tenth of the
population spread of each coordinate. 
Since the density concentrates in lower-dimensional region, the regularization provided by $\varrho$ is crucial to prevent components from collapsing.

In order to select $K$, we consider a grid of possible values from $1$ to $30$ and select
$K$ by BIC, over the ten representative patients. We selected $K=9$, the median of the per-patient selections, so that the complexity of the
prior does not depend on the patient's own sample. 


\paragraph{Combination with the acquisition function.} At acquisition iteration $n$, the prior is combined
with the base acquisition function $\alpha$ as $\alpha(\mathbf{x})\, \big(\pi(\mathbf{x}) +
\epsilon_\pi\big)^{\gamma_n}$, evaluated in logarithmic form, with $\gamma_n = \beta / n$. The constant
$\epsilon_\pi = 10^{-12}$ added to the density is the one used in the original implementation
\citep[Appendix F.1]{hvarfner_pbo_2022}, and it helps preventing numerical issues when the prior density is extremely low.

\paragraph{Calibration of $\beta$.} The schedule $\gamma_n$ sets the weight of the prior relative to the
data-driven term: $\beta$ fixes the tilt at the first acquisition iteration, and the tilt decays as $1/n$
as observations accumulate. We calibrate $\beta$ on the same ten patients, measuring the average
log-density the prior assigns to a plausible regimen above its floor,
\begin{equation}
  g \;=\; \frac{1}{S}\sum_{s=1}^S\log \pi\big(\mathcal{E}_w(\phi_s)\big) - \log \epsilon_\pi,
  \label{eq:pibo-guidance}
\end{equation}
estimated on the pushforward samples $\phi_s \overset{\text{iid}}{\sim} p(\phi)$. We then consider the median of $\hat{g}$ over the ten patients. 
A plausible regimen is then favoured over an implausible one by $\gamma_n\, \hat{g} = \beta\,\hat{g}/ n$ units of log-acquisition at iteration $n$. We set that at $n=1$ the prior effect is $\kappa = 30$, giving $\beta = \kappa / \hat{g} = 0.808$, so that the prior dominates during the first iterations and its tilt becomes comparable to the range of the log-acquisition function after a few of them. The stronger setting $\beta = 10$ is the value used by
\citet{hvarfner_pbo_2022} in their experiments; with the prior fitted here it amounts to $\kappa \approx
370$, and the prior dominates over a much longer initial phase ($\pi$BOASS-strong).

\paragraph{Acquisition optimization.} Half of the restarts of the multi-start optimizer
used for optimizing the acquisition function are drawn from the fitted prior and the rest from a Sobol sequence, both filtered
to the feasible set \eqref{eq:feasible-set}. Drawing part of the restarts from the prior helps guide the optimizer towards regions favored by the prior, which can improve the efficiency of the search.

\section{Bayesian optimization with PCA-BO}
\label{app:pcabo}

PCA-BO \citep{raponi2020pcabo} runs BO on the leading principal components of the design space. In the original
method, the principal component analysis is refitted at every iteration on the evaluated points, weighted by their
rank. With a budget of a few tens of evaluations, we instead fit it once per patient, offline, on regimens returned
by the dosing network. PCA-BO-on-$\mathcal{D}$ thus uses the same simulator information as BOASS and $\pi$BOASS, but
restricts the search to a linear subspace of $\mathcal{D}$ rather than to the image $\mathcal{M}$ of
$\mathcal{E}_w$.

\paragraph{Regimen samples.}
For each patient, we draw $N = 10^4$ samples of the optimized PK/PD
parameters from their population random-effect distribution, given the patient's covariates, and map each draw
$\phi_s$ through the dosing network, before rounding to the tablet grid. The draws are not restricted to the search
box $\Phi$. Each regimen $\mathcal{E}_w(\phi_s)$ is expressed in the unit-box coordinates of $\mathcal{D}$
(Appendix~\ref{app:bo}), which gives points $x_s \in [0, 1]^{2M}$.

\paragraph{Principal components.}
Let $\bar{x}$ and $\widehat{\Sigma}$ be the empirical mean and covariance of $\{x_s\}_{s=1}^N$, and
$V = [v_1, \dots, v_r] \in \mathbb{R}^{2M \times r}$ the eigenvectors of $\widehat{\Sigma}$ with the $r$ largest
eigenvalues. We use $r = 3$, the lower end of our estimate of the intrinsic dimension of $\mathcal{M}$
(Appendix~\ref{app:intrinsic_dimension}); these components explain more than $99\%$ of the variance of the samples for
every patient.

\paragraph{Search space.}
BO searches the scores $u \in [0, 1]^r$ of the $r$ components, rescaled so that the unit box spans the range of the
samples: with $s_j(x) = v_j^\top (x - \bar{x})$, $\ell_j = \min_s s_j(x_s)$ and $h_j = \max_s s_j(x_s)$, a search point
$u$ encodes
\begin{equation}
    x(u) = \bar{x} + \sum_{j=1}^{r} \big(\ell_j + u_j\,(h_j - \ell_j)\big)\, v_j ,
    \label{eq:pca_decode}
\end{equation}
the other components being fixed at their mean. The point $x(u)$ is decoded into a regimen with
equation~\ref{eq:decode}, the dose coordinates clipped to $[0, 1]$, and rounded to the tablet grid before administration.

\paragraph{Surrogate and acquisition.}
The GP is defined on the scores $u$, with a Mat\'ern-$5/2$ kernel with ARD and
$\Gamma(2, 2)$ priors on the lengthscales. The acquisition functions, their optimization, the dose constraints of
equation~\ref{eq:feasible-set}, evaluated on the regimen encoded by $x(u)$, and the posterior mean maximization are
those of BO-on-$\mathcal{D}$ (Appendix~\ref{app:bo}).

\paragraph{Initialization.}
The three initial regimens are shared with the other methods and administered unchanged. Since they do not belong
exactly to the subspace, the GP is given their projections,
$u_j = \big(s_j(x) - \ell_j\big) / (h_j - \ell_j)$, as inputs.

\section{MIPD baseline}
\label{app:mipd_eig}
Model-informed precision dosing (MIPD) is the standard model-based way to individualize a dose: a population PK/PD model combined with an observation model defines a likelihood function, the distribution of the PK/PD parameters in the population gives a prior on the patient's parameters, updated from the patient's observations, and the dose is chosen from the updated model. In our setting the observation is the daily feedback, not a drug concentration. The MIPD baseline
replaces the Gaussian-process belief of BOASS with a posterior over the patient's PK/PD parameters. Everything else is shared with BO: the three initialization days (two quasi-random posologies, then the population-mean prescription) and their feedback, the dose constraints, the dosing network that maps parameters to a posology, the oracle and the regrets. MIPD has two arms, which differ only in how they choose the next dosage: \emph{EIG} (most
informative posology) and \emph{greedy} (best predicted dosage).

\subsection{Parameters and prior}

The 14 searched PK/PD parameters are written through the standardized random effects $z\in\R^{14}$ of the population
model: $\theta_j(z) = g_j^{-1}\bigl(\mu_j + \omega_j z_j\bigr)$, with $g_j$ a log or logit link, $\mu_j$ the patient's
covariate-adjusted typical value and $\omega_j$ the inter-individual variability. The other six parameters stay at their
covariate-adjusted population values, as in the BO arms. The prior is the population model $z \sim \mathcal{N}(0, I_{14})$.

\subsection{Observation model}

\paragraph{Score.} Let $c_e(\tau;\theta,d)$ be the effect-site concentration of dosage $d$ (doses and intake times)
over the day, $[\tau_0,\tau_N]$, and $c^*(\theta)$ the concentration giving the target effect. MIPD scores a dosage
by minus its time-averaged absolute error to the target,
\begin{equation}\label{eq:mipd_score}
v(d;\theta) = -\frac{1}{\tau_N-\tau_0}\int_{\tau_0}^{\tau_N} \mid c_e(\tau;\theta,d)-c^*(\theta)\mid\,d\tau ,
\end{equation}

which coincide with the way the feedback is simulated, therefore the model is well-specified.

The score depends on $z$ through seven quantities only: the six levodopa PK parameters (the fraction $f_a$ of the dose
absorbed through the fast route, the mean transit times MTT$_{\text{fast}}$ and MTT$_{\text{slow}}$ of the two
absorption routes, central volume $V_c$, clearance $CL$ and the effect-site equilibration time $T_{k_{e0}}$) and $c^*$.
The carbidopa parameters do not affect $c_e$, and EC$_{50}$, $\rho$ and $\gamma$ enter only through
$c^* = \mathrm{EC}_{50}(\rho-1)^{-1/\gamma}$ (BASE drops out). Along the other directions the posterior is the prior.

\paragraph{Concentration feedback.} The feedback $y_t$ of day $t$ is modelled as an unknown affine function of the
score plus Gaussian noise,
\[
  y_t = a\,v\bigl(d_t;\theta(z)\bigr) + b + \varepsilon_t, \qquad \varepsilon_t \sim \mathcal{N}(0, s_y^2\sigma^2),
\]
with $s_y$ the standard deviation of the feedback observed so far: the noise is expressed on the standardized
feedback, as BoTorch standardizes the outcomes of the GP arms. The priors are
\begin{itemize}
  \item $\log a \sim \mathcal{N}(0,1)$: the feedback increases with the score; 90\% of the prior mass is on $a\in[0.19, 5.2]$;
  \item $b$ flat: a location parameter of unknown units. Integrated out, it leaves only the day-to-day variation of
    the feedback to inform $z$, with $n-1$ degrees of freedom for $n$ days;
  \item $\log\sigma^2 \sim \mathcal{N}(-4, 1)$: the default noise prior of BoTorch's \texttt{SingleTaskGP}, i.e.\ the noise
    prior of the GP arms. MIPD has the same information about the noise as BOASS.
\end{itemize}
The nuisance parameters $(a,b,\sigma^2)$ are integrated out: $b$ in closed form and $(\log a,\log\sigma^2)$ by
quadrature on a uniform grid ($1201\times241$ points over $\pm6$ prior standard deviations).

\paragraph{Preference feedback.} From the second day on, the simulator reports whether today's dosage beats
yesterday's, $c_t\in\{\pm1\}$, and MIPD (like the preference BO arms) sees only these comparisons. They are modelled
by a probit on the score difference,
\[
  P(c_t=+1\mid z,\kappa) = \Phi\bigl(\kappa\,[v(d_t;\theta(z)) - v(d_{t-1};\theta(z))]\bigr),
  \qquad \log\kappa\sim\mathcal{N}(0,1),
\]
with $\kappa$ integrated out on a grid of 101 values of $\log\kappa$. The level of the score cancels in the
difference, and $\kappa$ absorbs its scale and the comparison noise.

\subsection{Inference}

With the nuisance parameters integrated out, the posterior of $z$ is
$\log p(z\mid\text{data}) = \log p(\text{data}\mid z) - \tfrac12\lVert z\rVert^2 + \text{const}$. Each day MIPD computes
\begin{itemize}
  \item the MAP $\hat{z}_t$, by L-BFGS-B, started from the previous day's MAP and from the prior mode $z=0$ (the better
    one is kept). Gradients come from automatic differentiation through the exact solution of the PK ODE, which is
    linear between intakes (matrix exponential);
  \item the Laplace precision $\Lambda_t$: the prior precision plus the expected Fisher information of the observed
    days at $\hat{z}_t$ (Gauss--Newton). With $g_s = \nabla_z v(d_s;\theta(\hat{z}_t))$,
    \[
      \Lambda_t = I + \hat w \sum_{s} (g_s-\bar g)(g_s-\bar g)^\top
      \quad\text{(concentration)},\qquad
      \Lambda_t = I + \sum_s \bar\omega(\Delta v_s)\,h_s h_s^\top
      \quad\text{(preferences)},
    \]
    where $\hat w = E[a^2/(s_y^2\sigma^2)]$ under the posterior of the nuisances, $\bar g$ is the mean gradient, $\Delta v_s$ and $h_s$ are the score difference of a
    comparison (winner minus loser) and its gradient, and
    $\bar\omega(\Delta) = E_\kappa\bigl[\kappa^2\varphi(\kappa\Delta)^2/(\Phi(\kappa\Delta)\Phi(-\kappa\Delta))\bigr]$
    is the Fisher information of a probit observation averaged over the posterior of $\kappa$.
\end{itemize}

\subsection{Best guess and next candidate}

\paragraph{Best guess.} The dosing network's dosage for $\theta(\hat{z}_t)$, reported from the same day as the BO arms'
best guess (day 2). The network minimizes MIPD's own score, so the best guess is consistent with MIPD's model.

\paragraph{Candidates.} Each day, 8192 quasi-random points of the dosage box (doses and intake times, the domain of
BO on dosages) are rounded to administrable dosages, and those satisfying the dose constraints (total daily dose
within $\pm40\%$ of today's and at least 320\,mg) are kept.

\paragraph{EIG.} The candidate with the largest expected information gain about $z$, linearized at the MAP (local
Bayesian D-optimal design). An observation at $d$ adds a rank-one term to the precision, so the gain in
log-determinant is
\[
  \alpha_t(d) = \tfrac12\log\bigl(1 + w(d)\,h(d)^\top\Lambda_t^{-1}h(d)\bigr),
\]
with $h(d) = g(d)-\bar g$ and $w = \hat w\,n/(n+1)$ for concentration feedback, and, for preferences, where tomorrow's
posology is compared with today's $d_t$, $h(d) = g(d)-g(d_t)$ and $w=\bar\omega\bigl(v(d)-v(d_t)\bigr)$.

($\arg\max_d v$ with preferences), the best guess itself included. This is how MIPD is used in the clinic (Bayesian
forecasting): it doses to treat, and learns only from what it administers.

\subsection{Likelihood misspecification}

\begin{figure}
    \centering
    \includegraphics[width=\linewidth]{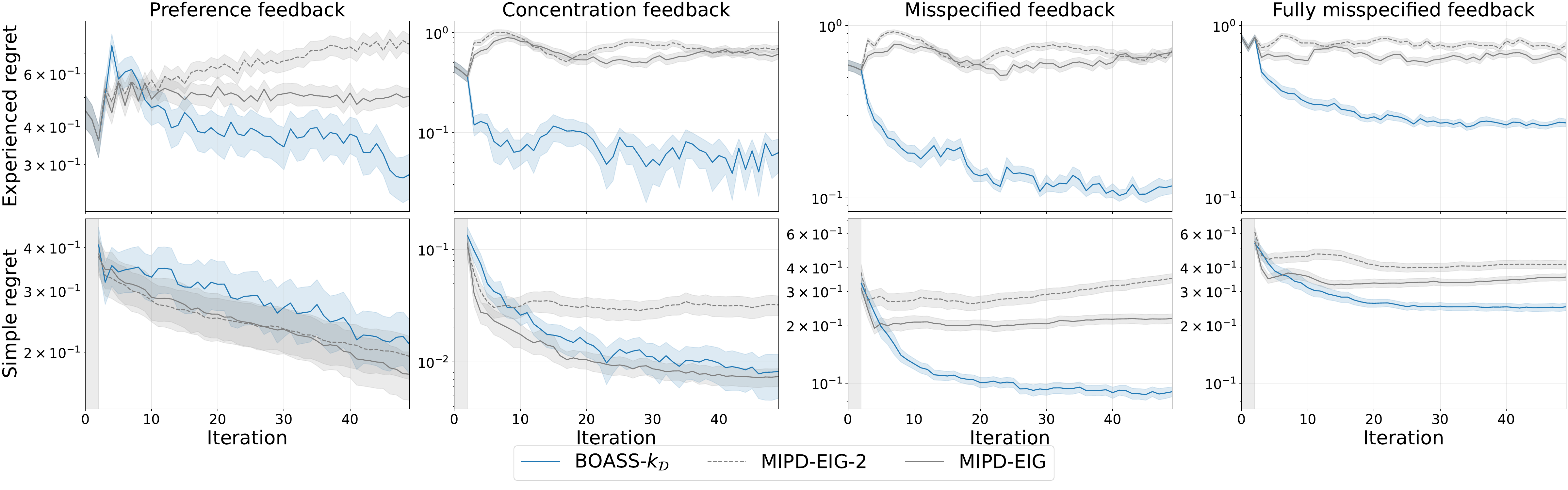}
    \caption{Performance comparison between well-specified score (absolute error on concentration) and misspecified score (squared error on concentration) in the likelihood of the MIPD model.}
    \label{fig:mipd_comparison}
\end{figure}

We implement a variant of the above MIPD model (MIPD-EIG-2), where the score that enters in the likelihood is the squared error instead of the absolute error in \eqref{eq:mipd_score}. This creates a small miss-specification in the likelihood. Figure \ref{fig:mipd_comparison} shows the comparison. While on preference feedback the difference in performance is negligible, on the single outcome settings this bias strongly affects the performance.

\section{Supplementary results for drug dosing optimization}
\label{app:low_first_dose}

\subsection{Low first-dose counting}

For the two well-specified settings in Section \ref{sec:results}, we also count the iterations at which the tested regimen has a first
dose smaller than the maintenance doses (\emph{low-first-dose}); Table~\ref{tab:first_dose_lowest} reports the
counts. A low first dose is a sufficient condition for a regimen to lie outside $\mathcal{M}$.
When the oracle follows the PK/PD model, this provides a proxy criterion to identify a subset of the
pharmacologically incoherent regimens, that is, regimens a clinician would likely not prescribe. By construction, our
method never tests such regimens, while the other methods test a small but non-zero number of them. All methods are initialised in $\mathcal{M}$.

\begin{table}[htbp]
\caption{Median [10th, 90th percentile] number of acquisition days (out of 47) on which the first prescribed dose is strictly lower than all other doses, per method and well-specified setting for the Levodopa-Carbidopa use-case in Section \ref{sec:levodopa}}
\label{tab:first_dose_lowest}
\centering
\small
\begin{tabular}{lrrrr}
\toprule
 & BOASS-* & $\pi$BOASS-on-$\mathcal{D}$ & $\pi$BOASS-on-$\mathcal{D}$-strong & BO-on-$\mathcal{D}$ \\
\midrule
preference feedback & 0.0 [0.0, 0.0] & 0.0 [0.0, 2.0] & 0.0 [0.0, 0.0] & 1.0 [0.0, 13.1] \\
concentration feedback & 0.0 [0.0, 0.0] & 0.0 [0.0, 4.0] & 0.0 [0.0, 3.1] & 1.5 [0.0, 7.0] \\
\bottomrule
\end{tabular}

\end{table}

\subsection{Computational time}

In order to compare the computational costs between the different search space optimization, we timed BOASS-$k_\mathcal{D}$, BOASS-$k_\Phi$, BO-on-$\mathcal{D}$, and PCA-BO-on-$\mathcal{D}$ on 5 patients with concentration feedback, over the 50 days
of the experiment. All methods ran in a single process, one after the other, after an untimed warm-up run. We report
the wall-clock time per patient of the BO loop: the setup, the daily GP fits and the next-candidate optimizations.
The computation of the best guess and the patient simulation are not timed. We also report the time of the optimization
of the acquisition function after the initialization days. All timings were measured on a MacBook Pro with an Apple M4 CPU
(10 cores: 4 performance, 6 efficiency) and 24\,GB of memory. Results are in Table~\ref{tab:timing_bo_loop}.

\begin{table}[t]
\centering
\caption{Wall time in seconds, mean (std), with concentration feedback on an Apple M4. BO loop: setup, GP fits and
next candidates over 50 days, per patient (5 patients). Acquisition: one next-candidate optimization.}
\label{tab:timing_bo_loop}
\begin{tabular}{lrrr}
\toprule
Method & BO loop (s) & Acquisition (s) \\
\midrule
BOASS-$k_\mathcal{D}$ & 89.56 (11.39) & 1.72 (1.39) \\
BOASS-$k_\Phi$ & 12.73 (2.63)& 0.16 (0.12) \\
BO-on-$\mathcal{D}$ & 16.35 (5.49) & 0.20 (0.59) \\
PCA-BO-on-$\mathcal{D}$ & 8.80 (0.94) & 0.14 (0.12) \\
\bottomrule
\end{tabular}
\end{table}

All methods choose the next query within seconds. With one dosage chosen per day,
computation is not a limitation for any method. PCA-BO-on-$\mathcal{D}$ is the fastest, as it searches the smallest
space (3 principal components), which shortens both its GP fits and its acquisition optimization. BOASS-$k_\Phi$ and
BO-on-$\mathcal{D}$ take similar times, as they search spaces of almost the same dimension: 14 PK/PD parameters and
13 dose degrees of freedom (7 doses and 6 dosing intervals). BOASS-$k_\mathcal{D}$ is the slowest, almost
entirely because of its acquisition optimization. Each evaluation of its kernel at a
candidate requires a forward pass of the dosing network.

\section{Synthetic experiments}
\label{app:synthetic_experiments}

Section~\ref{sec:regret} splits the simple regret of BOASS into two terms: the online optimization error, that is the gap between the best point found so far and the best point of the amortized search space $\mathcal{M}$, and the search-space bias, that is the gap between the best point of $\mathcal{M}$ and the best point of $\mathcal{D}$.  The synthetic benchmarks below control them separately, and vary one ingredient at a time.

We use two benchmarks with complementary roles.
\begin{itemize}
  \item \textbf{Branin embedded in an ambient space} (Section~\ref{sec:synth-branin}). The ambient objective, its optimum and the exact value of the search-space bias are known in closed form. We study the followings:
  \begin{itemize}
      \item We vary the ambient dimension $D$ and the search space bias $\delta$, and measure how the gain obtained by searching a two-dimensional space compares with the cost of the bias at a fixed evaluation budget;
      \item We train a NN with different sample size to induce different degrees of amortization error, measuring the impact in terms of regret (Section \ref{sec:synth-branin-learned});
  \end{itemize}
  \item \textbf{Objective drawn from a Gaussian process} (Section~\ref{sec:synth-gp}). The objective is a sample from the prior used by the surrogate, so the GP model is well-specified in the ambient space, and the reference value is the best value found on $\mathcal{M}$, so no search-space bias enters. What varies is the geometry of the embedding: the intrinsic dimension $p$, the ambient dimension $D$, and how far the embedding is from being injective. This is the setting in which the two kernel choices of Section~\ref{sec:pullback-vs-standard-kernel} differ, so this is the benchmark we use to compare them.
\end{itemize}

Table~\ref{tab:methods-synth} lists the methods used across all benchmarks. The Branin benchmark compares BOASS-$k_{\Phi}$ with BO-on-$\mathcal{D}$ and high-dimensional BO methods (TuRBO \citep{eriksson19a_turbo}, SAASBO \citep{eriksson21a_sassbo}), while the Gaussian process benchmark compares the three restricted methods.

\begin{table}[t]
\caption{Methods compared in the synthetic experiments.}
\label{tab:methods-synth}
\centering
\begin{tabular}{p{0.25\linewidth}p{0.29\linewidth}p{0.37\linewidth}}
\toprule
Method & Search space and GP input & Role \\
\midrule
BO-on-$\mathcal{D}$
  & Search $x\in\mathcal{D}$; GP input is $x$
  & Unrestricted reference. \\
BOASS-$k_{\mathcal{D}}$
  & Search $\phi\in\Phi$; GP input is $\mathcal{E}(\phi)$
  & Proposed method with the pullback kernel $k_{\mathcal{D}}\bigl(\mathcal{E}(\phi),\mathcal{E}(\phi')\bigr)$. \\
BOASS-$k_{\Phi}$
  & Search $\phi\in\Phi$; GP input is $\phi$
  & Same search space, but the kernel is defined on the simulator parameters and ignores the induced geometry. \\
BOASS-random-$k_{\Phi}$
  & Draw $\phi$ uniformly and evaluate $\mathcal{E}(\phi)$
  & Separates the benefit of restricting the search space from the benefit of the acquisition function. \\
TuRBO & Search $x\in\mathcal{D}$; GP input is $x$
  & Unrestricted high-dimensional BO reference \citep{eriksson19a_turbo} \\
SAASBO & Search $x\in\mathcal{D}$; GP input is $x$
  & Unrestricted high-dimensional BO reference \citep{eriksson21a_sassbo} \\
\bottomrule
\end{tabular}
\end{table}

\subsection{Branin embedded in an ambient space}
\label{sec:synth-branin}

\paragraph{Objective on the surface.}
The construction embeds the two-dimensional Branin function $B$ in an ambient decision space of dimension $D$.  The intrinsic parameter is $\phi=(\phi_1,\phi_2)\in\Phi=[-1,1]^2$.  It is mapped to the conventional Branin domain by :

\begin{equation}
  T(\phi)=
  \begin{pmatrix}
    7.5\phi_1+2.5\\
    7.5(\phi_2+1)
  \end{pmatrix},
\end{equation}

so that $T_1(\phi)\in[-5,10]$ and $T_2(\phi)\in[0,15]$.  The global minimum is $B_{\min}\simeq0.397887$.  The implementation uses the shifted and scaled cost

\begin{equation}
  \widetilde B(\phi)=\frac{B(T(\phi))-B_{\min}}{50}.
  \label{eq:normalized-branin}
\end{equation}

Thus the minimum of $\widetilde B$ is zero.

\paragraph{Embedding.}
The ambient space is $\mathcal{D}=[-1,1]^D$ and the surface is the image of the ideal embedding $\mathcal{E}^\star(\phi)=(\phi,m(\phi))$, with $m:\Phi\to\mathbb{R}^{D-2}$ smooth. We use the fixed-geometry family
\begin{equation}
  m_D^{\mathrm{fix}}(\phi)
  =a\begin{pmatrix}
      \sin(\pi\phi_1)\\
      \cos(\pi\phi_2)\\
      \sin\!\bigl(\pi(\phi_1+\phi_2)\bigr)\\
      \cos\!\bigl(\pi(\phi_1-\phi_2)\bigr)\\
      0_{D-6}
    \end{pmatrix},
  \qquad a=0.15,
  \label{eq:fixed-geometry-features}
\end{equation}
defined for $D\geq6$. Increasing $D$ only appends zero coordinates, so the pairwise distances on the surface, and therefore the problem seen by a method restricted to it, do not change. Only the dimension seen by BO-on-$\mathcal{D}$ grows.

\paragraph{Search-space bias.}
The bias is introduced by moving the optimum away from the surface.
Let $e_1=(1,0,\ldots,0)^\top\in\mathbb R^{D-2}$. This vector selects the first of the $D-2$ auxiliary coordinates. Consequently, $\delta e_1=(\delta,0,\ldots,0)^\top$ shifts the target surface only along the ambient coordinate $x_3$.
We define the ambient minimization objective as:

\begin{equation}
f_\delta(x)
  =\widetilde B(x_{1:2})
   +\lambda\left\|x_{3:D}-m(x_{1:2})-\delta e_1\right\|_2^2,
  \qquad x\in[-1,1]^D,
  \label{eq:true-objective-shifted}
\end{equation}

with $\lambda=1$ and $m=m_D^{\mathrm{fix}}$.  The global optimum has again cost zero, but lies on the shifted surface $\{x:x_{3:D}=m(x_{1:2})+\delta e_1\}$. The simulator-defined ideal embedding remains $\mathcal{E}^\star(\phi)=(\phi,m(\phi))$, but the ideal surface $\mathcal{M}^\star$ does not
 contain this optimum when $\delta>0$, so its global simple regret has the exact floor $\Delta_{\mathrm{model}}=\lambda\delta^2$. The two terms of the regret are thus controlled separately: $D$ acts only on the problem faced by BO-on-$\mathcal{D}$, and $\delta$ only on the bias of the restricted search.

\paragraph{Common protocol.}
Observations are $y_t=f_\delta(x_t)+\eta_t$, with $\eta_t\sim\mathcal N(0,0.01^2)$. Unless stated otherwise, the surrogates use a Matérn-$5/2$ covariance with the noise variance held at its true value $\sigma^{2}=0.01^{2}$, and BOASS and BO-on-$\mathcal{D}$ propose the next evaluation by minimizing the lower confidence bound $\operatorname{LCB}_t(z)=\mu_{t-1}(z)-2\sigma_{t-1}(z)$. Since $f_{\delta}$ attains the value $0$ on $\mathcal{D}$, we report simple regret as the best noiseless value observed so far.

\subsubsection{Comparison with high-dimensional BO methods and bias analysis}
\label{sec:synth-branin-hdbo}

This experiment measures how the search-space bias trades off against the ambient dimension, and compares BOASS with BO methods designed for high-dimensional spaces. The embedding $\mathcal{E}^\star$ is injective, so two distinct $\phi$ never produce the same design and the ambiguity that the pullback kernel is meant to resolve does not arise here. We therefore report BOASS-$k_{\Phi}$ only, and leave the comparison between the two kernels to Section~\ref{sec:synth-gp}.

TuRBO~\citep{eriksson19a_turbo} and SAASBO~\citep{eriksson21a_sassbo} adapt Bayesian
optimization to high-dimensional spaces through different mechanisms. 
TuRBO-1 restricts the search to a single axis-aligned trust region, whose side lengths are proportional to the ARD lengthscales of a GP fit only to the local data; candidates are selected by Thompson sampling inside the region. 
The region size is adapted rather than monotonically shrunk: doubled after $\tau_{\mathrm{succ}}$ consecutive improvements, halved after $\tau_{\mathrm{fail}}$ failures, and restarted from a fresh design once it falls below $L_{\min}$. 
Thus, exploration comes from restarts rather than from the acquisition function. 
SAASBO instead keeps a single global GP and places a half-Cauchy hierarchical prior on the inverse squared lengthscales, shrinking most coordinates to irrelevance and concentrating the model on a sparse axis-aligned subspace inferred from data,
with hyperparameters sampled by NUTS.
Both methods search the full ambient space $\mathcal{D}$, so, like BO-on-$\mathcal{D}$, they are not subject to the bias floor $\lambda\delta^{2}$.

\paragraph{Setting.}
We use $D\in\{6,8,16,32\}$ and $\delta\in\{0,0.1,0.25,0.5,1\}$. Each method has a budget of $60$ evaluations, and each pair $(D,\text{method})$ is repeated over $25$ independent seeds. All methods start from a single random point of $\mathcal{M}$, common to all methods, obtained by drawing $\phi$ uniformly in $\Phi$ and mapping it by $\mathcal{E}^\star$.

\paragraph{Methods.}
BOASS-$k_{\Phi}$ and BO-on-$\mathcal{D}$ estimate the amplitude and the length scale of their kernel by maximum marginal likelihood. 
TuRBO-1 uses an ARD Matérn-$5/2$ kernel, whose length scales set the shape of the trust region, and Thompson sampling as acquisition function. We run it with two pairs of failure and success tolerances $(\tau_{\mathrm{fail}},\tau_{\mathrm{succ}})$. 
With TuRBO-1-Restart (tolerances $(3,10)$), the trust region converges after an average of 35 iterations, after which it restarts with a new initial random point on $\mathcal{M}$, while with TuRBO-1 (tolerances $(8,3)$), the region shrinks slowly and a run has no restart for $99.6\%$ of experiments when the budget is $60$.
SAASBO is fitted by NUTS and uses q-log expected improvement as acquisition function \citep{ament_unexpected_2023}.  
Random search on $\Phi$ draws $\phi$ uniformly and evaluates $\mathcal{E}^\star(\phi)$.

\paragraph{Results.}
Tables~\ref{tab:regret-delta-D} and~\ref{tab:regret-median-quantiles} give the simple regret after $60$ evaluations, as mean with standard error and as median with quantile range. Table~\ref{tab:regret-delta-D_main} in the main paper is a subset of Table~\ref{tab:regret-delta-D}, and reports only TuRBO-1, which was consistently better than TuRBO-1-Restart.
When $\delta=0$, the optimum lies on $\mathcal{M}$ and BOASS-$k_{\Phi}$ has the lowest regret in every dimension. When $\delta>0$, its regret cannot go below $\lambda\delta^{2}$, and the methods that search $\mathcal{D}$ pass it in low dimension. The dimension from which BOASS-$k_{\Phi}$ is ahead grows with $\delta$: $D\geq8$ for $\delta=0.1$, $D\geq16$ for $\delta=0.25$, and $D=32$ for $\delta=0.5$. At $\delta=1$, SAASBO has the lowest median regret in every dimension.

\begin{table}[t]
\centering
\footnotesize
\setlength{\tabcolsep}{2pt}
\caption{Effect of search-space bias and ambient dimension on the simple regret of BO methods in the optimization of the Branin function. Simple regret at iteration $60$, in units of $10^{-3}$: mean $\pm$ standard error of the mean over $25$ seeds. Blocks correspond to the five values of $\delta$. $\Delta_{\mathcal{M}}=\delta^{2}$ (in units of $10^{-3}$) is the smallest regret attainable by a method restricted to $\Phi$; a dash marks the methods that search $\mathcal{D}$ directly, for which no such bias applies. Within each block, \textbf{bold} marks, for each $D$ column, the lowest mean together with every method whose $\hat m \pm \mathrm{s.e.m.}$ interval overlaps that of the lowest.}
\label{tab:regret-delta-D}
\begin{tabular}{lcrrrr}
\toprule
Method & \multicolumn{1}{c}{$\Delta_{\mathcal{M}}$} & \multicolumn{1}{c}{$D=6$} & \multicolumn{1}{c}{$D=8$} & \multicolumn{1}{c}{$D=16$} & \multicolumn{1}{c}{$D=32$} \\
\midrule
\multicolumn{6}{l}{\emph{$\delta=0$}} \\
\addlinespace[2pt]
BOASS-$k_\Phi$ & $0$ & $\bm{6.3 \pm 2.5}$ & $\bm{6.3 \pm 2.5}$ & $\bm{6.3 \pm 2.5}$ & $\bm{6.3 \pm 2.5}$ \\
\rowcolor[gray]{0.92}BO-on-$\mathcal{D}$ & -- & $37.2 \pm 7.1$ & $74.7 \pm 18.0$ & $149.1 \pm 26.0$ & $687.5 \pm 120.0$ \\
SAASBO & -- & $\bm{9.0 \pm 2.0}$ & $21.6 \pm 3.7$ & $177.0 \pm 36.0$ & $489.7 \pm 72.0$ \\
\rowcolor[gray]{0.92}TuRBO-1 & -- & $\bm{7.7 \pm 1.6}$ & $31.2 \pm 7.6$ & $275.2 \pm 48.0$ & $609.8 \pm 99.0$ \\
TuRBO-1-Restart & -- & $57.4 \pm 17.0$ & $118.6 \pm 26.0$ & $437.5 \pm 90.0$ & $648.6 \pm 110.0$ \\
\rowcolor[gray]{0.92}Random search on $\Phi$ & $0$ & $12.1 \pm 2.3$ & $12.1 \pm 2.3$ & $12.1 \pm 2.3$ & $12.1 \pm 2.3$ \\
\midrule
\multicolumn{6}{l}{\emph{$\delta=0.1$}} \\
\addlinespace[2pt]
BOASS-$k_\Phi$ & $10$ & $16.3 \pm 2.5$ & $\bm{16.3 \pm 2.5}$ & $\bm{16.3 \pm 2.5}$ & $\bm{16.3 \pm 2.5}$ \\
\rowcolor[gray]{0.92}BO-on-$\mathcal{D}$ & -- & $40.8 \pm 7.5$ & $68.2 \pm 17.0$ & $150.7 \pm 24.0$ & $651.7 \pm 100.0$ \\
SAASBO & -- & $\bm{11.4 \pm 2.2}$ & $25.4 \pm 4.9$ & $150.0 \pm 46.0$ & $464.5 \pm 68.0$ \\
\rowcolor[gray]{0.92}TuRBO-1 & -- & $\bm{8.5 \pm 2.4}$ & $33.7 \pm 10.0$ & $333.4 \pm 58.0$ & $662.1 \pm 130.0$ \\
TuRBO-1-Restart & -- & $94.3 \pm 41.0$ & $120.5 \pm 26.0$ & $472.8 \pm 94.0$ & $730.1 \pm 140.0$ \\
\rowcolor[gray]{0.92}Random search on $\Phi$ & $10$ & $22.1 \pm 2.3$ & $22.1 \pm 2.3$ & $22.1 \pm 2.3$ & $22.1 \pm 2.3$ \\
\midrule
\multicolumn{6}{l}{\emph{$\delta=0.25$}} \\
\addlinespace[2pt]
BOASS-$k_\Phi$ & $62.5$ & $68.8 \pm 2.5$ & $68.8 \pm 2.5$ & $\bm{68.8 \pm 2.5}$ & $\bm{68.8 \pm 2.5}$ \\
\rowcolor[gray]{0.92}BO-on-$\mathcal{D}$ & -- & $41.5 \pm 11.0$ & $81.0 \pm 20.0$ & $154.7 \pm 22.0$ & $761.8 \pm 110.0$ \\
SAASBO & -- & $\bm{11.4 \pm 2.3}$ & $\bm{19.0 \pm 3.0}$ & $108.3 \pm 17.0$ & $534.0 \pm 59.0$ \\
\rowcolor[gray]{0.92}TuRBO-1 & -- & $\bm{8.3 \pm 1.6}$ & $28.5 \pm 6.2$ & $292.2 \pm 54.0$ & $636.7 \pm 110.0$ \\
TuRBO-1-Restart & -- & $40.7 \pm 9.5$ & $66.1 \pm 14.0$ & $504.3 \pm 86.0$ & $638.4 \pm 110.0$ \\
\rowcolor[gray]{0.92}Random search on $\Phi$ & $62.5$ & $74.6 \pm 2.3$ & $74.6 \pm 2.3$ & $74.6 \pm 2.3$ & $74.6 \pm 2.3$ \\
\midrule
\multicolumn{6}{l}{\emph{$\delta=0.5$}} \\
\addlinespace[2pt]
BOASS-$k_\Phi$ & $250$ & $256.3 \pm 2.5$ & $256.3 \pm 2.5$ & $256.3 \pm 2.5$ & $\bm{256.3 \pm 2.5}$ \\
\rowcolor[gray]{0.92}BO-on-$\mathcal{D}$ & -- & $48.1 \pm 8.2$ & $69.7 \pm 14.0$ & $\bm{184.1 \pm 31.0}$ & $853.8 \pm 92.0$ \\
SAASBO & -- & $\bm{8.8 \pm 2.0}$ & $\bm{22.2 \pm 4.2}$ & $\bm{140.1 \pm 24.0}$ & $637.2 \pm 57.0$ \\
\rowcolor[gray]{0.92}TuRBO-1 & -- & $\bm{7.3 \pm 1.2}$ & $40.3 \pm 8.0$ & $308.3 \pm 51.0$ & $764.7 \pm 110.0$ \\
TuRBO-1-Restart & -- & $33.7 \pm 9.4$ & $76.9 \pm 13.0$ & $502.7 \pm 78.0$ & $868.3 \pm 130.0$ \\
\rowcolor[gray]{0.92}Random search on $\Phi$ & $250$ & $262.1 \pm 2.3$ & $262.1 \pm 2.3$ & $262.1 \pm 2.3$ & $262.1 \pm 2.3$ \\
\midrule
\multicolumn{6}{l}{\emph{$\delta=1$}} \\
\addlinespace[2pt]
BOASS-$k_\Phi$ & $1000$ & $1006.0 \pm 2.5$ & $1006.0 \pm 2.5$ & $1006.0 \pm 2.5$ & $1006.0 \pm 2.5$ \\
\rowcolor[gray]{0.92}BO-on-$\mathcal{D}$ & -- & $52.5 \pm 9.9$ & $88.2 \pm 20.0$ & $201.2 \pm 35.0$ & $1250.0 \pm 65.0$ \\
SAASBO & -- & $\bm{13.7 \pm 2.3}$ & $\bm{22.1 \pm 3.4}$ & $\bm{125.0 \pm 20.0}$ & $\bm{705.2 \pm 64.0}$ \\
\rowcolor[gray]{0.92}TuRBO-1 & -- & $\bm{12.7 \pm 1.9}$ & $50.6 \pm 14.0$ & $602.4 \pm 66.0$ & $1190.0 \pm 93.0$ \\
TuRBO-1-Restart & -- & $37.4 \pm 11.0$ & $173.5 \pm 43.0$ & $926.8 \pm 130.0$ & $1333.0 \pm 120.0$ \\
\rowcolor[gray]{0.92}Random search on $\Phi$ & $1000$ & $1012.0 \pm 2.3$ & $1012.0 \pm 2.3$ & $1012.0 \pm 2.3$ & $1012.0 \pm 2.3$ \\
\bottomrule
\end{tabular}
\end{table}

\begin{table}[t]
\centering
\footnotesize
\setlength{\tabcolsep}{1pt}
\caption{Effect of search-space bias and ambient dimension on the simple regret of BO methods in the optimization of the Branin function. Simple regret at iteration $60$, in units of $10^{-3}$: median with [lower, upper] quantiles over $25$ seeds, to three digits. Blocks correspond to the five values of $\delta$. $\Delta_{\mathcal{M}}=\delta^{2}$ (in units of $10^{-3}$) is the smallest regret attainable by a method restricted to $\Phi$; a dash marks the methods that search $\mathcal{D}$ directly, for which no such bias applies. Within each block, \textbf{bold} marks the lowest median in each $D$ column.}
\label{tab:regret-median-quantiles}
\begin{tabular}{lcrrrr}
\toprule
Method & \multicolumn{1}{c}{$\Delta_{\mathcal{M}}$} & \multicolumn{1}{c}{$D=6$} & \multicolumn{1}{c}{$D=8$} & \multicolumn{1}{c}{$D=16$} & \multicolumn{1}{c}{$D=32$} \\
\multicolumn{6}{l}{\emph{$\delta=0$}} \\
\addlinespace[2pt]
BOASS-$k_\Phi$ & 0 & $\bm{0.11\,[0.01, 30.9]}$ & $\bm{0.11\,[0.01, 30.9]}$ & $\bm{0.11\,[0.01, 30.9]}$ & $\bm{0.11\,[0.01, 30.9]}$ \\
\rowcolor[gray]{0.92}BO-on-$\mathcal{D}$ & -- & $31.6\,[5.00, 101]$ & $35.0\,[10.8, 217]$ & $100\,[49.6, 294]$ & $583\,[116, 1450]$ \\
SAASBO & -- & $6.51\,[2.16, 19.1]$ & $12.0\,[5.15, 43.4]$ & $93.0\,[33.5, 447]$ & $460\,[116, 882]$ \\
\rowcolor[gray]{0.92}TuRBO-1 & -- & $4.35\,[2.03, 20.3]$ & $18.6\,[4.76, 53.0]$ & $175\,[56.7, 707]$ & $446\,[104, 1300]$ \\
TuRBO-1-Restart & -- & $26.6\,[9.82, 108]$ & $60.2\,[26.4, 328]$ & $280\,[71.2, 1170]$ & $489\,[97.3, 1380]$ \\
\rowcolor[gray]{0.92}Random search on $\Phi$ & 0 & $8.90\,[2.54, 23.7]$ & $8.90\,[2.54, 23.7]$ & $8.90\,[2.54, 23.7]$ & $8.90\,[2.54, 23.7]$ \\
\midrule
\multicolumn{6}{l}{\emph{$\delta=0.1$}} \\
\addlinespace[2pt]
BOASS-$k_\Phi$ & $10$ & $10.1\,[10.0, 40.9]$ & $\bm{10.1\,[10.0, 40.9]}$ & $\bm{10.1\,[10.0, 40.9]}$ & $\bm{10.1\,[10.0, 40.9]}$ \\
\rowcolor[gray]{0.92}BO-on-$\mathcal{D}$ & -- & $32.1\,[5.98, 95.8]$ & $36.9\,[8.05, 176]$ & $107\,[70.4, 225]$ & $519\,[121, 1290]$ \\
SAASBO & -- & $8.58\,[2.86, 30.8]$ & $11.9\,[6.06, 55.1]$ & $85.1\,[33.1, 332]$ & $378\,[126, 1010]$ \\
\rowcolor[gray]{0.92}TuRBO-1 & -- & $\bm{4.01\,[1.39, 15.5]}$ & $17.2\,[5.25, 54.7]$ & $295\,[43.4, 835]$ & $457\,[115, 1430]$ \\
TuRBO-1-Restart & -- & $17.0\,[6.67, 247]$ & $79.2\,[17.3, 285]$ & $352\,[92.0, 1060]$ & $443\,[84.4, 1710]$ \\
\rowcolor[gray]{0.92}Random search on $\Phi$ & $10$ & $18.9\,[12.5, 33.7]$ & $18.9\,[12.5, 33.7]$ & $18.9\,[12.5, 33.7]$ & $18.9\,[12.5, 33.7]$ \\
\midrule
\multicolumn{6}{l}{\emph{$\delta=0.25$}} \\
\addlinespace[2pt]
BOASS-$k_\Phi$ & $62.5$ & $62.6\,[62.5, 93.4]$ & $62.6\,[62.5, 93.4]$ & $\bm{62.6\,[62.5, 93.4]}$ & $\bm{62.6\,[62.5, 93.4]}$ \\
\rowcolor[gray]{0.92}BO-on-$\mathcal{D}$ & -- & $31.4\,[3.12, 77.0]$ & $36.8\,[8.18, 238]$ & $92.9\,[66.3, 283]$ & $583\,[169, 1540]$ \\
SAASBO & -- & $5.90\,[2.00, 27.0]$ & $\bm{14.0\,[6.23, 37.7]}$ & $89.2\,[33.2, 190]$ & $522\,[179, 897]$ \\
\rowcolor[gray]{0.92}TuRBO-1 & -- & $\bm{3.96\,[2.53, 21.6]}$ & $16.4\,[6.27, 64.9]$ & $183\,[56.6, 658]$ & $422\,[145, 1450]$ \\
TuRBO-1-Restart & -- & $14.1\,[6.37, 102]$ & $31.9\,[17.4, 141]$ & $369\,[132, 1020]$ & $499\,[125, 1430]$ \\
\rowcolor[gray]{0.92}Random search on $\Phi$ & $62.5$ & $71.4\,[65.0, 86.2]$ & $71.4\,[65.0, 86.2]$ & $71.4\,[65.0, 86.2]$ & $71.4\,[65.0, 86.2]$ \\
\midrule
\multicolumn{6}{l}{\emph{$\delta=0.5$}} \\
\addlinespace[2pt]
BOASS-$k_\Phi$ & $250$ & $250\,[250, 281]$ & $250\,[250, 281]$ & $250\,[250, 281]$ & $\bm{250\,[250, 281]}$ \\
\rowcolor[gray]{0.92}BO-on-$\mathcal{D}$ & -- & $33.5\,[5.08, 98.9]$ & $47.8\,[9.19, 200]$ & $102\,[67.9, 449]$ & $710\,[345, 1490]$ \\
SAASBO & -- & $5.32\,[2.12, 25.0]$ & $\bm{15.2\,[5.48, 55.7]}$ & $\bm{86.8\,[49.8, 350]}$ & $546\,[359, 1080]$ \\
\rowcolor[gray]{0.92}TuRBO-1 & -- & $\bm{5.31\,[1.86, 15.0]}$ & $23.8\,[5.62, 105]$ & $227\,[110, 625]$ & $604\,[300, 1410]$ \\
TuRBO-1-Restart & -- & $22.3\,[4.06, 48.0]$ & $55.0\,[16.6, 168]$ & $410\,[133, 967]$ & $633\,[293, 1650]$ \\
\rowcolor[gray]{0.92}Random search on $\Phi$ & $250$ & $259\,[253, 274]$ & $259\,[253, 274]$ & $259\,[253, 274]$ & $259\,[253, 274]$ \\
\midrule
\multicolumn{6}{l}{\emph{$\delta=1$}} \\
\addlinespace[2pt]
BOASS-$k_\Phi$ & $1000$ & $1000\,[1000, 1030]$ & $1000\,[1000, 1030]$ & $1000\,[1000, 1030]$ & $1000\,[1000, 1030]$ \\
\rowcolor[gray]{0.92}BO-on-$\mathcal{D}$ & -- & $32.6\,[31.3, 109]$ & $35.7\,[21.8, 248]$ & $114\,[65.4, 448]$ & $1210\,[868, 1720]$ \\
SAASBO & -- & $\bm{8.19\,[2.91, 32.5]}$ & $\bm{14.5\,[7.95, 43.6]}$ & $\bm{86.1\,[53.5, 237]}$ & $\bm{682\,[393, 888]}$ \\
\rowcolor[gray]{0.92}TuRBO-1 & -- & $10.4\,[3.65, 26.3]$ & $29.7\,[14.0, 63.8]$ & $685\,[223, 994]$ & $1160\,[656, 1940]$ \\
TuRBO-1-Restart & -- & $17.0\,[3.51, 64.9]$ & $59.9\,[14.2, 396]$ & $1010\,[159, 1780]$ & $1120\,[688, 2110]$ \\
\rowcolor[gray]{0.92}Random search on $\Phi$ & $1000$ & $1010\,[1000, 1020]$ & $1010\,[1000, 1020]$ & $1010\,[1000, 1020]$ & $1010\,[1000, 1020]$ \\
\bottomrule
\end{tabular}

\end{table}

\begin{figure}[t]
    \centering
    \includegraphics[width=0.8\linewidth]{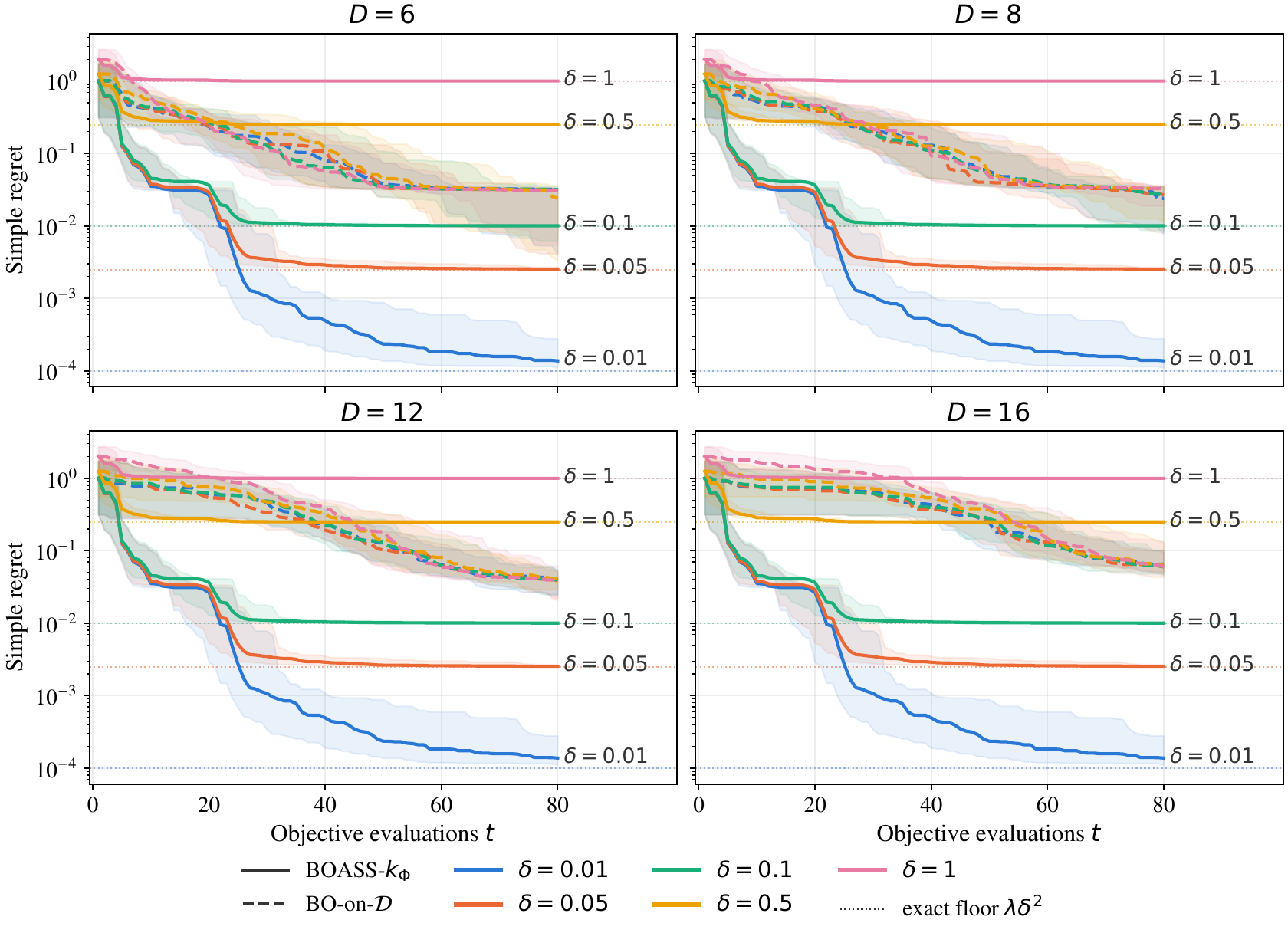}
    \caption{Effect of search-space bias and ambient dimension on the simple regret of BOASS in the optimization of the Branin function. Median simple regret of BOASS-$k_{\Phi}$ (solid lines) and BO-on-$\mathcal{D}$ (dashed lines), interquartile range highlighted. 
    Each plot corresponds to a dimension of $\mathcal{D}$. Within each plot, the results for the different bias parameters $\delta$ are plotted.}
    \label{fig:regret-curves-delta}
\end{figure}

\begin{figure}[t]
    \centering
    \includegraphics[width=0.8\linewidth]{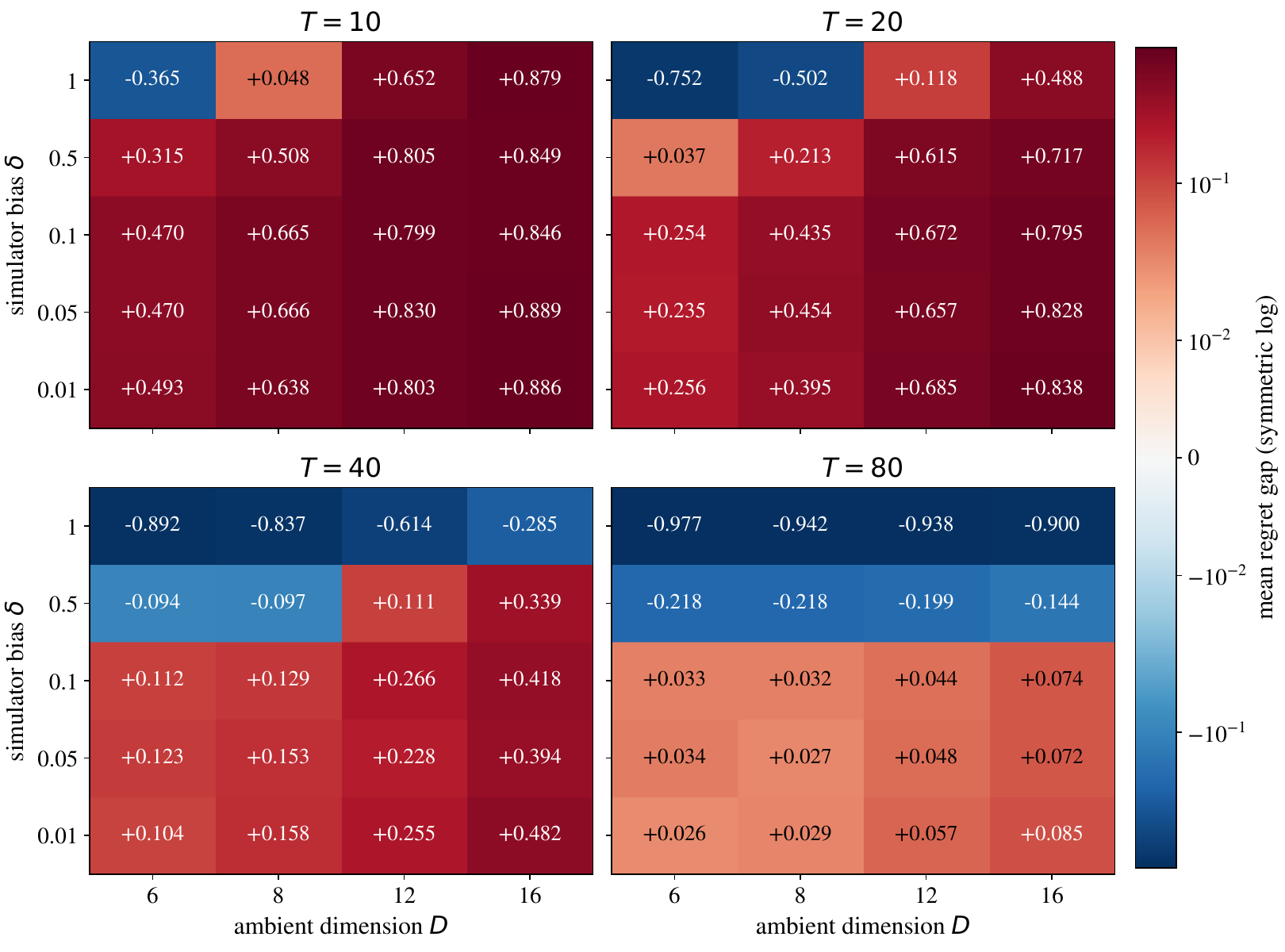}
    \caption{Comparison of simple regret in the optimization of the Branin function between BOASS-$k_{\Phi}$ and BO-on-$\mathcal{D}$, depending on search-space bias and ambient dimension.
    Each plot corresponds to a dimension of $\mathcal{D}$. Within each plot, each cell is the gap between the simple regret of BO-on-$\mathcal{D}$ and that of BOASS-$k_{\Phi}$ (higher means BOASS-$k_{\Phi}$ has a lower regret).}
    \label{fig:regret-gap-heatmaps}
\end{figure}

\paragraph{Regret curves.}
Figures~\ref{fig:regret-curves-delta} and~\ref{fig:regret-gap-heatmaps} follow BOASS-$k_{\Phi}$ and BO-on-$\mathcal{D}$ along the whole run, in a slightly different setting: $D\in\{6,8,12,16\}$, a finer grid of biases $\delta\in\{0.01,0.05,0.1,0.5,1\}$, a budget of $80$ evaluations, and $50$ paired seeds. 
All models estimate the amplitude and lengthscale, fixing one lengthscale common to all dimensions.
The floor $\lambda\delta^{2}$ is the horizontal asymptote of every BOASS-$k_{\Phi}$ curve, drawn as a dotted line in Figure~\ref{fig:regret-curves-delta}. Since the restricted problem does not depend on $D$, the BOASS-$k_{\Phi}$ curves are the same in all panels, and all differences between panels come from BO-on-$\mathcal{D}$. Figure~\ref{fig:regret-gap-heatmaps} reports, for $T\in\{10,20,40,80\}$, the mean over seeds of $r^{\text{BO-on-}\mathcal{D}}_{T}-r^{\text{BOASS-}k_{\Phi}}_{T}$, positive when BOASS-$k_{\Phi}$ is ahead.
For $\delta\leq0.1$, BOASS-$k_{\Phi}$ is ahead at every budget and every dimension. At $\delta=0.5$, BO-on-$\mathcal{D}$ passes it between $5$ and $30$ evaluations for $D=6$, and between $20$ and $40$ for $D=8$. This is the trade-off of Section~\ref{sec:regret}: the restriction pays as long as $\lambda\delta^{2}$ stays below the regret BO-on-$\mathcal{D}$ can reach within the budget.

\subsubsection{Learned embedding and amortization error}
\label{sec:synth-branin-learned}

In the previous experiments, the BO search space was defined using the optimal embedding $\mathcal{E}^\star$.  However, our BOASS method assumes that the embedding is learned using a neural network (Algorithm  \ref{app:algorithm}). To study the effect of amortization error due, e.g., to poor training of the network, we train a neural network  $\widehat{\mathcal{E}}$  to learn the embedding  with varying amounts of training data.

\paragraph{Setting.}
The objective is $f_\delta$, as defined in Section~\ref{sec:synth-branin}, with $D=10$ and $\delta=0.1$, so a search restricted to $\mathcal{M}^\star$ has the floor $\lambda\delta^{2}=0.01$.
We vary the size of the training set, $n_{\mathrm{train}}\in\{25,50,200,1000\}$.

For each replicate, we draw $n_{\mathrm{train}}$ Sobol points in $\Phi$ and fit a multilayer perceptron to the full ideal embedding $\mathcal{E}^\star(\phi)=(\phi,m(\phi))\in\mathbb{R}^{10}$. 
The network has two hidden layers of $128$ units with $\tanh$ activations. It is trained with Adam (learning rate $10^{-3}$, weight decay $10^{-5}$), with early stopping on a $15\%$ validation split. Its outputs are clipped to $\mathcal{D}=[-1,1]^{10}$. 

\paragraph{Methods and protocol.}
We compare four methods. BOASS-$k_{\mathcal{D}}$, BOASS-$k_{\Phi}$, BOASS-$k_{\mathcal{D}}$-oracle, and BO-on-$\mathcal{D}$.
BOASS-$k_{\mathcal{D}}$ and BOASS-$k_{\Phi}$ differ only in the GP input, $\widehat{\mathcal{E}}(\phi)$ or $\phi$ respectively, and evaluate the objective at $\widehat{\mathcal{E}}(\phi)$. 
BOASS-$k_{\mathcal{D}}$-oracle uses $\mathcal{E}^\star$ instead, and BO-on-$\mathcal{D}$ is the unrestricted BO reference.
 
The surrogates, the noise and the LCB acquisition are as in Section~\ref{sec:synth-branin}.  The optimization budget is $80$ iterations, with the acquisition minimized over $4096$ candidates per step. Each configuration is repeated over $20$ seeds. To each seed corresponds  a trained network, the initial candidate,  and the noise.

\paragraph{Results.}

\begin{figure}[t]
    \centering
    \includegraphics[width=0.8\linewidth]{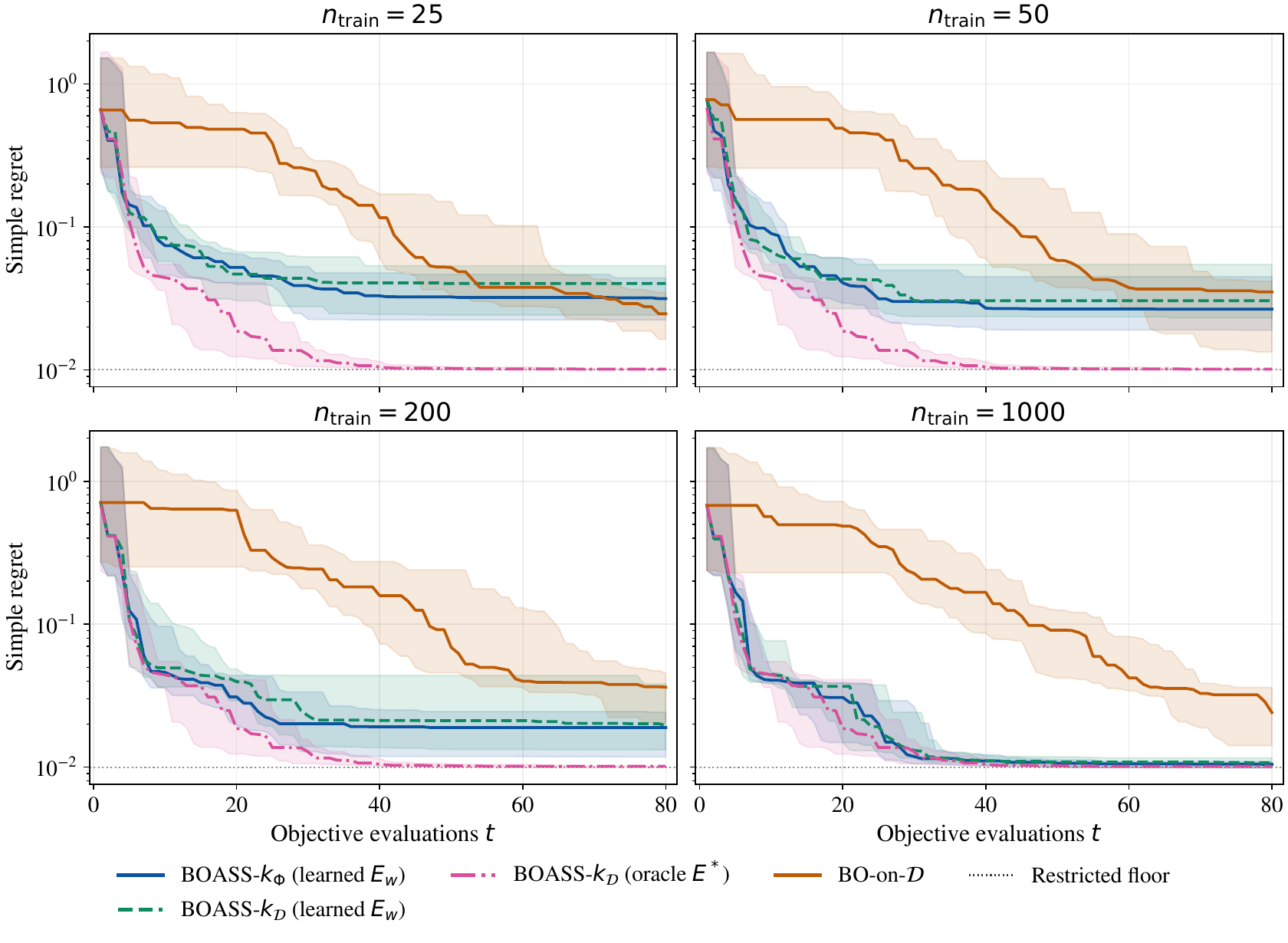} 
    \caption{Effect of amortization error on the simple regret of BOASS during the optimization of the Branin function, with a learned embedding $\widehat{\mathcal{E}}$. Lines correspond to the median simple regret, interquartile range highlighted. Each panel corresponds to a training-set size $n_{\mathrm{train}}$. The dotted line is the floor $\lambda\delta^{2}$ of the search on $\mathcal{M}^\star$.}
    \label{fig:regret-learned-embedding}
\end{figure}

Figure~\ref{fig:regret-learned-embedding} shows the median simple regret and interquartile range, one panel per $n_{\mathrm{train}}$. The floor $\lambda\delta^{2}$ is drawn as a dotted line.  
The oracle and BO-on-$\mathcal{D}$ do not depend on $\widehat{\mathcal{E}}$, so they give similar curves at every $n_{\mathrm{train}}$ and serve as fixed references. The approximation error incurs larger regret for BOASS-$k_{\mathcal{D}}$ and BOASS-$k_{\Phi}$ compared to BOASS-$k_{\mathcal{D}}$-oracle , since they rely $\widehat{\mathcal{E}}(\phi)$ instead of $\mathcal{E}^\star(\phi)$.  
More precisely, for a small training set, both  BOASS-$k_{\mathcal{D}}$ and BOASS-$k_{\Phi}$ it a regret lower bound that is higher than the floor $\lambda \delta^2$,  but still have better or similar performance compared to BO-on-$\mathcal{D}$. This can be explained by the gains from a restricted search space that compensate for the amortization bias, which is consistent with our regret analysis in Section \ref{sec:regret}. As the size of the training set increases, the amortization gap vanishes, so that the performance of BOASS-$k_{\mathcal{D}}$ and BOASS-$k_{\Phi}$ reaches the one of BOASS-$k_{\mathcal{D}}$-oracle.

\subsection{Objective drawn from a Gaussian process}
\label{sec:synth-gp}
\paragraph{Setting.} 
The Branin objective is not a sample from the GP prior that is used when fitting the GP regression surrogate model. To remove any effect that may be due to the surrogate GP kernel being misspecified, we therefore use a second benchmark in which the objective is drawn from the same Gaussian process prior the surrogate assumes.

 Let $\Phi = [-1, 1]^p$, and $\mathcal{D} = [-1, 1]^D$. Let $f\sim \text{GP}(0, k_D)$ be the objective function defined in $\mathcal{D}$.
We define
\begin{align}
\mathcal{E}_D:\Phi & \rightarrow \mathcal{D}\\
 \phi & \mapsto Wg(\phi)
\end{align}
where $g(\phi) = (\phi+1)(\phi-1)\sin{(a\pi\phi+b)}/2 + \phi$ acts coordinatewise, and $a, b$ are scalar hyperparameters.

The function $g$ maps $[-1, 1]$ onto (surjectively) $[-1, 1]$, and parameter $a$ dictates the level of non-injectivity of the function.
To ensure full rank of $W$ each row is sampled independently from a Dirichlet distribution, $W_i\sim \text{Dir}(\alpha=(1,\dots,1))$.
Since $W$ has full column rank almost surely, the non-injectivity of $\mathcal{E}_D$ comes from $g$ alone.

The kernel $k_D$ is a Matérn-$5/2$ kernel whose lengthscale is set to one half of the median pairwise distance of $\mathcal{E}_D(\Phi)$, so that $f_D$ does not become rougher as $D$ grows. Observations are corrupted by Gaussian noise $N(0, 0.01^2)$.

\paragraph{What we vary.} We run every combination of $p\in\{2,4,8\}$ and $D\in\{p,10,20\}$, with the linear embedding $\mathcal{E}_D(\phi)=W\phi$ and with $a\in\{0.5,0.7,0.9,1.1\}$. At each experiment we sample a random value of $b$, uniformly in $[0, \pi]$. The pair $(p,D)$ sets the dimension of the search space and the dimension of the design space it is mapped into, while $a$ sets how much the embedding folds, which we summarize by the fraction $\nu$ defined below. Note that $a=0.5$ makes $g$ curved while keeping it injective, which separates the effect of curvature from the effect of folding. No search-space bias is introduced, since the reference value is the best value found on $\mathcal{M}$.

\paragraph{Methods and protocol.} We compare the three restricted methods of Table~\ref{tab:methods-synth}: BOASS-$k_{\mathcal{D}}$, BOASS-$k_{\Phi}$, and BOASS-random-$k_{\Phi}$.
Both BOASS methods use GP-LCB\footnote{GP-UCB for minimization}, $\mu_t-\beta\sigma_t$ with $\beta=2$ fixed, minimised at each step over $2048$ fresh candidates.
Each run has 5 initial Sobol points, common to the three methods. The surrogates are a Matérn-$5/2$ Kernel with a single lengthscale.
The signal variance $s^{2}$ and the length scale $\ell$ are re-estimated by maximum marginal likelihood at every step, while the noise variance is held at its true value $\sigma^{2}=0.01^{2}$.
The prior mean is zero, which is exact here.
Each configuration is repeated over $50$ replicates, and within a replicate the three methods share the same realization of $f$.

\paragraph{Sampling the objective.} 
We instantiate the GP sample $f$ sequentially, by sampling, for each new query, $f_{t+1}\sim\mathcal{N}\left(\mu_t(\mathbf{x}_{t+1}),\sigma_t^2(\mathbf{x}_{t+1})\right)$, where $\mu_t$ and $\sigma^2_t$ are the posterior mean of variance of  the GP given $\{(\mathbf{x}_s,f_s)\}_{s\le t}$, the samples pooled across the evaluated methods. Pooling the samples ensure that all methods share the same GP sample.  . 
The observations are corrupted by i.i.d. Gaussian noise, that is $y_{t+1}=f_{t+1}+\epsilon_{t+1}$ with $\epsilon_{t+1}\sim\mathcal{N}(0,\sigma^2)$.
 
By the chain rule for Gaussian conditionals, this approach leads to an exact draw from the finite-dimensional distribution of the prior.

\paragraph{Non-injectivity of $\mathcal{E}_D$.} For each configuration, we report $\nu$, the fraction of
$\Phi=[-1,1]^p$ whose image is shared with some other point,
\[
  \nu \;=\; \frac{\lambda\!\left(\left\{\phi\in\Phi\ :\ \exists\,\phi'\neq\phi,\
  \mathcal{E}_D(\phi')=\mathcal{E}_D(\phi)\right\}\right)}{\lambda(\Phi)},
\]
with $\lambda$ the Lebesgue measure. Because $W$ has full column rank $p$ almost surely,
$\mathcal{E}_D(\phi)=\mathcal{E}_D(\phi')$ holds iff $g(\phi)=g(\phi')$, and $g$ acts
coordinatewise; hence $\phi$ has a twin iff \emph{at least one} of its coordinates does. Writing
\[
  q_i \;=\; \tfrac{1}{2}\,\lambda\!\left(\left\{u\in[-1,1]\ :\ g_i^{-1}(g_i(u))\neq\{u\}\right\}\right)
\]
for the folded fraction of the $i$-th coordinate map, independence of the coordinates under the uniform measure on $\Phi$ gives
\[
  \nu \;=\; 1-\prod_{i=1}^{p}\left(1-q_i\right).
\]
Each $q_i$ is evaluated numerically: $g_i$ is split into monotone branches at its turning points, each branch covering an interval of values, and $u$ is counted whenever $g_i(u)$ lies in the value range of a branch other than its own. We use a uniform grid of $2\times10^{4}+1$ points, at which the estimate is converged to five decimal places.
At fixed $a$, $\nu$ grows with $p$, since a point is folded as soon as one of its coordinates is.

\paragraph{Results.}
We report, for each replicate, the gap between the best value a method has found and the best value found by any of the three methods on that replicate. The best method of a replicate has a gap of zero by construction, so only comparisons within a configuration are meaningful. Tables~\ref{tab:gap-it30} and~\ref{tab:gap-it60} give the mean gap and its standard error over the $50$ replicates, after $30$ and $60$ iterations.

When the function $\mathcal{E}_D$ is not linear, BOASS-$k_{\mathcal{D}}$ has the lowest mean gap in every configuration after $30$ and $60$ iterations. The two kernels are indistinguishable when the embedding is linear and $p=2$, and they separate as soon as $g$ is curved, already at $a=0.5$ where the map is still injective. The distance between them grows with $p$: at $p=8$ the mean gap of BOASS-$k_{\Phi}$ is about twice that of BOASS-$k_{\mathcal{D}}$ after $30$ iterations, and more after $60$, whether the embedding folds or not. The ambient dimension $D$ has a smaller effect than $p$, in line with the analysis of Section~\ref{sec:regret}, where the rate depends on the intrinsic dimension of $\mathcal{M}$ rather than on $D$. Random search on $\Phi$ is behind both BOASS variants everywhere, so the gain does not come from the restriction of the search space alone.

\definecolor{pband}{gray}{0.92}
 
\begin{table}[htbp]
\centering
\caption{Comparison of BOASS variants in the optimization of a sample from a Gaussian process. Gap to the best method after 30 iterations, in units of $10^{-2}$: mean $\pm$ standard error of the mean over $N=50$ paired replicates. On each replicate the gap of a method is its best objective value found minus the lowest value found by any of the three methods on that same replicate; the best method therefore has a gap of $0$ by construction, and only within-row contrasts are interpretable. \textbf{Bold}: lowest mean in the row, together with every method whose $\hat m \pm \mathrm{s.e.m.}$ interval overlaps that of the lowest.}
\label{tab:gap-it30}
\begin{footnotesize}
\renewcommand{\arraystretch}{0.85}
\begin{tabular}{llrrrr}
\toprule
$p$ & $D$ & \multicolumn{1}{c}{$\nu$} & \multicolumn{1}{c}{BOASS-$k_{\mathcal{D}}$} & \multicolumn{1}{c}{BOASS-$k_{\Phi}$} & \multicolumn{1}{c}{BOASS-random-$k_{\Phi}$} \\
\midrule
\multicolumn{6}{l}{\emph{no fold ($\mathcal{E}_D(\phi)=W\phi$)}} \\
\addlinespace[2pt]
2 & 2 & 0.0\% & $\bm{1.80 \pm 1.50}$ & $\bm{2.70 \pm 1.60}$ & $23.01 \pm 4.80$ \\
2 & 10 & 0.0\% & $\bm{4.58 \pm 1.70}$ & $\bm{3.63 \pm 1.50}$ & $27.35 \pm 4.40$ \\
2 & 20 & 0.0\% & $\bm{6.86 \pm 2.70}$ & $\bm{6.36 \pm 2.60}$ & $27.20 \pm 4.80$ \\
\rowcolor{pband}4 & 4 & 0.0\% & $\bm{17.34 \pm 4.40}$ & $32.80 \pm 4.40$ & $59.70 \pm 5.90$ \\
\rowcolor{pband}4 & 10 & 0.0\% & $\bm{25.08 \pm 4.50}$ & $\bm{18.26 \pm 3.20}$ & $77.67 \pm 5.70$ \\
\rowcolor{pband}4 & 20 & 0.0\% & $\bm{17.47 \pm 4.40}$ & $27.96 \pm 6.00$ & $70.30 \pm 6.30$ \\
8 & 8 & 0.0\% & $\bm{31.44 \pm 5.30}$ & $62.61 \pm 5.90$ & $136.50 \pm 7.70$ \\
8 & 10 & 0.0\% & $\bm{43.86 \pm 6.30}$ & $75.59 \pm 7.00$ & $145.10 \pm 8.90$ \\
8 & 20 & 0.0\% & $\bm{44.97 \pm 6.40}$ & $67.48 \pm 6.50$ & $144.10 \pm 8.40$ \\
\midrule
\multicolumn{6}{l}{\emph{$a=0.5$}} \\
\addlinespace[2pt]
2 & 2 & 0.0\% & $\bm{1.76 \pm 0.83}$ & $7.55 \pm 3.00$ & $24.51 \pm 4.10$ \\
2 & 10 & 0.0\% & $\bm{9.55 \pm 3.40}$ & $20.35 \pm 4.90$ & $30.84 \pm 4.80$ \\
2 & 20 & 0.0\% & $\bm{5.04 \pm 2.50}$ & $11.53 \pm 3.70$ & $25.60 \pm 4.60$ \\
\rowcolor{pband}4 & 4 & 0.0\% & $\bm{16.50 \pm 5.00}$ & $34.98 \pm 5.60$ & $58.51 \pm 7.50$ \\
\rowcolor{pband}4 & 10 & 0.0\% & $\bm{16.39 \pm 3.60}$ & $37.00 \pm 6.10$ & $67.79 \pm 5.70$ \\
\rowcolor{pband}4 & 20 & 0.0\% & $\bm{17.32 \pm 4.00}$ & $\bm{24.05 \pm 4.20}$ & $67.08 \pm 6.10$ \\
8 & 8 & 0.0\% & $\bm{29.31 \pm 4.30}$ & $61.30 \pm 5.60$ & $118.80 \pm 8.50$ \\
8 & 10 & 0.0\% & $\bm{30.68 \pm 4.50}$ & $77.13 \pm 7.20$ & $117.80 \pm 9.20$ \\
8 & 20 & 0.0\% & $\bm{42.72 \pm 6.00}$ & $73.93 \pm 6.70$ & $127.70 \pm 8.80$ \\
\midrule
\multicolumn{6}{l}{\emph{$a=0.7$}} \\
\addlinespace[2pt]
2 & 2 & 25.7\% & $\bm{4.29 \pm 2.00}$ & $14.54 \pm 4.60$ & $37.14 \pm 6.20$ \\
2 & 10 & 25.7\% & $\bm{4.62 \pm 2.60}$ & $13.55 \pm 4.00$ & $36.91 \pm 5.40$ \\
2 & 20 & 25.7\% & $\bm{0.16 \pm 0.09}$ & $10.22 \pm 4.10$ & $28.84 \pm 4.80$ \\
\rowcolor{pband}4 & 4 & 25.7\% & $\bm{13.42 \pm 3.90}$ & $31.43 \pm 4.20$ & $57.10 \pm 6.40$ \\
\rowcolor{pband}4 & 10 & 25.7\% & $\bm{16.74 \pm 3.90}$ & $47.88 \pm 6.60$ & $65.14 \pm 7.10$ \\
\rowcolor{pband}4 & 20 & 25.7\% & $\bm{14.96 \pm 3.70}$ & $35.73 \pm 6.50$ & $68.55 \pm 6.20$ \\
8 & 8 & 40.4\% & $\bm{29.32 \pm 5.50}$ & $81.83 \pm 7.00$ & $118.50 \pm 9.20$ \\
8 & 10 & 40.4\% & $\bm{33.64 \pm 5.80}$ & $72.24 \pm 7.00$ & $115.50 \pm 9.00$ \\
8 & 20 & 40.4\% & $\bm{38.47 \pm 5.30}$ & $75.48 \pm 6.40$ & $113.30 \pm 8.00$ \\
\midrule
\multicolumn{6}{l}{\emph{$a=0.9$}} \\
\addlinespace[2pt]
2 & 2 & 43.6\% & $\bm{2.16 \pm 1.20}$ & $12.57 \pm 3.90$ & $31.16 \pm 5.70$ \\
2 & 10 & 43.6\% & $\bm{4.87 \pm 2.70}$ & $13.52 \pm 3.80$ & $31.94 \pm 5.20$ \\
2 & 20 & 43.6\% & $\bm{3.38 \pm 2.00}$ & $11.43 \pm 3.60$ & $27.82 \pm 5.20$ \\
\rowcolor{pband}4 & 4 & 46.4\% & $\bm{11.46 \pm 3.50}$ & $39.52 \pm 4.60$ & $51.43 \pm 5.50$ \\
\rowcolor{pband}4 & 10 & 46.4\% & $\bm{13.35 \pm 3.10}$ & $40.31 \pm 5.20$ & $66.67 \pm 5.20$ \\
\rowcolor{pband}4 & 20 & 46.4\% & $\bm{8.87 \pm 2.60}$ & $28.44 \pm 4.30$ & $57.38 \pm 5.50$ \\
8 & 8 & 68.0\% & $\bm{28.01 \pm 4.70}$ & $81.38 \pm 9.00$ & $118.70 \pm 9.10$ \\
8 & 10 & 68.0\% & $\bm{27.15 \pm 4.30}$ & $63.07 \pm 5.60$ & $109.30 \pm 7.30$ \\
8 & 20 & 68.0\% & $\bm{40.06 \pm 6.00}$ & $81.85 \pm 6.60$ & $127.30 \pm 8.90$ \\
\midrule
\multicolumn{6}{l}{\emph{$a=1.1$}} \\
\addlinespace[2pt]
2 & 2 & 67.8\% & $\bm{4.84 \pm 2.80}$ & $15.90 \pm 3.80$ & $32.10 \pm 5.90$ \\
2 & 10 & 67.8\% & $\bm{2.59 \pm 2.10}$ & $22.97 \pm 5.30$ & $35.80 \pm 5.60$ \\
2 & 20 & 67.8\% & $\bm{5.84 \pm 2.80}$ & $17.75 \pm 4.60$ & $30.41 \pm 5.20$ \\
\rowcolor{pband}4 & 4 & 87.3\% & $\bm{8.15 \pm 2.60}$ & $33.69 \pm 4.30$ & $37.91 \pm 3.90$ \\
\rowcolor{pband}4 & 10 & 87.3\% & $\bm{13.08 \pm 3.40}$ & $45.76 \pm 6.40$ & $63.14 \pm 6.40$ \\
\rowcolor{pband}4 & 20 & 87.3\% & $\bm{27.56 \pm 6.30}$ & $45.01 \pm 7.60$ & $75.84 \pm 7.30$ \\
8 & 8 & 98.3\% & $\bm{33.17 \pm 6.40}$ & $85.22 \pm 8.30$ & $111.60 \pm 8.80$ \\
8 & 10 & 98.3\% & $\bm{33.48 \pm 5.40}$ & $87.27 \pm 6.60$ & $130.10 \pm 8.00$ \\
8 & 20 & 98.3\% & $\bm{38.85 \pm 6.10}$ & $88.63 \pm 7.40$ & $125.00 \pm 8.90$ \\
\bottomrule
\end{tabular}
\end{footnotesize}
\end{table}
 
\begin{table}[htbp]
\centering
\caption{Comparison of BOASS variants in the optimization of a sample from a Gaussian process.  Gap to the best method after 60 iterations, same specifications as in Table \ref{tab:gap-it30}.}
\label{tab:gap-it60}
\begin{footnotesize}
\renewcommand{\arraystretch}{0.85}
\begin{tabular}{llrrrr}
\toprule
$p$ & $D$ & \multicolumn{1}{c}{$\nu$} & \multicolumn{1}{c}{BOASS-$k_{\mathcal{D}}$} & \multicolumn{1}{c}{BOASS-$k_{\Phi}$} & \multicolumn{1}{c}{BOASS-random-$k_{\Phi}$} \\
\midrule
\multicolumn{6}{l}{\emph{no fold ($\mathcal{E}_D(\phi)=W\phi$)}} \\
\addlinespace[2pt]
2 & 2 & 0.0\% & $\bm{1.77 \pm 1.50}$ & $\bm{0.54 \pm 0.38}$ & $7.65 \pm 2.20$ \\
2 & 10 & 0.0\% & $\bm{4.45 \pm 1.70}$ & $\bm{2.83 \pm 1.40}$ & $9.43 \pm 2.10$ \\
2 & 20 & 0.0\% & $\bm{6.41 \pm 2.60}$ & $\bm{6.15 \pm 2.60}$ & $\bm{10.62 \pm 3.00}$ \\
\rowcolor{pband}4 & 4 & 0.0\% & $\bm{12.35 \pm 4.00}$ & $\bm{19.84 \pm 4.40}$ & $34.01 \pm 4.70$ \\
\rowcolor{pband}4 & 10 & 0.0\% & $17.49 \pm 3.70$ & $\bm{9.62 \pm 2.50}$ & $47.09 \pm 5.20$ \\
\rowcolor{pband}4 & 20 & 0.0\% & $\bm{10.93 \pm 3.70}$ & $\bm{7.35 \pm 2.50}$ & $44.08 \pm 5.40$ \\
8 & 8 & 0.0\% & $\bm{9.36 \pm 3.40}$ & $29.33 \pm 5.10$ & $95.65 \pm 7.20$ \\
8 & 10 & 0.0\% & $\bm{18.04 \pm 5.30}$ & $29.43 \pm 5.10$ & $100.70 \pm 7.70$ \\
8 & 20 & 0.0\% & $\bm{15.68 \pm 3.70}$ & $29.33 \pm 4.40$ & $97.40 \pm 7.80$ \\
\midrule
\multicolumn{6}{l}{\emph{$a=0.5$}} \\
\addlinespace[2pt]
2 & 2 & 0.0\% & $\bm{1.74 \pm 0.83}$ & $6.81 \pm 2.90$ & $12.84 \pm 2.90$ \\
2 & 10 & 0.0\% & $\bm{9.41 \pm 3.40}$ & $\bm{15.99 \pm 4.60}$ & $17.29 \pm 3.70$ \\
2 & 20 & 0.0\% & $\bm{4.94 \pm 2.50}$ & $\bm{9.59 \pm 3.50}$ & $10.77 \pm 3.00$ \\
\rowcolor{pband}4 & 4 & 0.0\% & $\bm{7.12 \pm 2.90}$ & $20.11 \pm 4.80$ & $28.68 \pm 4.60$ \\
\rowcolor{pband}4 & 10 & 0.0\% & $\bm{8.67 \pm 2.80}$ & $16.74 \pm 5.00$ & $40.63 \pm 4.50$ \\
\rowcolor{pband}4 & 20 & 0.0\% & $\bm{11.70 \pm 3.80}$ & $\bm{12.82 \pm 4.00}$ & $41.03 \pm 5.00$ \\
8 & 8 & 0.0\% & $\bm{10.27 \pm 2.70}$ & $31.26 \pm 4.40$ & $77.00 \pm 7.20$ \\
8 & 10 & 0.0\% & $\bm{9.34 \pm 3.00}$ & $45.72 \pm 6.10$ & $77.97 \pm 7.30$ \\
8 & 20 & 0.0\% & $\bm{16.27 \pm 4.40}$ & $42.80 \pm 6.70$ & $90.87 \pm 8.30$ \\
\midrule
\multicolumn{6}{l}{\emph{$a=0.7$}} \\
\addlinespace[2pt]
2 & 2 & 25.7\% & $\bm{4.12 \pm 2.10}$ & $\bm{5.84 \pm 1.90}$ & $14.93 \pm 2.90$ \\
2 & 10 & 25.7\% & $\bm{3.69 \pm 2.40}$ & $\bm{8.92 \pm 3.10}$ & $23.23 \pm 4.80$ \\
2 & 20 & 25.7\% & $\bm{0.03 \pm 0.01}$ & $7.65 \pm 3.90$ & $14.26 \pm 3.90$ \\
\rowcolor{pband}4 & 4 & 25.7\% & $\bm{7.91 \pm 2.80}$ & $15.87 \pm 3.30$ & $30.40 \pm 4.00$ \\
\rowcolor{pband}4 & 10 & 25.7\% & $\bm{8.31 \pm 3.00}$ & $24.29 \pm 4.60$ & $43.37 \pm 5.90$ \\
\rowcolor{pband}4 & 20 & 25.7\% & $\bm{8.92 \pm 3.00}$ & $21.18 \pm 4.90$ & $36.97 \pm 5.00$ \\
8 & 8 & 40.4\% & $\bm{7.97 \pm 2.50}$ & $51.86 \pm 6.90$ & $79.82 \pm 8.80$ \\
8 & 10 & 40.4\% & $\bm{8.13 \pm 2.40}$ & $47.16 \pm 6.00$ & $73.57 \pm 7.70$ \\
8 & 20 & 40.4\% & $\bm{15.43 \pm 3.80}$ & $41.30 \pm 5.80$ & $80.56 \pm 7.50$ \\
\midrule
\multicolumn{6}{l}{\emph{$a=0.9$}} \\
\addlinespace[2pt]
2 & 2 & 43.6\% & $\bm{2.10 \pm 1.20}$ & $\bm{4.76 \pm 1.50}$ & $20.01 \pm 5.10$ \\
2 & 10 & 43.6\% & $\bm{4.83 \pm 2.70}$ & $\bm{10.53 \pm 3.70}$ & $23.16 \pm 5.00$ \\
2 & 20 & 43.6\% & $\bm{3.32 \pm 2.00}$ & $\bm{4.47 \pm 2.50}$ & $14.32 \pm 3.90$ \\
\rowcolor{pband}4 & 4 & 46.4\% & $\bm{6.70 \pm 2.70}$ & $23.35 \pm 4.30$ & $32.05 \pm 4.40$ \\
\rowcolor{pband}4 & 10 & 46.4\% & $\bm{9.75 \pm 2.80}$ & $25.45 \pm 4.80$ & $36.42 \pm 3.90$ \\
\rowcolor{pband}4 & 20 & 46.4\% & $\bm{4.95 \pm 2.10}$ & $12.33 \pm 3.10$ & $33.82 \pm 4.20$ \\
8 & 8 & 68.0\% & $\bm{9.32 \pm 2.70}$ & $53.59 \pm 8.20$ & $86.77 \pm 8.70$ \\
8 & 10 & 68.0\% & $\bm{9.34 \pm 3.00}$ & $42.80 \pm 4.80$ & $70.03 \pm 6.70$ \\
8 & 20 & 68.0\% & $\bm{10.39 \pm 3.70}$ & $45.01 \pm 5.10$ & $86.90 \pm 7.80$ \\
\midrule
\multicolumn{6}{l}{\emph{$a=1.1$}} \\
\addlinespace[2pt]
2 & 2 & 67.8\% & $\bm{4.61 \pm 2.80}$ & $\bm{10.56 \pm 3.30}$ & $15.99 \pm 3.30$ \\
2 & 10 & 67.8\% & $\bm{2.48 \pm 2.10}$ & $13.24 \pm 4.30$ & $25.25 \pm 4.90$ \\
2 & 20 & 67.8\% & $\bm{5.80 \pm 2.80}$ & $\bm{9.78 \pm 3.10}$ & $21.16 \pm 4.40$ \\
\rowcolor{pband}4 & 4 & 87.3\% & $\bm{3.12 \pm 1.70}$ & $23.25 \pm 3.90$ & $26.17 \pm 3.30$ \\
\rowcolor{pband}4 & 10 & 87.3\% & $\bm{9.22 \pm 2.90}$ & $26.04 \pm 5.20$ & $44.96 \pm 5.90$ \\
\rowcolor{pband}4 & 20 & 87.3\% & $\bm{15.90 \pm 4.60}$ & $\bm{21.94 \pm 5.60}$ & $53.12 \pm 6.50$ \\
8 & 8 & 98.3\% & $\bm{8.18 \pm 2.50}$ & $58.61 \pm 7.50$ & $72.53 \pm 8.00$ \\
8 & 10 & 98.3\% & $\bm{13.49 \pm 3.60}$ & $55.22 \pm 6.60$ & $91.12 \pm 7.30$ \\
8 & 20 & 98.3\% & $\bm{8.38 \pm 2.80}$ & $59.35 \pm 6.70$ & $92.40 \pm 8.30$ \\
\bottomrule
\end{tabular}
\end{footnotesize}
\end{table}

\stopcontents[appendices]
\end{appendices}

\end{document}